\documentclass[review]{elsarticle}

\usepackage{soul}
\usepackage{amsmath}
\usepackage{amssymb}
\usepackage{mathtools}
\usepackage{enumitem}
\usepackage{amsthm}

\usepackage{graphicx}
\usepackage{natbib}
\usepackage{url} 
\usepackage{booktabs}
\usepackage{multirow}
\usepackage{longtable}
\usepackage{supertabular}
\usepackage{float}
\usepackage{subfig}
\usepackage{algorithm}
\usepackage{algpseudocode}

\date{}
\journal{XXX}

\begin{document}

\begin{frontmatter}

\title{Differential evolution variants for Searching $D$- and $A$-optimal designs}

\author[mymainaddress,mysecondaryaddress]{Lyuyang Tong}

\address[mymainaddress]{National Engineering Research Center for Multimedia Software, Institute of Artificial Intelligence, School of Computer Science and Hubei Key Laboratory of Multimedia and Network Communication Engineering, Wuhan University, Wuhan, China.}

\address[mysecondaryaddress]{
Department of Computer Science, City University of Hong Kong, Hong Kong.
}

%
%
%

\begin{abstract}
Optimal experimental design is an essential subfield of statistics that maximizes the chances of experimental success. The $D$- and $A$-optimal design is a very challenging problem in the field of optimal design, namely minimizing the determinant and trace of the inverse Fisher information matrix. Due to the flexibility and ease of implementation, traditional evolutionary algorithms (EAs) are applied to deal with a small part of experimental optimization design problems without mathematical derivation and assumption. However, the current EAs remain the issues of determining the support point number, handling the infeasible weight solution, and the insufficient experiment. To address the above issues, this paper investigates differential evolution (DE) variants for finding $D$- and $A$-optimal designs on several different statistical models. The repair operation is proposed to automatically determine the support point by combining similar support points with their corresponding weights based on Euclidean distance and deleting the support point with less weight. Furthermore, the repair operation fixes the infeasible weight solution into the feasible weight solution. To enrich our optimal design experiments, we utilize the proposed DE variants to test the $D$- and $A$-optimal design problems on 12 statistical models. Compared with other competitor algorithms, simulation experiments show that LSHADE can achieve better performance on the $D$- and $A$-optimal design problems.
\end{abstract}

\begin{keyword}
Approximate design, $D$-optimal design, $A$-optimal design, Differential Evolution (DE).
\end{keyword}

\end{frontmatter}


\section{Introduction}
The sub-field of optimal experimental design is an increasingly important  statistical research topic for two reasons.  The first and traditional reason is that experimental costs are always increasing and a well designed study can provide the most accurate inference at minimum cost \cite{smucker2018optimal}.  The second reason is  its newly found applications to devise  optimal subsampling schemes  to make inference from big data based on the selected sample. Optimal design techniques and ideas have been increasingly applied across disciplines to save experimental costs \citep{atkinson1996usefulness}.  Some examples are in the design of dose-response experiments \cite{fedorov2001optimal}, petroleum engineering \cite{song2014d}, clinical pharmacology experiments \cite{ogungbenro2009application}, and image retrieval \cite{he2009laplacian}, among many others.  These optimal experimental designs are model-based and are found by optimizing the design criterion, which is usually formulated as a convex function of the Fisher's information matrix defined below.  This matrix measures the worth of the  design and  depends on the unknown model parameters if the model is nonlinear. $D$- and $A$-locally-optimal designs are among the most popular designs for estimating model parameters but there are other design criteria with different objectives. Fundamentals of optimal design construction are given in \cite{atkinson2007optimum,fedorov2013theory,pukelsheim1991optimal} and  \cite{wong1994comparing} compared robustness properties of  optimal designs under different models and criteria assumptions.

The traditional approach in the statistics literature leans heavily on mathematical derivations. Our experience is that such an approach can be limited, especially when we have a  regression model complicated.  Some of the algorithms in the statistics literature have a proof of convergence but in practice, they may not converge and get trapped in local optima \cite{garcia2020comparison}.  Other problems are usually due to the fact that most proofs are for linear models and have the assumption that the criterion or objective function is differentiable. Nature-inspired metaheuristic algorithms have emerged as powerful tools to solve all kinds of  optimization problems without technical assumptions imposed on the problem \cite{stokes2020using}. Examples of such  evolutionary algorithms (EAs) and nature-inspired algorithms are simulated annealing (SA) \cite{kirkpatrick1983optimization, vcerny1985thermodynamical}, genetic algorithm (GA) \cite{holland1992adaptation,goldberg1989genetic}, particle swarm optimization (PSO) \cite{kennedy1995particle,zhang2020competitive}, Differential evolution (DE) \cite{storn1997differential} and some of them have been applied to solve $D$- and $A$-optimal design problems \cite{garcia2020comparison}. However, it is hard to know the real support point number and appears many similar support points when the design variables can be optimized by naive EAs. It is also challenging that EAs should seek feasible solutions on the support points and their corresponding weights that must be satisfied with the condition of the sum of the corresponding weights to 1. And it is insufficient that existing naive EAs are utilized to test on a small number of optimal design problems.

In this paper, we investigates DE variants for finding $D$- and $A$-optimal designs on several different statistical models. To address above issues, we propose the repair operation to combine similar support points with their corresponding weights based on Euclidean distance and delete the corresponding support point with less weight to automatically recognize the number of support points. To solve the infeasible weight solutions of optimal design problems, the repair operation ensures that the infeasible weight solutions are repaired into the feasible weight solutions. To enrich the $D$- and $A$-optimal design experiment, we adopt the advanced DE variants: JADE \cite{zhang2009jade}, CoDE \cite{wang2011differential}, SHADE \cite{tanabe2013success}, and LSHADE \cite{tanabe2014improving} to test the performance for the $D$- and $A$-optimal design. Simulation experiments of 12 statistical models are shown that LSHADE obtains better performance on the $D$- and $A$-optimal problems among comparative algorithms.

The rest paper is organized as follows: Section 2 describes the related work about the approximate design and the information matrix, the $D$-optimal design, the $A$-optimal design, and the traditional Differential evolution. In Section 3, we utilize the proposed DE variants with the repair operation to solve the $D$- and $A$-optimal design. In Section 4, the simulated experiments of 12 statistical models are conducted on the $D$- and $A$-optimal design criterion, the experiment result of EAs are compared, and we conduct the discussion and analysis. Section 5 gives a conclusion.

\section{Statistical Background}
\subsection{Statistical models, approximate designs and   information matrices}
Our statistical models have the form:
\begin{equation}
Y = \eta \left( {{\bf{x}},\theta } \right) + \varepsilon ,{\rm{ }}{\bf{x}} \in {\bf{X},}
\end{equation}
where ${Y}$ is univariate response and the mean response function $\eta \left( {{\bf{x}},\theta } \right)$ is a known continuous function of a vector of input variables or design variables $\bf{x}$ assumed to belong a user-specified compact design space $\bf{X}$. The vector of unknown model parameters in the model is ${\theta}$ and has dimension ${p}$ and the error ${\varepsilon}$ is error term  with zero mean and constant variance.  All errors are assumed to be identically and independently distributed, but the methodology can be directly extended to the case when errors are independent with a known heteroscedastic structure.

Design problems concern choosing the combinations of the levels of the input variables so that the model unknown parameter  $\theta$ is estimated as precisely as possible, subject to fixed number of observations $N$ determined by the budget.  The optimization problem then determines the optimal number of design points, the optimal values of the input values at each point and the number of  replicates at each point. In practice, approximate designs are used because they are easier to find and study \citep{kiefer74}.  Approximate designs are simply probability measures $\xi$  represented by  ${\xi {\rm{ = }}\left\{ {\left( {{{\bf{x}}_i},{\omega _i}} \right),i = 1,2,...,m} \right\}}$, where each  ${{{\bf{x}}_i} \in {\bf{X}}}$ is a support point and  ${{\omega _i} > 0}$ is the proportion of total observations to be taken at ${{\bf{x}}_i}$, subject to the constraint ${\sum\limits_{i = 1}^m {{\omega _i}}  = 1}$. They are then implemented by taking $[N*\omega_i]$ observations at ${\bf x_i}, i=1,2,\ldots,m$, subject to $[N*\omega_1]+[N*\omega_2]+...+[N*\omega_m]=N$ and $[s]$ is the positive integer nearest to s.

Following convention, the worth of a design is measured by its Fisher information matrix \citep{atkinson2007optimum,pazman1986foundations}. This is because the covariance matrix of the maximum likelihood estimates of $\theta$ is inversely proportional to the determinant of the information matrix and so choosing input variables to make the information matrix large in some sense is equivalent to making the covariance matrix small. For the approximate design ${\xi}$, its normalized $p\times p$ information matrix is given by
\begin{equation}
    {\bf{M}}\left( {\xi ,\theta } \right) = \sum\limits_{i = 1}^m {{\omega _i}} {\bf{M}}\left( {{{\bf{x}}_i},\theta } \right)
\end{equation}
where
\begin{equation}
    {\bf{M}}\left( {{{\bf{x}}_i},\theta } \right) = \frac{{\partial \eta \left( {{{\bf{x}}_i},\theta } \right)}}{{\partial \theta }}\frac{{\partial \eta \left( {{{\bf{x}}_i},\theta } \right)}}{{\partial {\theta ^{\rm{T}}}}}.
\end{equation}

The next two subsections present two common design optimality criteria for estimating model parameters. We note that they are formulated as convex functional functions of the information matrix so that the optimum design found is globally optimum and we can use the directional derivative of the convex function to confirm a design optimality. We also note that if the mean response  is a nonlinear function of $\theta$, then the information matrix contains the unknown parameters.

\subsection{The $D$- and $A$-optimality criteria}
Inference for a single parameter can be based on a confidence intervals and shorter intervals imply more accurate inference. Similarly, for two or more parameters,  smaller area of the confidence region or smaller volume of the confidence ellipsoid  indicates more accurate estimates.  Since the volume of the confidence ellipsoid is inversely proportional to the determinant of the information matrix \cite{pazman1986foundations}, a design that maximizes the determinant of the information matrix among all approximate designs on ${\bf X}$ is desirable.  For nonlinear models,  the determinant depends on  the unknown model parameters  $\theta$ so it cannot be optimized directly.   In practice, the user uses a prior guess of its value (nominal value) and the  resulting design  ${\xi_\theta^*}$ is called locally $D$-optimal. Specifically,
$$ \xi_\theta^*=\text{arg min}~ \text{log}~\text{det}~ {{\bf{M}}^{ - 1}}{\rm{(}}\xi {\rm{,}}\theta {\rm{}})$$
and the minimization is over all approximate designs $\xi$ in ${\bf X}$.  Let $\bar \xi_x$ be the design supported at the point ${x}$ and let $  \xi ' = (1 - \alpha )\xi  + \alpha \bar \xi_x$.  Then

\begin{equation}
    {\bf{M}}(\xi ') = (1 - \alpha ){\bf{M}}(\xi ) + \alpha {\bf{M}}(\bar \xi_x)= (1 - \alpha ){\bf{M}}(\xi ) + \alpha {f(x)f(x)^T}
\end{equation}

\noindent and it follows that if $\psi (x,\xi )= \text{log~det} ({{\bf{M}}^{ - 1}}{\rm{(}}\xi {\rm{,}}\theta {\rm{)}}$, its directional derivative is
\begin{equation}
    \begin{array}{l}
\varphi (x,\xi ) = \mathop {\lim }\limits_{\alpha  \to {0^ + }} \frac{1}{\alpha }[\psi \{ {\bf{M}}(\xi ',\theta )\}  - \psi \{ {\bf{M}}(\xi ,\theta )\} ].\\
\end{array}
\end{equation}

The equivalence theorem \cite{kiefer74} states that a design $\xi^*$
is $D$-optimal if and only if the minimum of the directional derivative $\varphi (x,{\xi ^*}) \ge 0$  for all $x$ in {\bf X}, with inequity at the support points of the design. The sensitive function of the design $\xi$  is:
\begin{equation}
    \begin{array}{l}
S(x,{\xi _D}) = -\varphi (x,{\xi _D})\\
 = trace({\bf{M}}(\bar \xi ,\theta ){{\bf{M}}^{ - 1}}({\xi _D},\theta )) - p\\
 = trace(f(x)f{(x)^T}{{\bf{M}}^{ - 1}}({\xi _D},\theta )) - p\\
 = f{(x)^T}{{\bf{M}}^{ - 1}}({\xi _D},\theta )f(x) - p.
\end{array}
\end{equation}

The practical implication is that $D$-optimality of a design $\xi$ can now be verified by plotting its sensitivity function over the design space and determining if it is the conditions of the equivalence theorem.  If the design space is uni-dimensional, the sensitivity plot provides an easy visual confirmation of whether the design is $D$-optimal.  If a design is not optimal, a $D$-efficiency lower bound of the design $\xi$ can be deduced from properties of a convex functional to be:
\begin{equation}
    D\_Eff(\xi ) = \exp ( - \frac{{\max (S(x,{\xi _D}))}}{p});
\end{equation}

\noindent see details in \cite{fedorov2013optimal} and \cite{pazman1986foundations}.  The usefulness of the bound is that it provides an easy assessment how close a design is to the optimum without knowing the optimum.  For example, when an algorithm to find a $D$-optimal design is terminally prematurely or stalled, the $D$-efficiency lower bound can be computed directly.  If the design has a high efficiency, it may suffice for practical purposes and there is no need to find the true optimum.

The $A$-optimality design criterion minimizes  the sum of the variances of the estimated parameters in the model.  This is equivalent to find a design that minimizes the trace of the inverse of the information matrix over all approximate designs on ${\bf X}$. The resulting design is $A$-optimal
$$ \xi_\theta^*=\text{arg min}~ \text{trace}~ {{\bf{M}}^{ - 1}}{\rm{(}}\xi {\rm{,}}\theta {\rm{}})$$ and a similar argument.  A similar argument shows that the sensitivity function of a design under the $A$-optimality criterion is

$$S(x,{\xi _A})
 = {\mathop{\rm trace}\nolimits} ({\bf{M}}(\bar \xi ,\theta ){{\bf{M}}^{ - 2}}({\xi _A},\theta )) - {\mathop{\rm trace}\nolimits} ({{\bf{M}}^{ - 1}}({\xi _A},\theta ))$$

\noindent and a design $\xi$ is $A$-optimal if and only if $S(x,\xi)$ is bounded above by 0 for all $x\in{\bf X}$, with equality at the support points of $\xi$.  Likewise, the $A$-efficiency lower bound of a design $\xi$ can be directly derived to be

\begin{equation}
    A\_Eff(\xi ) = 1- \frac{{\max (S(x,{\xi _A}))}}{{trace\left( {{{\bf{M}}^{ - 1}}({\xi _A},\theta )} \right)}}
\end{equation}
In our work, if a design has at least a $D$-efficiency or $A$-efficiency of 95\%,  we accept the design as close enough to the optimum.

\subsection{The traditional DE}
Differential Evolution (DE) is a competitive evolutionary algorithm (EA) that utilizes initialization, mutation, crossover, and selection operations to guide the population toward the global optimum \cite{opara2019differential}. The last three operations are repeated in the evolutionary process until the maximum number of  function evaluations (FES) is reached.

In the DE algorithm, each individual in the population encodes the candidate solution as follows:
\begin{equation}\label{EQ_6}
  {{\bf{x}}_{i,G}} = [{\bf{x}}_{i,G}^1,{\bf{x}}_{i,G}^2,...,{\bf{x}}_{i,G}^D],{\rm{ }}i = 1,2,...NP,
\end{equation}
where ${D}$ is the individual dimensionality, ${G}$ is the current generation, and ${NP}$ is the population size. In the initialization operation, the population is randomly generated in the feasible domain.
\begin{equation}\label{EQ_7}
  {{\bf{x}}_{i,G}} = {{\bf{x}}_{\min }} + rand(0,1) \times ({{\bf{x}}_{\max }} - {{\bf{x}}_{\min }}),
\end{equation}
where ${{\bf{x}}_{\min }}$ and ${{\bf{x}}_{\max }}$ are, respectively,  the lower bound and upper bound of the search space,  and ${rand(0,1)}$ is a random number generated from $[0,1]$.

In the mutation operation, the mutation vector ${{\bf{v}}_{i,G}}$ is generated by the mutation strategies:\\ \ \\
\noindent DE/rand/1:
\begin{equation}\label{EQ_8}
  {{\bf{v}}_{i,G}} = {{\bf{x}}_{{r_1},G}} + F \cdot ({{\bf{x}}_{{r_2},G}} - {{\bf{x}}_{{r_3},G}})
\end{equation}
DE/rand/2:
\begin{equation}\label{EQ_9}
  {{\bf{v}}_{i,G}} = {{\bf{x}}_{{r_1},G}} + F \cdot ({{\bf{x}}_{{r_2},G}} - {{\bf{x}}_{{r_3},G}}) + F \cdot ({{\bf{x}}_{{r_4},G}} - {{\bf{x}}_{{r_5},G}})
\end{equation}
DE/best/1:
\begin{equation}\label{EQ_10}
  {{\bf{v}}_{i,G}} = {{\bf{x}}_{best,G}} + F \cdot ({{\bf{x}}_{{r_1},G}} - {{\bf{x}}_{{r_2},G}})
\end{equation}
DE/best/2:
\begin{equation}\label{EQ_11}
  {{\bf{v}}_{i,G}} = {{\bf{x}}_{best,G}} + F \cdot ({{\bf{x}}_{{r_1},G}} - {{\bf{x}}_{{r_2},G}}) + F \cdot ({{\bf{x}}_{{r_3},G}} - {{\bf{x}}_{{r_4},G}}),
\end{equation}
where ${r_1}$, ${r_2}$, ${r_3}$, ${r_4}$, and ${r_5}$ are random integers ranging from ${[1,NP]}$, ${{\bf{x}}_{best,G}}$ is the best individual in  the ${G}^{th}$ generation population, and ${F}$ is the positive control parameter called the scaling factor.

In the crossover operation, the trial vector ${{\bf{u}}_{i,G}}$ is generated by the target vector ${{\bf{x}}_{i,G}}$ and the mutation vector ${{{\bf{v}}_{i,G}}}$. The binomial crossover can be expressed as follows:
\begin{equation}\label{EQ_12}
\mathbf{u}_{i, G}=\left\{\begin{array}{cc}
\mathbf{v}_{i, G} & \text { if }(\operatorname{rand}(0,1) \leq \mathrm{CR}) \text { or }\left(j==j_{\text {rand }}\right) \\
\mathbf{x}_{i, G} & \text { otherwise }
\end{array}\right.
\end{equation}
where CR is the positive control parameter called the crossover probability and $j_{rand}$ is a random integer ranged from ${[1,D]}$.

In the selection operation, DE selects the next candidate solution from the target vector and the trial vector  as follows:
\begin{equation}\label{EQ_13}
  {{\bf{x}}_{i,G + 1}} = \left\{ {\begin{array}{*{20}{c}}
{{{\bf{u}}_{i,G}}}&{\text{if} ~f({{\bf{u}}_{i,G}}) \le f({{\bf{x}}_{i,G}})}\\
{{{\bf{x}}_{i,G}}}&{\text{otherwise}}
\end{array}} \right.
\end{equation}

\section{DE variants with repair operation for $D$- and $A$-optimal design}

This paper uses DE variants to find $D$- and $A$-optimal designs for  statistical models and compare their performance. And the D-optimal design and the A-optimal design problem is considered as single objective optimization problem. A design is represented as an individual in the EA algorithm and the $k^{th}$ individual ${\bf{X_k}}$ is encoded as
\begin{equation}\label{EQ_14}
  \bf{X_k}= {[{{\bf{x}}_1},{\omega _1},...,{{\bf{x}}_i},{\omega _i},...,{{\bf{x}}_{nSupp}},{\omega _{nSupp}}]},
\end{equation}

\noindent where ${nSupp}$ is the  number of support points of the design,  ${\omega _i}$ is the proportion of observations to  take at $x_i$ and ${\sum\limits_{i = 1}^m {{\omega _i}}  = 1}$. A  typical choice for  ${nSupp}$ is a number slightly larger or equal to the number of parameters in the model.  Figure \ref{fig:individual} shows how  the $k^{th}$ individual is encoded as a design:
\begin{figure}[h]
    \centering
    \includegraphics[width=0.99\textwidth]{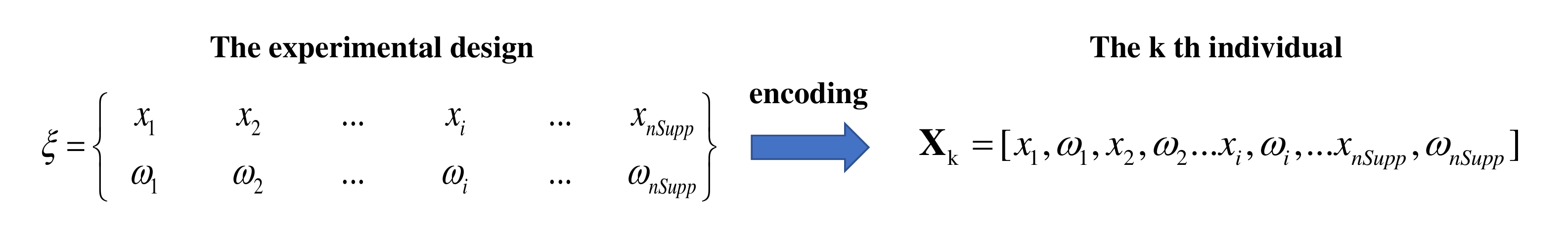}
    \caption{The $k^{th}$ individual encoding.}
    \label{fig:individual}
\end{figure}

\subsection{Repair operation}
Our experience is that the current EAs are not able to find the correct number of support points in the optimal design effectively. We introduce a repair operation to tackle the issue for DE variants. In the repair process, since the true number of  support points is unknown and there are usually similar support points in the generated designs, we combine the weights of similar support points as the weight for the combined point using Euclidean distance clustering. Support points with very small weights are deleted and have their weights distributed uniformly across the rest of the support points after ensuring that the sum of the weights is one. The pseudo code of the repair operation is illustrated in  Algorithm \ref{The_repair_operation}.

\begin{algorithm}[htpb]
\small
\caption{The repair operation}
\label{The_repair_operation}
\begin{algorithmic}[1]
    \State {$\bf{X}= min(max(\bf{X},\bf{x_{min}}),\bf{x_{max}});$}
    \State{$index = D/nSupp : D/nSupp : D;$}
    \State {$\bf{X}(:,index)=\bf{X}(:,index)./repmat(sum(\bf{X}(:,index),2),1,length(index));$}
    \For {$k=1$ to $size(\bf{X},1)$}
        \State{$Temp= reshape(\bf{X}(k,:),[\quad],nSupp);$}
        \State{$aux\_w= Temp(end,:)';aux\_x = Temp(1:end-1,:)';$}
        \State{$data\_not\_order = [aux\_w, aux\_x];[rows, cols] = size(data\_not\_order);$}
        \State{$data = sortrows(data\_not\_order, (2:cols))$};
        \State{$Distance = squareform(pdist(data(:,2:size(data,2)),'euclidean'));$}
        \State{$rows = size(data,1);i = 1; j = 2;$}
        \While{${i < rows}$}
            \While {${j \leq rows}$}
                \If{$Distance(i,j)< \epsilon$}
                    \State {$e = mean([data(i,:);data(j,:)]); e(1) = 2*e(1);$}
                    \State{$data(j,:) = [\quad];Distance(:,j) = [\quad];rows = rows - 1;$}
                    \State{$data(i,:) = e;$}
                \Else
                    \State{$j = j + 1;$}
                \EndIf
            \EndWhile
            \State{$i = i + 1; j = i + 1;$}
            \State{$Distance=squareform(pdist(data(:,2:size(data,2)),'euclidean'));$}
        \EndWhile
         \State{$condition = data(:,1)< \hat{w}; data(condition,:) = [\quad];$}
        \If{$size(data,1)<nSupp$}
            \State{$data=[data;repmat(zeros(1,size(data,2)),nSupp-size(data,1),1);];$}
        \EndIf
        \State{$data=[data(:,2:end),data(:,1)];X(k,:)=reshape(data',1,[\quad]);$}
    \EndFor
    \State {$index = D/nSupp : D/nSupp : D;$}
    \State {$\bf{X}(:,index)=\bf{X}(:,index)./repmat(sum(\bf{X}(:,index),2),1,length(index));$}
\end{algorithmic}
\end{algorithm}

\subsection{DE variants}
DE is a well known and widely used evolutionary metaheuristic algorithm that can search in large spaces for a solution or an approximate solution to a complex optimization problem   without the use of gradient information. Like any notable metaheuristic algorithm, it has many variants which are modifications of DE in different ways for improved performance. We focus on the following  seemingly more popular DE variants, such as, JADE, CoDE, SHADE, and LSHADE, and compare their abilities   to search for  $D$- and $A$-optimal designs for different statistical models.

JADE \cite{zhang2009jade}  is based on a novel greedy mutation strategy called DE/current-to-pbest/1 with the archive. The DE/current-to-pbest/1 strategy utilizes the multiple best solutions to balance the greediness of the mutation strategy and the population diversity. This archive provides information about the direction of progress and contributes to the improvement of the diversity of the population. The DE/current-to-pbest/1 strategy with the archive is  as follows:
\begin{equation}\label{EQ_15}
    {{\bf{v}}_{i,G}} = {{\bf{x}}_{i,G}} + F \cdot ({{\bf{x}}_{pbest,G}} - {{\bf{x}}_{i,G}}) + F \cdot ({{\bf{x}}_{{r_1},G}} - {{\bf{\tilde x}}_{{r_2},G}})
\end{equation}
where ${{\bf{x}}_{pbest,G}}$ is randomly selected in the top of the current population with the probability ${p}$. JADE also adopts the parameter adaptation to generate the scaling factors and crossover probabilities based on the success record. During such an update, ${F}$ and ${Cr}$ values have been produced by successful offspring in recent generations. The pseudocode of JADE is illustrated in the Algorithm \ref{alg2}.

\begin{algorithm}[htpb]
\small
\caption{JADE\cite{zhang2009jade}}
\label{alg2}
\begin{algorithmic}[1]
    \State {Set the initial generation ${G}=0$, ${\mu_F=0.5}$, ${\mu_{CR}=0.5}$, ${A=\emptyset}$;}
    \State {\bf Step 1: The initialization operation}
    \State {Randomly initialize the population;}
    \While{the stopping condition is not satisfied}
        \State {${S_F=\emptyset}$; ${S_{CR}=\emptyset}$;}
        \For {$i=1$ to $NP$}
            \State {$F_i=randc_i(\mu_F,0.1);$ $CR_i=randn_i(\mu_{CR}, 0.1);$}
            \State {\bf  Step 2: The ``DE/current-to-pbest/1'' mutation with the archive operation}
            \State {${{\bf{v}}_{i,G}} = {{\bf{x}}_{i,G}} + F \cdot ({{\bf{x}}_{pbest,G}} - {{\bf{x}}_{i,G}}) + F \cdot ({{\bf{x}}_{{r_1},G}} - {{\bf{\tilde x}}_{{r_2},G}})$;}
            \State {\bf  Step 3: The binomial crossover operation}
            \For {$j=1$ to $D$}
                \If {{${\left( {rand_{i,j}(0,1) \le CR} \right){\rm{ or }}\left( {{j_{rand}}==j} \right)}$}}
                    \State ${{\bf {u}}_{i,G}^j = {\bf {v}}_{i,G}^j}$;
                \Else
                    \State ${{\bf {u}}_{i,G}^j = {\bf {x}}_{i,G}^j}$;
                \EndIf
            \EndFor
            \State {\bf  Step 4: The selection operation}
            \If {$ f({{\bf {u}}_{i,G}}) \leq f({{\bf {x}}_{i,G}})$}
                \State ${{\bf {x}}_{i,G+1}={\bf {u}}_{i,G}}; {{A}\longleftarrow{{\bf {x}}_{i,G}}}; S_F\longleftarrow{F_i}; S_{CR} \longleftarrow {CR_i};$
            \EndIf
        \EndFor
        \State {Randomly remove solutions of the archive $|| A || > NP$;}
        \State {${\mu_{F}=(1-c)\cdot\mu_{F}+c\cdot mean_L(S_{F})}$; ${\mu_{CR}=(1-c)\cdot\mu_{CR}+c\cdot mean_A(S_{CR})}$;}
        \State {${G = G + 1}$};
    \EndWhile
\end{algorithmic}
\end{algorithm}

The composite DE (CoDE) (\cite{wang2011differential}) is proposed to generate trial vectors by different DE mutation strategies and the control parameter pool. The CoDE utilizes three mutation strategies (DE/rand/1, DE/rand/2, and DE/current-to-rand/1) and the control parameter pool is $[F=1.0, CR=0.1], [F=1.0, CR=0.9],$ and $[F=0.8, CR=0.2]$.
Three candidates are generated for each trial vector by using an independent strategy, which uses a set of control parameters that are randomly selected for each trial vector. The final trial vector is chosen from the candidate with the best fitness value. The rand/1 and rand/2 strategy can generate the population to enhance the diversity in the random direction, and the current-to rand/1 strategy utilizes the rotation-invariant arithmetic crossover strategy  suitable for the rotation problem. The various control parameters balance the global exploration and exploitation in the search space and Algorithm \ref{CoDE} is a pseudo code of CoDE.

\begin{algorithm}[htpb]
\small
\caption{The composite DE (CoDE) \cite{wang2011differential}}
\label{CoDE}
\begin{algorithmic}[1]
    \State {Set the initial generation ${G}=0$, the parameter pool:$[F=1.0, CR=0.1], [F=1.0, CR=0.9],$ and $[F=0.8, CR=0.2]$;}
    \State {Randomly initialize the population;}
    \While{$FES < MaxFEs$}
        \For {$i=1$ to $NP$}
            \State {Utilize three mutation strategies ``rand/1/bin'',``rand/2/bin'', ``current-to-rand/1/bin'' to generate the trial vectors ${{\bf {u}}_{i_1,G}}$, ${{\bf {u}}_{i_2,G}}$, and ${{\bf {u}}_{i_3,G}}$;}
            \State {Choose the best trial vector ${{\bf {u}}_{i,G}^*}$ from three trial vectors ${{\bf {u}}_{i_1,G}}$, ${{\bf {u}}_{i_2,G}}$, and ${{\bf {u}}_{i_3,G}}$;}
            \If {$ f({{\bf {u}}_{i,G}^*}) \leq f({{\bf {x}}_{i,G}})$}
                \State ${{\bf {x}}_{i,G+1}={\bf {u}}_{i,G}^*}$;
            \EndIf
            \State {${FES = FES + 3}$};
        \EndFor
        \State {${G = G + 1}$};
    \EndWhile
\end{algorithmic}
\end{algorithm}

SHADE \cite{tanabe2013success} adopts the different parameter mechanism based on the successful parameter historical record. SHADE calculates the history memory parameter content ${M_F}$ and ${M_{CR}}$ using the weighted Lehmer mean to generate the parameter ${F}$ and ${CR}$ adaptively.  Algorithm \ref{SHADE} is our pseudo code for SHADE. In LSHADE \cite{tanabe2014improving}, the linear population size reduction strategy is utilized to improve performance and the population size is dynamically reduced by the linear function. In the   early search phase, the emphasis is on exploration  and at the end of the search process,  a smaller population is used to exploit  the best regions. Algorithm \ref{LSHADE} is our pseudo code for LSHADE.

\begin{algorithm}[htpb]
\small
\caption{SHADE \cite{tanabe2013success} }
\label{SHADE}
\begin{algorithmic}[1]
    \State {Set the initial generation ${G}=0$, ${M_F=0.5}$, ${M_{CR}=0.5}$, ${A=\emptyset}$;}
    \State {Randomly initialize the population;}
    \State {Index counter k=1};
    \While{the stopping condition is not satisfied}
        \State {${S_F=\emptyset}$; ${S_{CR}=\emptyset}$;}
        \For {$i=1$ to $NP$}
            \State {$r_i$=Select from [1,H] randomly;}
            \State {$F_{i,G}=randc_i(M_{F,r_i},0.1);$ $CR_{i,G}=randn_i(M_{CR,r_i}, 0.1);$}
            \State {${{\bf{v}}_{i,G}} = {{\bf{x}}_{i,G}} + F_{i,G} \cdot ({{\bf{x}}_{pbest,G}} - {{\bf{x}}_{i,G}}) + F_{i,G} \cdot ({{\bf{x}}_{{r_1},G}} - {{\bf{\tilde x}}_{{r_2},G}})$;}
            \For {$j=1$ to $D$}
                \If {{${\left( {rand_{i,j}(0,1) \le CR} \right){\rm{ or }}\left( {{j_{rand}}==j} \right)}$}}
                    \State ${{\bf {u}}_{i,G}^j = {\bf {v}}_{i,G}^j}$;
                \Else
                    \State ${{\bf {u}}_{i,G}^j = {\bf {x}}_{i,G}^j}$;
                \EndIf
            \EndFor
            \If {$ f({{\bf {u}}_{i,G}}) \leq f({{\bf {x}}_{i,G}})$}
                \State ${{\bf {x}}_{i,G+1}={\bf {u}}_{i,G}}; {{A}\longleftarrow{{\bf {x}}_{i,G}}}; S_F\longleftarrow{F_{i,G}}; S_{CR} \longleftarrow {CR_{i,G}};$
            \EndIf
        \EndFor
        \State {Randomly remove solutions of the archive $\| A \| > NP$;}
        \If {$S_F \neq \emptyset$ and $S_{CR} \neq \emptyset$}
            \State {Update ${M_{CR,k}}$, ${M_{F,k}}$ based on ${S_{CR}}$, ${S_{F}}$;}
            \State {$k++$};
            \If {$k>H$}
                \State {$k=1$};
            \EndIf
        \EndIf
        \State {${G = G + 1}$};
    \EndWhile
\end{algorithmic}
\end{algorithm}

\begin{algorithm}[htpb]
\small
\caption{LSHADE \cite{tanabe2014improving}}
\label{LSHADE}
\begin{algorithmic}[1]
    \State {Set the initial generation ${G}=0$, ${M_F=0.5}$, ${M_{CR}=0.5}$, ${A=\emptyset}$, Index counter k=1;}
    \State {Randomly initialize the population;}
    \While{the stopping condition is not satisfied}
        \State {${S_F=\emptyset}$; ${S_{CR}=\emptyset}$;}
        \For {$i=1$ to $NP$}
            \State {$r_i$=Select from [1,H] randomly;}
            \State {$F_{i,G}=randc_i(M_{F,r_i},0.1);$
            \bf{If} $M_{CR,r_i}=\perp,$ $CR_{i,G}=0.$ \bf{Else} $CR_{i,G}=randn_i(M_{CR,r_i}, 0.1);$}
            \State {${{\bf{v}}_{i,G}} = {{\bf{x}}_{i,G}} + F_{i,G} \cdot ({{\bf{x}}_{pbest,G}} - {{\bf{x}}_{i,G}}) + F_{i,G} \cdot ({{\bf{x}}_{{r_1},G}} - {{\bf{\tilde x}}_{{r_2},G}})$;}
            \For {$j=1$ to $D$}
                \If {{${\left( {rand_{i,j}(0,1) \le CR} \right){\rm{ or }}\left( {{j_{rand}}==j} \right)}$}}
                    \State ${{\bf {u}}_{i,G}^j = {\bf {v}}_{i,G}^j}$;
                \Else
                    \State ${{\bf {u}}_{i,G}^j = {\bf {x}}_{i,G}^j}$;
                \EndIf
            \EndFor
            \If {$ f({{\bf {u}}_{i,G}}) \leq f({{\bf {x}}_{i,G}})$}
                \State ${{\bf {x}}_{i,G+1}={\bf {u}}_{i,G}}; {{A}\longleftarrow{{\bf {x}}_{i,G}}}; S_F\longleftarrow{F_{i,G}}; S_{CR} \longleftarrow {CR_{i,G}};$
            \EndIf
        \EndFor
        \State {Randomly remove solutions of the archive $A > NP$;}
        \If {$S_F \neq \emptyset$ and $S_{CR} \neq \emptyset$}
            \State{Update ${M_{CR,k}}$, ${M_{F,k}}$ based on ${S_{CR}}$, ${S_{F}}$;}
            \State{$k++$};
            \If {$k>H$}
                \State{$k=1$};
            \EndIf
        \EndIf
        \State {${NP_{G + 1}} = round[(\frac{{{NP^{\min }} - {NP^{init}}}}{{MaxFES}}) \cdot FES + {NP^{init}}]$;}
        \If{$NP_{G}<NP_{G+1}$}
            \State {Sort individuals in the population on the their fitness values and delete lowest ${NP_G}-{NP_{G+1}}$ individuals, and resize archive size ${A}$ according to new population;}
        \EndIf
        \State {${G = G + 1}$};
    \EndWhile
\end{algorithmic}
\end{algorithm}

\section{Experimental Results}
In this section, we conduct experiments to find the $D$- and $A$-optimal design experiment for 12 statistical problems in \cite{garcia2020comparison}. In Table \ref{table1}, the usefulness of these models is described in the accompanying paper cited.   Problems 1 and 4 are exponential models that can be written as a sum of two exponential terms \cite{dette2006locally,dette2011note}. The problems 2 and 8 combine the linear term with the interaction terms \cite{yang2013optimal,johnson1983some} and Problem 5 is for a  Kinetics of the catalytic dehydrogenation of n-hexil alcohol model \cite{box1965experimental}. Problem 6 is the Michaelis-Menten model \cite{lopez2002design} and  Problem 7 concerns a mixture-type inhibition model \cite{bogacka2011optimum}. Problems 9-11 are for the probit regression model, logistic regression model, gamma regression model, respectively and Problems 3 and 12 are the multinomial logistic regression models with different number of factors \cite{yang2013optimal,garcia2020comparison}. Our goal is to apply the various algorithms to find $D$- and $A$-optimal designs for these models.  This means we first  determine the number of support points in the optimal design, where they are in the design space and the proportion of observations to take at each of the design points.

In the subsection, we apply several algorithms and compare their performance to find $D$- and $A$-optimal designs for 12 different types of models.  They include traditional metaheuristic algorithms like Simulating Annealing (SA), Standard Particle Swarm Optimization (SPSO) \cite{zambrano2013standard}, Genetic Algorithm (GA), Optimal foraging algorithm (OFA) \cite{zhu2017optimal}  and Competitive Swarm Optimizer (CSO) \cite{cheng2014competitive}.  Additionally, we include DE variants, which are various modifications of differential evolution, a widely used evolutionary algorithm and the variants of interest   are JADE, CoDE, SHADE, and LSHADE to find the $D$-optimal and $A$-optimal design for the 12 statistical models. In the experimental setting, the population size is 50, the function evaluations (FES) of Problem 1-7 are 10000, and the function evaluations of Problem 8-12 are 500000. Each algorithm runs 25 times on each problem independently.

\begin{table}[htpb]
\tiny
  \centering
  \caption{Design Problems 1-12 for Various Statistical Models, along with their sources in square parentheses.}
    \begin{tabular}{clcccc}
    \hline
    \textbf{Problem} & \textbf{Model} & \textbf{Factors} & \textbf{Parameters} & \begin{tabular}[c]{@{}c@{}}\textbf{Assumed  number of }  \\ \textbf{Support points} \end{tabular} & \textbf {Variables} \\
    \hline
    1 \cite{dette2006locally}    &
    ${\begin{array}{l}
        Y{\rm{\sim}}{\theta _1}{e^{{\rm{ - }}{\theta _2}x}} + {\theta _3}{e^{{\rm{ - }}{\theta _4}x}} + N\left( {0,{\sigma ^2}} \right),x \in \left[ {0,3} \right]\\
        g(\theta ) = \theta\\
        \text{\rm{nominal values }} \theta {\rm{ = }}{\left( {{\rm{1,1,1,2}}} \right)^T}
    \end{array}}$
    & 1     & 4     & 6     & 12 \\
    \hline

    2   \cite{yang2013optimal}  &
    $\begin{array}{l}
        Y{\rm{\sim}}{\theta _1} + {\theta _2}{x_1} + {\theta _3}x_1^2 + {\theta _4}{x_2} + {\theta _5}{x_1}{x_2} + N\left( {0,{\sigma ^2}} \right),\\
        \left( {{x_1},{x_2}} \right) \in \left[ { - 1,1} \right] \times \left[ {0,1} \right]\\
        g(\theta ) = \theta
    \end{array}$
    & 2     & 5     & 10    & 30 \\
    \hline

    3 \cite{yang2013optimal}    &
    $\begin{aligned}
&Y \sim \pi_{i}(\mathbf{x})=P\left(Y_{i}=1 \mid \mathbf{x}\right)
=\frac{e^{h(\mathbf{x})^{T} \theta_{i}}}{1+e^{h(\mathbf{x})^{T} \theta_{1}+e^{h(\mathbf{x})^{T} \theta_{2}}}} \\
&\mathbf{x} \in[0,6]^{3} ,i=1,2 \\
&h(\mathbf{x})=\left[1, \mathbf{x}^{T}\right]^{T} ; g(\theta)=\theta \\
&\text { nominal values } \theta_{1}=(1,1,-1,2)^{T} ; \theta_{2}=(-1,2,1,-1)^{T}
\end{aligned}$
    & 3     & 8     & 15    & 60\\
    \hline

    4 \cite{dette2011note}    &
    $\begin{array}{l}
        Y{\rm{\sim}}{\theta _1}{e^{{\theta _2}x}} + {\theta _3}{e^{{\theta _4}x}} + N\left( {0,{\sigma ^2}} \right),\\
        x \in \left[ {0,{\rm{1}}} \right]\\
        g(\theta ) = \theta\\
        \text{\rm{nominal values }}\theta {\rm{ = }}{\left( {{\rm{1,0}}{\rm{.5,1,1}}} \right)^T}
    \end{array}$       & 1     & 4     & 8     & 16 \\
    \hline

    5  \cite{box1965experimental}   &
    $\begin{array}{l}
        Y{\rm{\sim}}\frac{{{\theta _1}{\theta _3}{x_1}}}{{1 + {\theta _1}{x_1} + {\theta _2}{x_2}}} + N\left( {0,{\sigma ^2}} \right),\\
        \left( {{x_1},{x_2}} \right) \in \left[ {0,3} \right] \times \left[ {0,3} \right]\\
        g(\theta ) = \theta \\
        \text{\rm{nominal values }}\theta  = {(2.9,12.2,0.69)^T}
    \end{array}$
    & 2     & 3     & 10    & 30  \\
    \hline

    6   \cite{lopez2002design}  &
    $\begin{array}{l}
        Y{\rm{\sim}}\frac{{{\theta _1}x}}{{{\theta _2} + x}} + N\left( {0,{\sigma ^2}} \right),\\
        x \in \left[ {0,5} \right]\\
        g(\theta ) = \theta \\
        \text{\rm{nominal values }}\theta  = {(1,1)^T}
    \end{array}$ & 1     & 2     & 5     & 10 \\
    \hline

    7  \cite{bogacka2011optimum}   &
    $\begin{array}{l}
        Y{\rm{\sim}}\frac{{{\theta _1}{x_1}}}{{\left( {1 + \frac{{{x_2}}}{{{\theta _3}}}} \right){\theta _2} + \left( {1 + \frac{{{x_2}}}{{{\theta _4}}}} \right){x_1}}} + N\left( {0,{\sigma ^2}} \right),\\
        \left( {{x_1},{x_2}} \right) \in \left[ {0,30} \right] \times \left[ {0,60} \right]\\
        g(\theta ) = \theta \\
        \text{\rm{nominal values }}\theta  = {(1,4,2,4)^T}
    \end{array}$
    & 2     & 4     & 5     & 15 \\
    \hline

    8  \cite{johnson1983some}   &
    $\begin{array}{l}
        Y{\rm{\sim}}{\theta _1}{x_1} + {\theta _2}{x_2} + {\theta _3}{x_3} + {\theta _4}{x_1}{x_2} + {\theta _5}{x_1}{x_3} + {\theta _6}{x_2}{x_3}\\
        + \frac{{{\theta _7}}}{{{x_1}}} + \frac{{{\theta _8}}}{{{x_2}}} + \frac{{{\theta _9}}}{{{x_3}}} + N\left( {0,{\sigma ^2}} \right),\\
        \left( {{x_1},{x_2},{x_3}} \right) \in \left[ {0.5,2} \right] \times \left[ {0.5,2} \right] \times \left[ {0.5,2} \right]\\
        g(\theta ) = \theta
    \end{array}$       & 3     & 9     & 20    & 80  \\
    \hline

    9  \cite{garcia2020comparison}   &
    $\begin{array}{l}
        Y{\rm{\sim Binomial}}\left( {1,\mu } \right);\mu  = \Phi (h{(x)^T}\theta )\\
        \text{$\Phi ( \cdot )${\rm{is the normal cumulative  distribution function}}}\\
        h(x) = {[1,{x^T}]^T};x \in {\left[ { - 2,2} \right]^5}\\
        g(\theta ) = \theta \\
        \text{\rm{nominal values }}\theta  = {(0.5,0.7,0.18, - 0.20, - 0.58,0.51)^T}
    \end{array}$       & 5     & 6     & 25    & 150 \\
    \hline

    10 \cite{garcia2020comparison}   &
    $\begin{array}{l}
        Y{\rm{\sim Binomial}}\left( {1,\mu } \right);\mu  = \frac{1}{{1 + {e^{ - h{{(x)}^T}\theta }}}}\\
        h(x) = {[1,{x^T}]^T};x \in {\left[ { - 2,2} \right]^5}\\
        g(\theta ) = \theta \\
        \text{\rm{nominal values }}\theta  = {(0.5,0.7,0.18, - 0.20, - 0.58,0.51)^T}
    \end{array}$      & 5     & 6     & 25    & 150 \\
    \hline

    11 \cite{garcia2020comparison}   &
    $\begin{array}{l}
        Y{\rm{\sim}}\Gamma \left( {1,\mu } \right);\mu  = {\left( {{\theta _1}{x_1} + \sum\nolimits_{i = 2}^5 {{x_{i - 1}}{x_i}{\theta _i}} } \right)^2}\\
        x \in {\left[ {0,10} \right]^5}\\
        g(\theta ) = \theta \\
        \text{\rm{nominal values }}\theta  = {(0.25,0.5,0.20,0.58,0.51)^T}
    \end{array}$       & 5     & 5     & 25    & 150 \\
    \hline

    12 \cite{garcia2020comparison}   &
    $\begin{aligned}
&Y \sim \pi_{i}(\mathbf{x})=P\left(Y_{i}=1 \mid \mathbf{x}\right)
=\frac{e^{h(\mathbf{x})^{T} \theta_{i}}}{1+e^{h(\mathbf{x})^{T} \theta_{1}+e^{h(\mathbf{x})^{T} \theta_{2}}}} \\
&\mathbf{x} \in[0,3]^{10} ,i=1,2 \\
&h(\mathbf{x})=\left[1, \mathbf{x}^{T}\right]^{T} ; g(\theta)=\theta \\
&\text { nominal values }\\ &\theta_{1}=(1,1,-1,2,-2,1,0.5,-0.25,0.5,-0.75,2)^{T} ; \\
&\theta_{2}=(-1,2,1,-1,-1,-1,-0.5,1,0.75,0.25,-2)^{T}
\end{aligned}$
           & 10    & 22    & 17    & 187\\
    \hline
    \end{tabular}%
  \label{table1}%
\end{table}%


\subsection{Algorithm-generated designs under the $D$-optimality criterion}

Tables \ref{table2_1}, \ref{table2_2}, and \ref{table2_3} show the summary statistics for $D$-optimality criterion values of designs generated by different algorithms for 1-12 models. The corresponding columns show the best, median, worst, and mean objective function value on the $D$-optimal design criterion and the last two columns represent the standard derivation of each algorithm and the average Computation time in seconds. From the summary tables, we observe that LSHADE outperforms the other algorithms for the problems 2, 3, 5, 7, and 12 in terms of median value. SHADE also achieves better performance on the problems 8-10 and JADE obtains better result on problem 11 in terms of the median value. Compared with CSO, LSHADE obtains the inferior results on the problems 4 and 6, and outperforms on the rest problems. Overall the DE variants outperform OFA and the traditional EA such as the SA, SPSO, GA.

\begin{table}[htbp]
\tiny
  \centering
  \caption{Summary statistics for $D$-optimality criterion values of designs generated by different algorithms for 1-4 models.}
    \begin{tabular}{clcccccc}
    \toprule
    Problem & {Algorithm} & {best} & {median} & {worst} & {mean} & {std} & {time} \\
    \midrule
    \multirow{9}[18]{*}{1} & SA    & 2.0950E+01 & 2.1369E+01 & 2.1931E+01 & 2.1441E+01 & 2.9427E-01 & 5.3642E+00 \\
\cmidrule{2-8}          & SPSO  & 2.0517E+01 & 2.0585E+01 & 2.0654E+01 & 2.0587E+01 & 3.7838E-02 & 6.1441E+00 \\
\cmidrule{2-8}          & GA    & 2.0525E+01 & 2.0558E+01 & 2.0602E+01 & 2.0556E+01 & 2.0318E-02 & 4.2281E+00 \\
\cmidrule{2-8}          & OFA   & 2.0523E+01 & 2.0573E+01 & 2.0653E+01 & 2.0575E+01 & 3.1526E-02 & 3.7729E+00 \\
\cmidrule{2-8}          & CSO   & 2.0508E+01 & 2.0508E+01 & 2.0517E+01 & 2.0509E+01 & 1.6465E-03 & 3.8454E+00 \\
\cmidrule{2-8}          & JADE  & 2.0508E+01 & 2.0508E+01 & 2.0508E+01 & 2.0508E+01 & 6.8025E-10 & 3.9214E+00 \\
\cmidrule{2-8}          & CoDE  & 2.0508E+01 & 2.0509E+01 & 2.0511E+01 & 2.0509E+01 & 4.7857E-04 & 4.3009E+00 \\
\cmidrule{2-8}          & SHADE & 2.0508E+01 & 2.0508E+01 & 2.0508E+01 & 2.0508E+01 & 1.2866E-08 & 4.5524E+00 \\
\cmidrule{2-8}          & LSHADE & 2.0508E+01 & 2.0508E+01 & 2.0508E+01 & 2.0508E+01 & 1.8354E-05 & 6.0343E+00 \\
    \midrule
    \multirow{9}[18]{*}{2} & SA    & 5.6730E+00 & 6.0433E+00 & 7.5963E+00 & 6.1876E+00 & 4.9394E-01 & 8.2503E+00 \\
\cmidrule{2-8}          & SPSO  & 5.6155E+00 & 6.0435E+00 & 6.4494E+00 & 6.0886E+00 & 2.2373E-01 & 9.3157E+00 \\
\cmidrule{2-8}          & GA    & 6.1859E+00 & 6.5607E+00 & 7.0657E+00 & 6.6318E+00 & 2.2003E-01 & 6.9724E+00 \\
\cmidrule{2-8}          & OFA   & 5.4921E+00 & 6.0236E+00 & 6.6990E+00 & 6.0876E+00 & 3.3401E-01 & 8.0033E+00 \\
\cmidrule{2-8}          & CSO   & 5.0443E+00 & 5.2064E+00 & 5.5156E+00 & 5.2167E+00 & 1.1736E-01 & 5.5990E+00 \\
\cmidrule{2-8}          & JADE  & 5.0811E+00 & 5.1441E+00 & 5.2143E+00 & 5.1506E+00 & 3.4243E-02 & 6.6867E+00 \\
\cmidrule{2-8}          & CoDE  & 5.4858E+00 & 5.7458E+00 & 5.9051E+00 & 5.7220E+00 & 1.2025E-01 & 7.4948E+00 \\
\cmidrule{2-8}          & SHADE & 5.1742E+00 & 5.2753E+00 & 5.3571E+00 & 5.2672E+00 & 5.0745E-02 & 8.1721E+00 \\
\cmidrule{2-8}          & LSHADE & 5.0219E+00 & 5.0227E+00 & 5.2656E+00 & 5.0501E+00 & 5.6540E-02 & 8.2123E+00 \\
    \midrule
    \multirow{9}[18]{*}{3} & SA    & 1.9539E+01 & 2.1143E+01 & 2.4438E+01 & 2.1250E+01 & 1.2990E+00 & 1.7355E+01 \\
\cmidrule{2-8}          & SPSO  & 1.8433E+01 & 1.9642E+01 & 2.0614E+01 & 1.9708E+01 & 5.3792E-01 & 2.0101E+01 \\
\cmidrule{2-8}          & GA    & 1.9198E+01 & 2.0144E+01 & 2.1234E+01 & 2.0149E+01 & 5.7839E-01 & 1.7473E+01 \\
\cmidrule{2-8}          & OFA   & 1.9602E+01 & 2.0752E+01 & 2.1667E+01 & 2.0770E+01 & 5.0845E-01 & 1.9847E+01 \\
\cmidrule{2-8}          & CSO   & 1.6691E+01 & 1.7707E+01 & 1.8682E+01 & 1.7761E+01 & 5.0523E-01 & 1.4178E+01 \\
\cmidrule{2-8}          & JADE  & 1.7409E+01 & 1.7880E+01 & 1.8385E+01 & 1.7871E+01 & 2.8381E-01 & 1.7929E+01 \\
\cmidrule{2-8}          & CoDE  & 1.8320E+01 & 1.9258E+01 & 1.9684E+01 & 1.9112E+01 & 3.8079E-01 & 1.6960E+01 \\
\cmidrule{2-8}          & SHADE & 1.7304E+01 & 1.8248E+01 & 1.8757E+01 & 1.8227E+01 & 2.8792E-01 & 2.1254E+01 \\
\cmidrule{2-8}          & LSHADE & 1.6121E+01 & 1.6283E+01 & 1.6695E+01 & 1.6360E+01 & 1.6745E-01 & 2.8408E+01 \\
    \midrule
    \multirow{9}[18]{*}{4} & SA    & 2.2182E+01 & 2.2629E+01 & 2.3160E+01 & 2.2641E+01 & 2.5755E-01 & 5.6349E+00 \\
\cmidrule{2-8}          & SPSO  & 2.1027E+01 & 2.1111E+01 & 2.1228E+01 & 2.1112E+01 & 5.5062E-02 & 6.2601E+00 \\
\cmidrule{2-8}          & GA    & 2.1053E+01 & 2.1133E+01 & 2.1254E+01 & 2.1144E+01 & 5.4330E-02 & 4.6637E+00 \\
\cmidrule{2-8}          & OFA   & 2.1041E+01 & 2.1085E+01 & 2.1188E+01 & 2.1101E+01 & 4.4650E-02 & 4.3282E+00 \\
\cmidrule{2-8}          & CSO   & 2.1022E+01 & 2.1022E+01 & 2.1022E+01 & 2.1022E+01 & 1.1110E-06 & 4.0850E+00 \\
\cmidrule{2-8}          & JADE  & 2.1022E+01 & 2.1022E+01 & 2.1022E+01 & 2.1022E+01 & 1.0122E-09 & 4.1126E+00 \\
\cmidrule{2-8}          & CoDE  & 2.1022E+01 & 2.1023E+01 & 2.1023E+01 & 2.1023E+01 & 6.5874E-05 & 4.2364E+00 \\
\cmidrule{2-8}          & SHADE & 2.1022E+01 & 2.1022E+01 & 2.1022E+01 & 2.1022E+01 & 1.3467E-09 & 4.8679E+00 \\
\cmidrule{2-8}          & LSHADE & 2.1022E+01 & 2.1022E+01 & 2.1029E+01 & 2.1023E+01 & 1.2089E-03 & 6.1771E+00 \\
    \bottomrule
    \end{tabular}%
  \label{table2_1}%
\end{table}%

\begin{table}[htbp]
\tiny
  \centering
  \caption{Summary statistics for $D$-optimality criterion values of designs generated by different algorithms for 5-8 models.}
    \begin{tabular}{clcccccc}
    \toprule
    Problem & {Algorithm} & {best} & {median} & {worst} & {mean} & {std} & {time} \\
    \midrule
    \multirow{9}[18]{*}{5} & SA    & 1.9014E+01 & 1.9829E+01 & 2.0649E+01 & 1.9837E+01 & 4.4351E-01 & 1.0149E+01 \\
\cmidrule{2-8}          & SPSO  & 1.8559E+01 & 1.8834E+01 & 1.9201E+01 & 1.8839E+01 & 1.6180E-01 & 1.0475E+01 \\
\cmidrule{2-8}          & GA    & 1.8521E+01 & 1.8759E+01 & 1.8915E+01 & 1.8744E+01 & 1.0214E-01 & 8.4169E+00 \\
\cmidrule{2-8}          & OFA   & 1.8715E+01 & 1.9061E+01 & 1.9503E+01 & 1.9085E+01 & 2.1220E-01 & 9.3761E+00 \\
\cmidrule{2-8}          & CSO   & 1.8328E+01 & 1.8335E+01 & 1.8388E+01 & 1.8341E+01 & 1.5112E-02 & 5.8712E+00 \\
\cmidrule{2-8}          & JADE  & 1.8340E+01 & 1.8351E+01 & 1.8369E+01 & 1.8352E+01 & 8.6070E-03 & 8.3880E+00 \\
\cmidrule{2-8}          & CoDE  & 1.8563E+01 & 1.8660E+01 & 1.8779E+01 & 1.8664E+01 & 6.0923E-02 & 9.1251E+00 \\
\cmidrule{2-8}          & SHADE & 1.8338E+01 & 1.8371E+01 & 1.8412E+01 & 1.8373E+01 & 1.8124E-02 & 9.6797E+00 \\
\cmidrule{2-8}          & LSHADE & 1.8328E+01 & 1.8328E+01 & 1.8330E+01 & 1.8328E+01 & 4.3515E-04 & 9.0270E+00 \\
    \midrule
    \multirow{9}[18]{*}{6} & SA    & 5.2563E+00 & 5.3232E+00 & 5.6250E+00 & 5.3571E+00 & 1.1280E-01 & 4.6965E+00 \\
\cmidrule{2-8}          & SPSO  & 5.2529E+00 & 5.2556E+00 & 5.2700E+00 & 5.2566E+00 & 4.1190E-03 & 5.0812E+00 \\
\cmidrule{2-8}          & GA    & 5.2563E+00 & 5.2618E+00 & 5.2690E+00 & 5.2616E+00 & 4.0658E-03 & 3.7057E+00 \\
\cmidrule{2-8}          & OFA   & 5.2532E+00 & 5.2568E+00 & 5.2734E+00 & 5.2579E+00 & 4.8636E-03 & 3.4635E+00 \\
\cmidrule{2-8}          & CSO   & 5.2528E+00 & 5.2528E+00 & 5.2528E+00 & 5.2528E+00 & 3.1869E-15 & 3.0429E+00 \\
\cmidrule{2-8}          & JADE  & 5.2528E+00 & 5.2528E+00 & 5.2528E+00 & 5.2528E+00 & 4.9944E-11 & 3.4204E+00 \\
\cmidrule{2-8}          & CoDE  & 5.2528E+00 & 5.2529E+00 & 5.2543E+00 & 5.2530E+00 & 3.2672E-04 & 3.6795E+00 \\
\cmidrule{2-8}          & SHADE & 5.2528E+00 & 5.2528E+00 & 5.2528E+00 & 5.2528E+00 & 4.0355E-12 & 4.2826E+00 \\
\cmidrule{2-8}          & LSHADE & 5.2528E+00 & 5.2528E+00 & 5.2529E+00 & 5.2528E+00 & 1.1133E-05 & 4.9213E+00 \\
    \midrule
    \multirow{9}[18]{*}{7} & SA    & 2.5206E+01 & 2.6849E+01 & 2.9695E+01 & 2.7104E+01 & 1.3432E+00 & 6.4176E+00 \\
\cmidrule{2-8}          & SPSO  & 2.5987E+01 & 2.8507E+01 & 2.9983E+01 & 2.8335E+01 & 1.0567E+00 & 5.6918E+00 \\
\cmidrule{2-8}          & GA    & 2.5222E+01 & 2.5731E+01 & 2.6698E+01 & 2.5743E+01 & 3.2034E-01 & 4.6229E+00 \\
\cmidrule{2-8}          & OFA   & 2.5584E+01 & 2.6437E+01 & 2.7540E+01 & 2.6424E+01 & 4.8131E-01 & 4.1098E+00 \\
\cmidrule{2-8}          & CSO   & 2.4755E+01 & 2.4817E+01 & 2.5076E+01 & 2.4840E+01 & 9.0283E-02 & 4.4402E+00 \\
\cmidrule{2-8}          & JADE  & 2.4755E+01 & 2.4764E+01 & 2.4786E+01 & 2.4765E+01 & 6.8432E-03 & 4.8036E+00 \\
\cmidrule{2-8}          & CoDE  & 2.4970E+01 & 2.5145E+01 & 2.5401E+01 & 2.5164E+01 & 1.1437E-01 & 4.8842E+00 \\
\cmidrule{2-8}          & SHADE & 2.4764E+01 & 2.4773E+01 & 2.4812E+01 & 2.4778E+01 & 1.4106E-02 & 5.5157E+00 \\
\cmidrule{2-8}          & LSHADE & 2.4752E+01 & 2.4752E+01 & 2.4946E+01 & 2.4760E+01 & 3.8811E-02 & 7.3086E+00 \\
    \midrule
    \multirow{9}[18]{*}{8} & SA    & 1.7440E+01 & 1.9019E+01 & 1.9510E+01 & 1.9084E+01 & 4.0072E-01 & 8.0268E+02 \\
\cmidrule{2-8}          & SPSO  & 1.3380E+01 & 1.4211E+01 & 1.4912E+01 & 1.4213E+01 & 3.5979E-01 & 7.8987E+02 \\
\cmidrule{2-8}          & GA    & 1.9646E+01 & 2.1166E+01 & 2.2210E+01 & 2.0989E+01 & 7.1101E-01 & 6.2538E+02 \\
\cmidrule{2-8}          & OFA   & 1.3687E+01 & 1.4605E+01 & 1.5296E+01 & 1.4534E+01 & 4.5288E-01 & 5.4602E+02 \\
\cmidrule{2-8}          & CSO   & 1.3079E+01 & 1.5750E+01 & 1.7425E+01 & 1.5572E+01 & 1.0201E+00 & 6.2543E+02 \\
\cmidrule{2-8}          & JADE  & 1.0123E+01 & 1.0137E+01 & 1.0188E+01 & 1.0140E+01 & 1.4061E-02 & 7.3665E+02 \\
\cmidrule{2-8}          & CoDE  & 1.1238E+01 & 1.1481E+01 & 1.1660E+01 & 1.1477E+01 & 1.0074E-01 & 6.7893E+02 \\
\cmidrule{2-8}          & SHADE & 1.0120E+01 & 1.0132E+01 & 1.0151E+01 & 1.0131E+01 & 7.2142E-03 & 8.3684E+02 \\
\cmidrule{2-8}          & LSHADE & 1.0141E+01 & 1.0355E+01 & 1.7984E+01 & 1.1012E+01 & 1.6465E+00 & 8.5980E+02 \\
    \bottomrule
    \end{tabular}%
  \label{table2_2}%
\end{table}%

\begin{table}[htbp]
\tiny
  \centering
  \caption{Summary statistics for $D$-optimality criterion values of designs generated by different algorithms for 9-12 models.}
    \begin{tabular}{clcccccc}
    \toprule
    Problem & {Algorithm} & {best} & {median} & {worst} & {mean} & {std} & {time} \\
    \midrule
    \multirow{9}[18]{*}{9} & SA    & 2.1138E+00 & 3.5850E+00 & 5.5857E+00 & 3.6311E+00 & 8.0692E-01 & 1.7705E+03 \\
\cmidrule{2-8}          & SPSO  & 1.9491E+00 & 2.2749E+00 & 2.7721E+00 & 2.2755E+00 & 1.6800E-01 & 2.1851E+03 \\
\cmidrule{2-8}          & GA    & 2.3355E+00 & 2.7209E+00 & 2.8683E+00 & 2.6935E+00 & 1.4640E-01 & 2.0297E+03 \\
\cmidrule{2-8}          & OFA   & -4.7730E-01 & -1.4708E-01 & 6.3570E-01 & -4.7291E-02 & 3.2590E-01 & 1.4896E+03 \\
\cmidrule{2-8}          & CSO   & -1.0171E+00 & -5.0889E-01 & 1.0382E-01 & -4.5812E-01 & 3.2599E-01 & 1.5052E+03 \\
\cmidrule{2-8}          & JADE  & -1.4051E+00 & -1.3942E+00 & -1.3722E+00 & -1.3935E+00 & 7.2466E-03 & 1.6989E+03 \\
\cmidrule{2-8}          & CoDE  & -9.5511E-01 & -6.2197E-01 & -3.3163E-01 & -5.9983E-01 & 1.4441E-01 & 1.5607E+03 \\
\cmidrule{2-8}          & SHADE & -1.4058E+00 & -1.3957E+00 & -1.3645E+00 & -1.3946E+00 & 9.4775E-03 & 1.2752E+03 \\
\cmidrule{2-8}          & LSHADE & -1.4099E+00 & -1.3803E+00 & -1.3335E+00 & -1.3788E+00 & 1.7491E-02 & 1.4326E+03 \\
    \midrule
    \multirow{9}[18]{*}{10} & SA    & 9.6203E+00 & 1.1093E+01 & 1.3250E+01 & 1.1146E+01 & 1.0032E+00 & 8.1256E+02 \\
\cmidrule{2-8}          & SPSO  & 6.8778E+00 & 7.3106E+00 & 7.6479E+00 & 7.3302E+00 & 1.8711E-01 & 1.2783E+03 \\
\cmidrule{2-8}          & GA    & 7.3239E+00 & 7.7247E+00 & 7.9990E+00 & 7.6912E+00 & 1.6754E-01 & 1.1488E+03 \\
\cmidrule{2-8}          & OFA   & 5.3458E+00 & 6.4498E+00 & 6.7960E+00 & 6.3674E+00 & 3.9753E-01 & 7.9646E+02 \\
\cmidrule{2-8}          & CSO   & 4.1360E+00 & 4.6397E+00 & 5.2884E+00 & 4.6058E+00 & 3.2267E-01 & 6.5452E+02 \\
\cmidrule{2-8}          & JADE  & 3.7108E+00 & 3.7162E+00 & 3.7363E+00 & 3.7198E+00 & 8.9034E-03 & 8.5128E+02 \\
\cmidrule{2-8}          & CoDE  & 4.3516E+00 & 4.5441E+00 & 4.7225E+00 & 4.5422E+00 & 8.5509E-02 & 7.9899E+02 \\
\cmidrule{2-8}          & SHADE & 3.7092E+00 & 3.7161E+00 & 3.7288E+00 & 3.7173E+00 & 4.3694E-03 & 8.3846E+02 \\
\cmidrule{2-8}          & LSHADE & 3.7087E+00 & 3.7354E+00 & 3.9989E+00 & 3.7526E+00 & 5.6766E-02 & 8.9841E+02 \\
    \midrule
    \multirow{9}[18]{*}{11} & SA    & -1.9743E+00 & -1.8435E+00 & 1.2587E+01 & 3.3043E+00 & 7.1055E+00 & 1.0568E+03 \\
\cmidrule{2-8}          & SPSO  & -5.1770E+00 & -3.5075E+00 & -1.5682E+00 & -3.5155E+00 & 9.5282E-01 & 1.1954E+03 \\
\cmidrule{2-8}          & GA    & -4.9769E+00 & -1.1644E+00 & 7.3234E-01 & -1.2660E+00 & 1.2505E+00 & 1.0956E+03 \\
\cmidrule{2-8}          & OFA   & -7.2037E+00 & -6.7809E+00 & -5.9558E+00 & -6.7419E+00 & 3.0416E-01 & 9.3847E+02 \\
\cmidrule{2-8}          & CSO   & -7.1550E+00 & -6.2678E+00 & -5.6565E+00 & -6.3541E+00 & 4.6642E-01 & 1.0135E+03 \\
\cmidrule{2-8}          & JADE  & -8.6005E+00 & -8.6003E+00 & -8.6001E+00 & -8.6003E+00 & 1.1716E-04 & 1.0640E+03 \\
\cmidrule{2-8}          & CoDE  & -7.9452E+00 & -7.8098E+00 & -7.6923E+00 & -7.8139E+00 & 7.0938E-02 & 8.8282E+02 \\
\cmidrule{2-8}          & SHADE & -8.5980E+00 & -8.5944E+00 & -8.5902E+00 & -8.5945E+00 & 2.0392E-03 & 1.0306E+03 \\
\cmidrule{2-8}          & LSHADE & -8.5987E+00 & -8.5795E+00 & -5.3369E+00 & -8.0841E+00 & 8.8577E-01 & 1.0936E+03 \\
    \midrule
    \multirow{9}[18]{*}{12} & SA    & 9.1286E+01 & 1.0172E+02 & 1.1639E+02 & 1.0205E+02 & 6.2509E+00 & 1.0663E+03 \\
\cmidrule{2-8}          & SPSO  & 7.0244E+01 & 7.3279E+01 & 7.5312E+01 & 7.3122E+01 & 1.2190E+00 & 1.2432E+03 \\
\cmidrule{2-8}          & GA    & 6.9499E+01 & 7.3638E+01 & 7.6067E+01 & 7.3484E+01 & 1.6996E+00 & 1.0153E+03 \\
\cmidrule{2-8}          & OFA   & 6.0067E+01 & 6.4620E+01 & 6.6220E+01 & 6.4273E+01 & 1.6884E+00 & 8.4061E+02 \\
\cmidrule{2-8}          & CSO   & 4.9104E+01 & 5.6595E+01 & 6.7650E+01 & 5.7496E+01 & 4.4471E+00 & 8.6097E+02 \\
\cmidrule{2-8}          & JADE  & 3.5215E+01 & 3.5923E+01 & 3.6730E+01 & 3.5914E+01 & 3.2457E-01 & 1.0156E+03 \\
\cmidrule{2-8}          & CoDE  & 3.9260E+01 & 4.4461E+01 & 4.8282E+01 & 4.4077E+01 & 2.5368E+00 & 1.0337E+03 \\
\cmidrule{2-8}          & SHADE & 3.4703E+01 & 3.5327E+01 & 3.6080E+01 & 3.5334E+01 & 3.0471E-01 & 9.8981E+02 \\
\cmidrule{2-8}          & LSHADE & 3.3481E+01 & 3.4330E+01 & 4.8622E+01 & 3.4906E+01 & 2.8913E+00 & 9.9196E+02 \\
    \bottomrule
    \end{tabular}%
  \label{table2_3}%
\end{table}%

In Table \ref{table3}, the Wilcoxon rank test was applied to test equality of the $D$-optimal design criterion values from various algorithms at the 0.05 significance level. The notations `-', `+', and `=' indicate that the corresponding comparative algorithm in the column of table is significantly worse than, better than, or similar to the criterion value of the target algorithm shown in the row of the table, respectively. Table \ref{table3} shows that LSHADE outperforms  other comparative algorithms   and the OFA and the traditional evolutionary algorithms(SA, SPSO, and GA) are inferior to DE variants such as JADE, CoDE, SHADE, and LSHADE. CSO is also a competitive algorithm better than CoDE, OFA, and the traditional EAs, but inferior to JADE, SHADE, and LSHADE.

\begin{table}[htbp]
\tiny
  \centering
  \caption{Results of the Wilcoxon rank tests for the equality of the medians of  values of the $D$-optimality criterion values of designs  generated from various  algorithms at the 0.05 significance level.}
    \begin{tabular}{cccccccccc}
    \toprule
    -/+/= & SA    & SPSO  & GA    & OFA   & CSO   & JADE  & CoDE  & SHADE & LSHADE \\
    \midrule
    SA    & [0/0/12] & [10/1/1] & [9/2/1] & [9/0/3] & [12/0/0] & [12/0/0] & [12/0/0] & [12/0/0] & [12/0/0] \\
    \midrule
    SPSO  & [1/10/1] & [0/0/12] & [2/7/3] & [5/3/4] & [11/1/0] & [12/0/0] & [12/0/0] & [12/0/0] & [12/0/0] \\
    \midrule
    GA    & [2/9/1] & [7/2/3] & [0/0/12] & [8/4/0] & [12/0/0] & [12/0/0] & [12/0/0] & [12/0/0] & [12/0/0] \\
    \midrule
    OFA   & [0/9/3] & [3/5/4] & [4/8/0] & [0/0/12] & [10/2/0] & [12/0/0] & [12/0/0] & [12/0/0] & [12/0/0] \\
    \midrule
    CSO   & [0/12/0] & [1/11/0] & [0/12/0] & [2/10/0] & [0/0/12] & [8/3/1] & [3/7/2] & [7/5/0] & [10/2/0] \\
    \midrule
    JADE  & [0/12/0] & [0/12/0] & [0/12/0] & [0/12/0] & [3/8/1] & [0/0/12] & [0/12/0] & [2/6/4] & [7/5/0] \\
    \midrule
    CoDE  & [0/12/0] & [0/12/0] & [0/12/0] & [0/12/0] & [7/3/2] & [12/0/0] & [0/0/12] & [12/0/0] & [12/0/0] \\
    \midrule
    SHADE & [0/12/0] & [0/12/0] & [0/12/0] & [0/12/0] & [5/7/0] & [6/2/4] & [0/12/0] & [0/0/12] & [7/4/1] \\
    \midrule
    LSHADE & [0/12/0] & [0/12/0] & [0/12/0] & [0/12/0] & [2/10/0] & [5/7/0] & [0/12/0] & [4/7/1] & [0/0/12] \\
    \bottomrule
    \end{tabular}%
  \label{table3}%
\end{table}%

\begin{figure}[htp]
\centering
	{\includegraphics[width=0.32\linewidth]{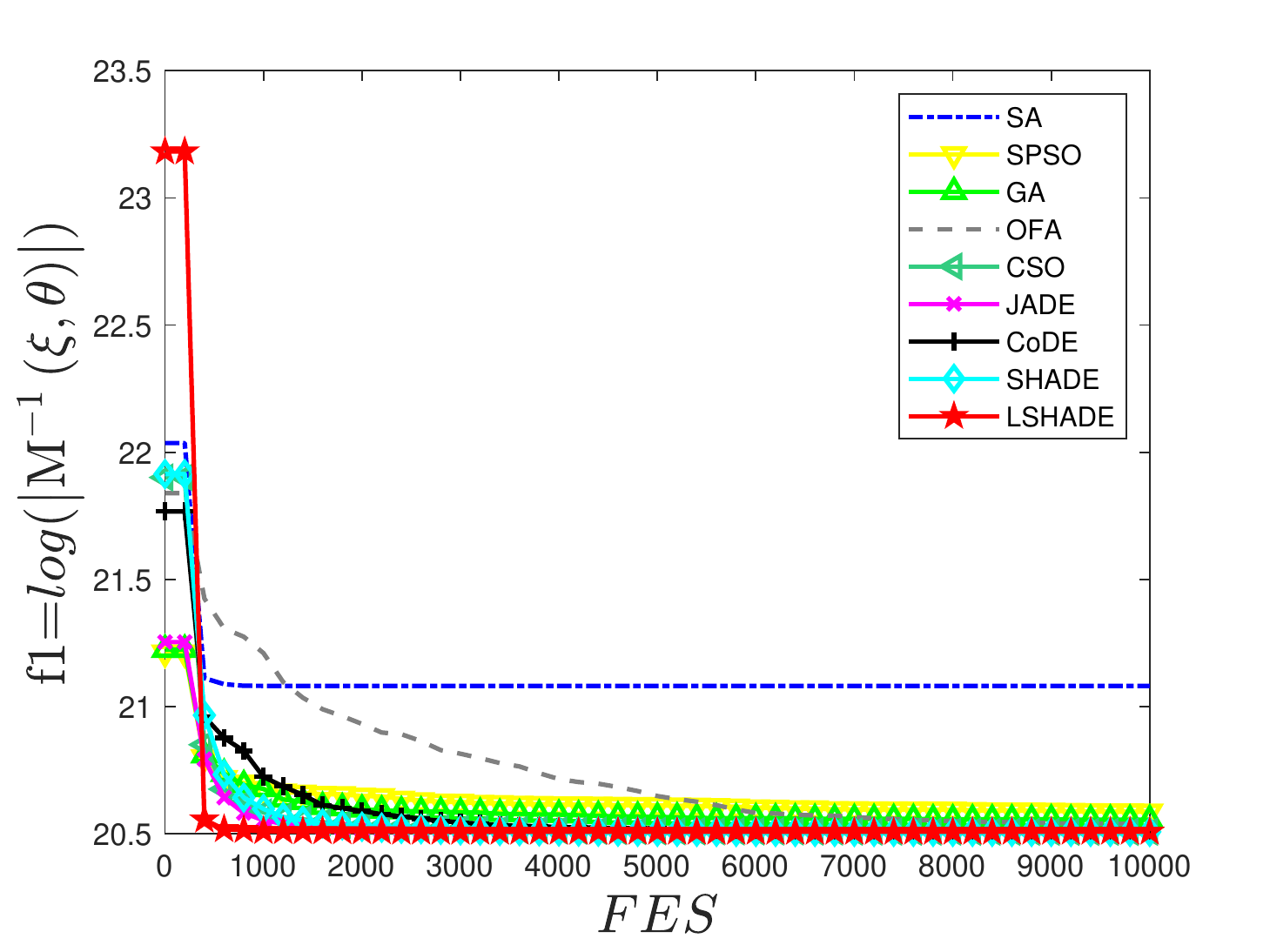}}
    \label{figure1D}\hfill
	{\includegraphics[width=0.32\linewidth]{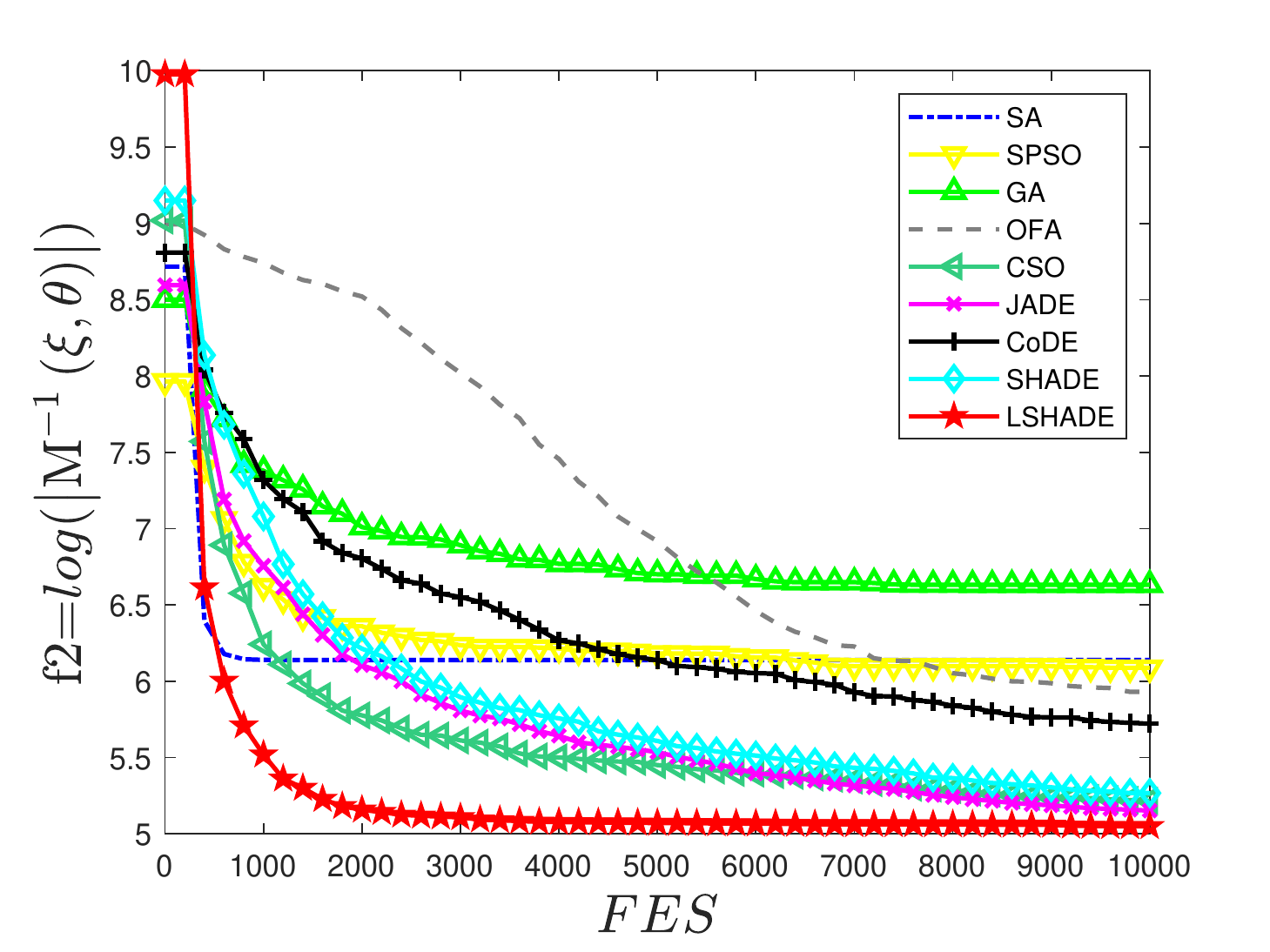}}
    \label{figure2D}\hfill
    {\includegraphics[width=0.32\linewidth]{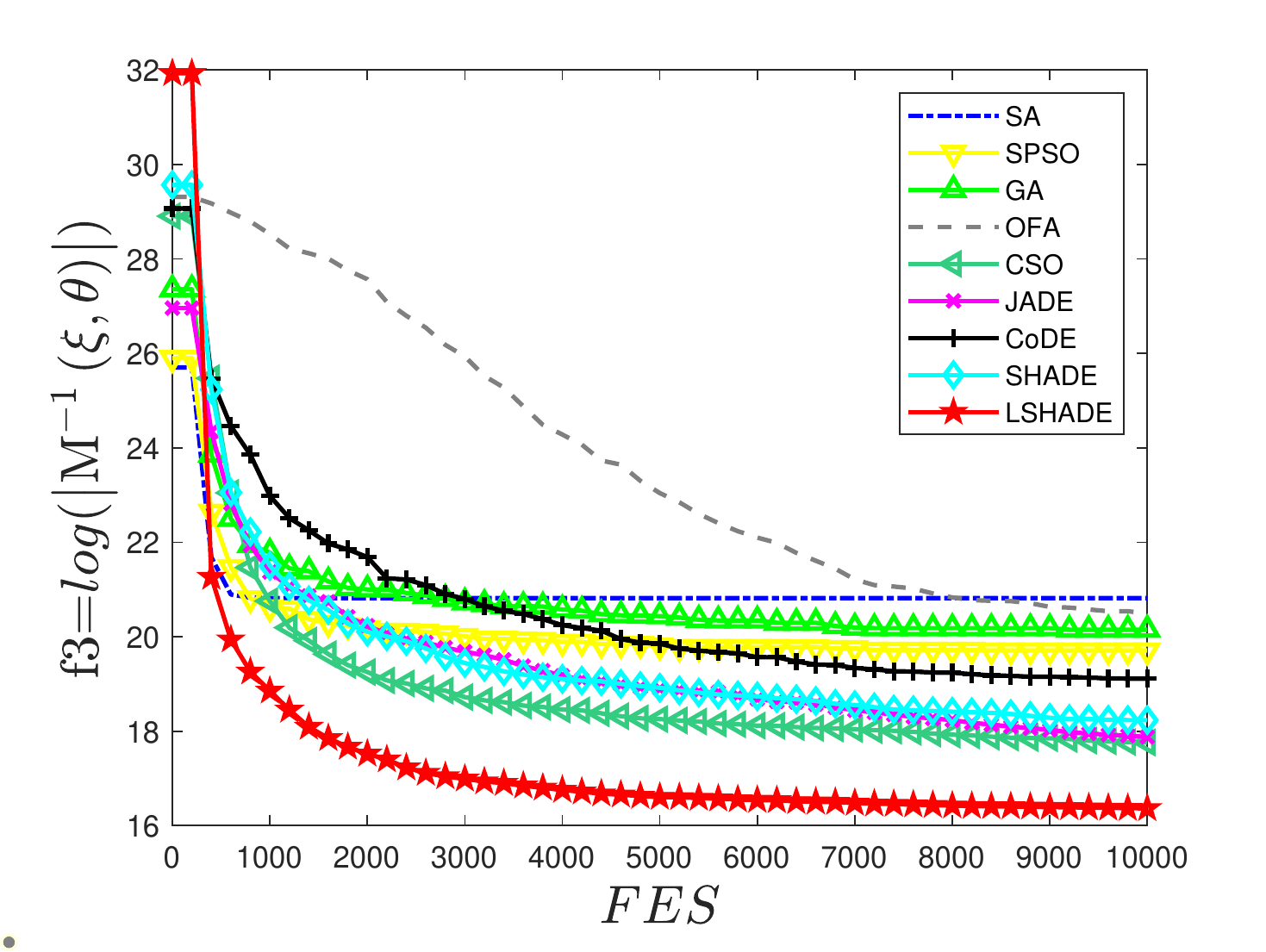}}
    \label{figure3D}\\
    {\includegraphics[width=0.32\linewidth]{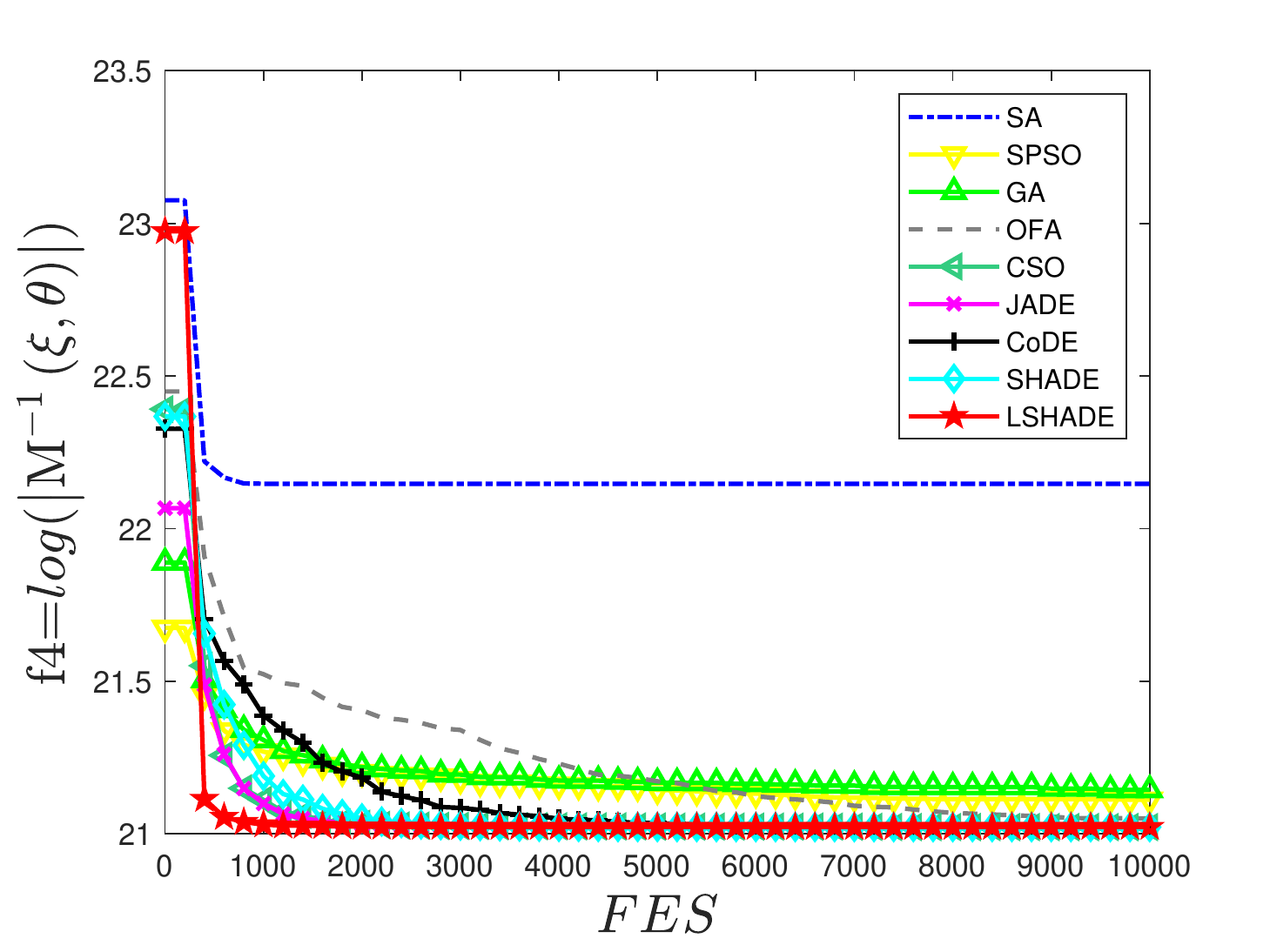}}
    \label{figure4D}\hfill
    {\includegraphics[width=0.32\linewidth]{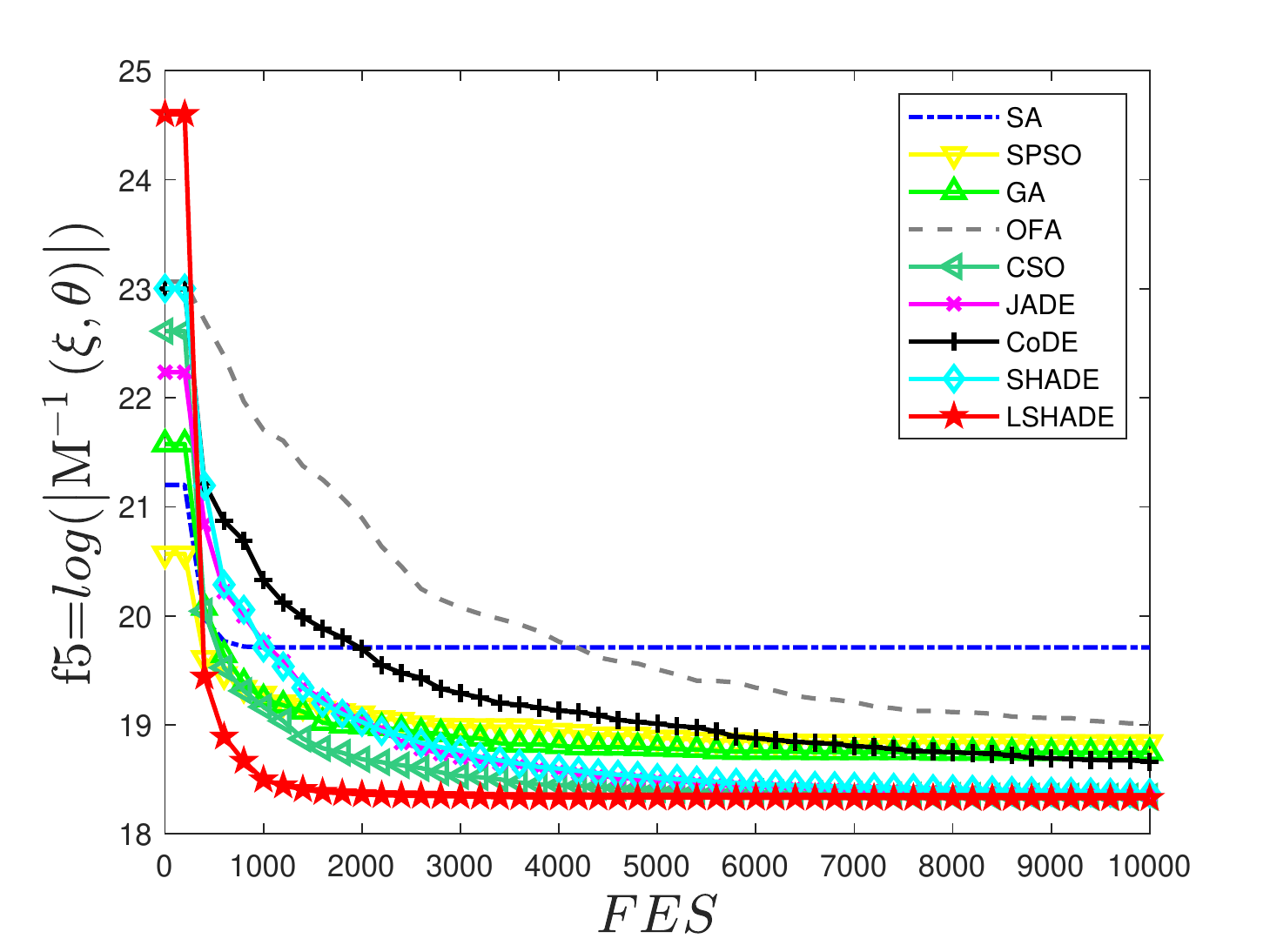}}
    \label{figure5D}\hfill
    {\includegraphics[width=0.32\linewidth]{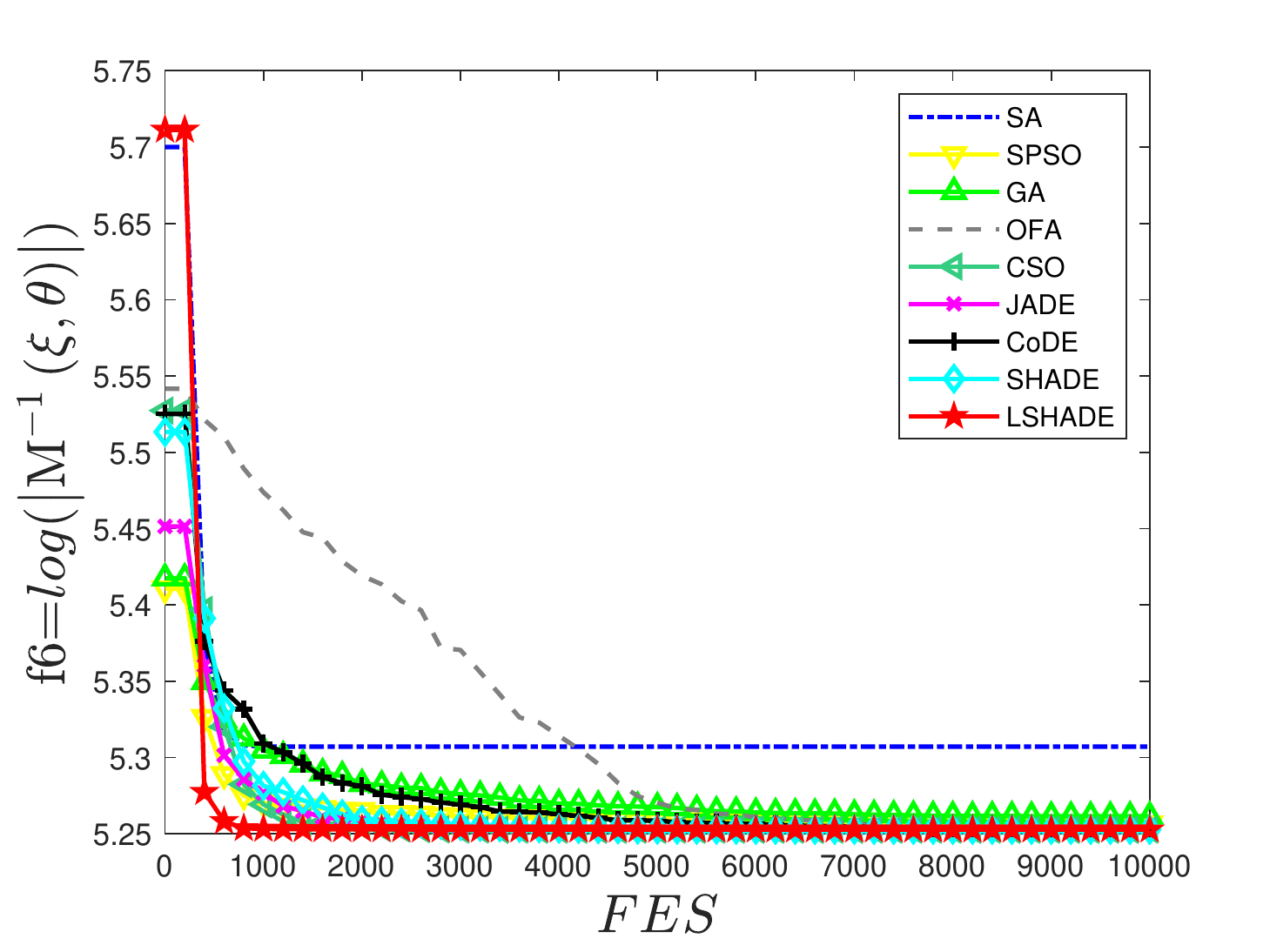}}
    \label{figure6D}\\
    {\includegraphics[width=0.32\linewidth]{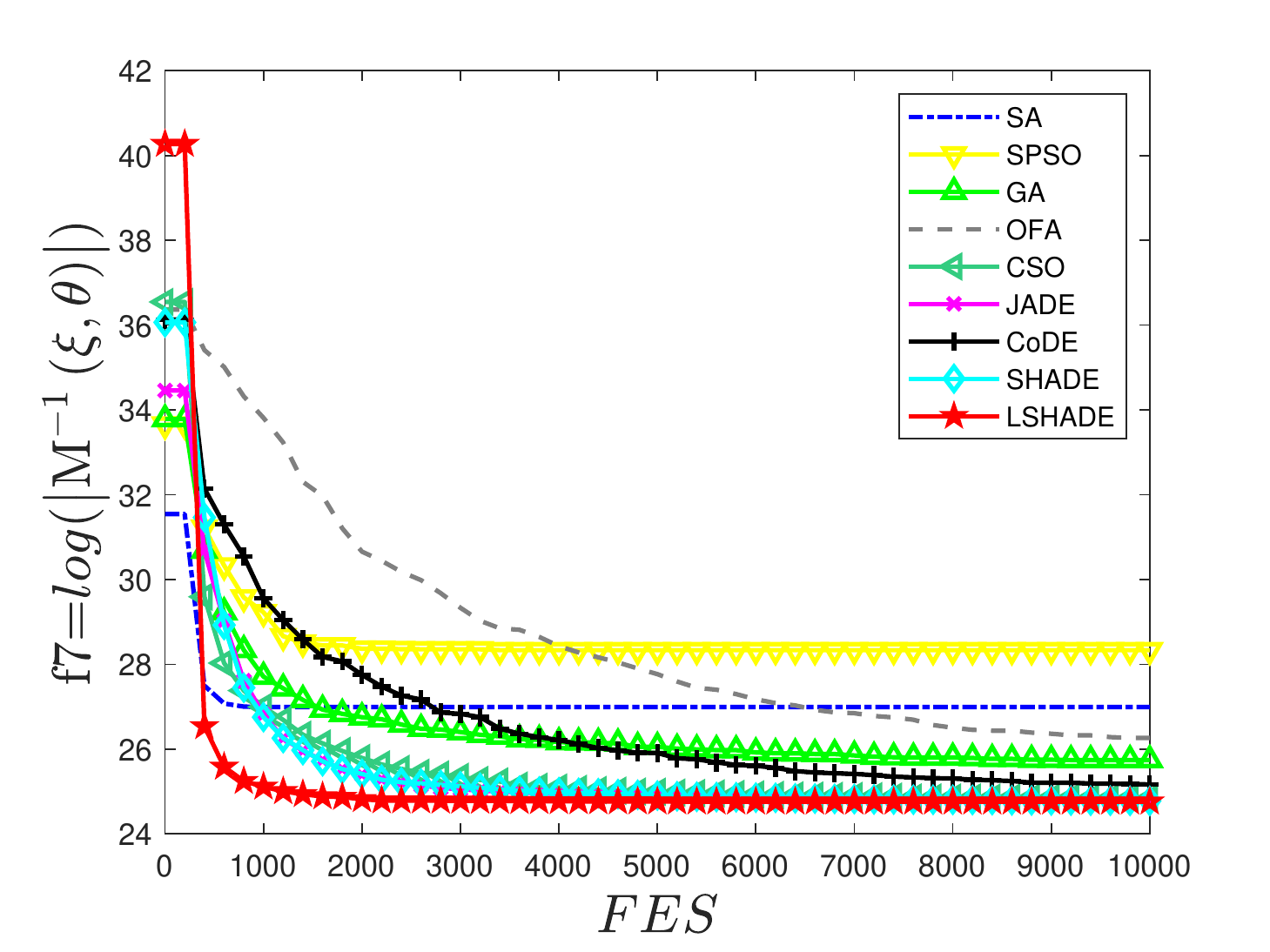}}
    \label{figure7D}\hfill
	{\includegraphics[width=0.32\linewidth]{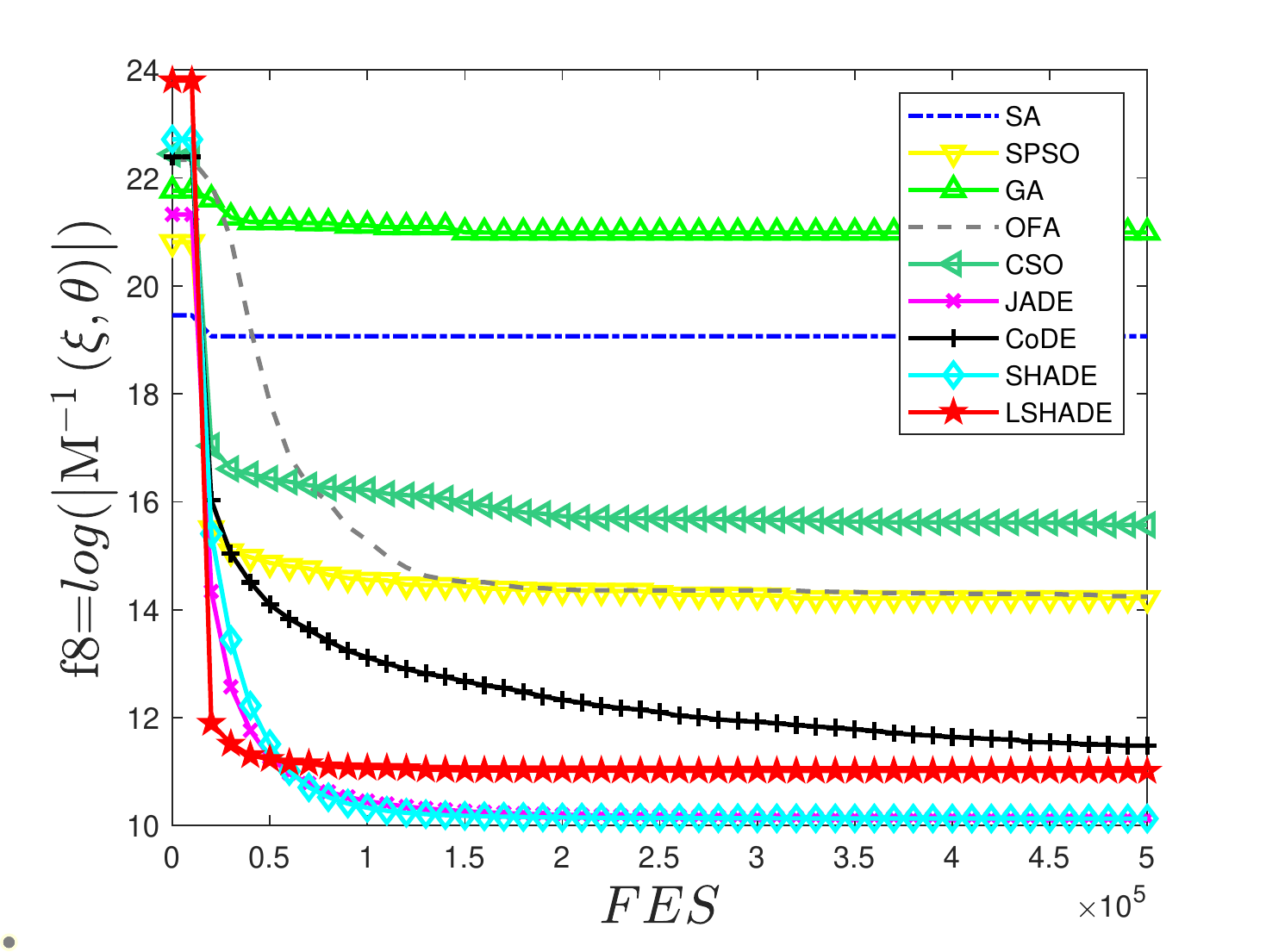}}
    \label{figure8D}\hfill
    {\includegraphics[width=0.32\linewidth]{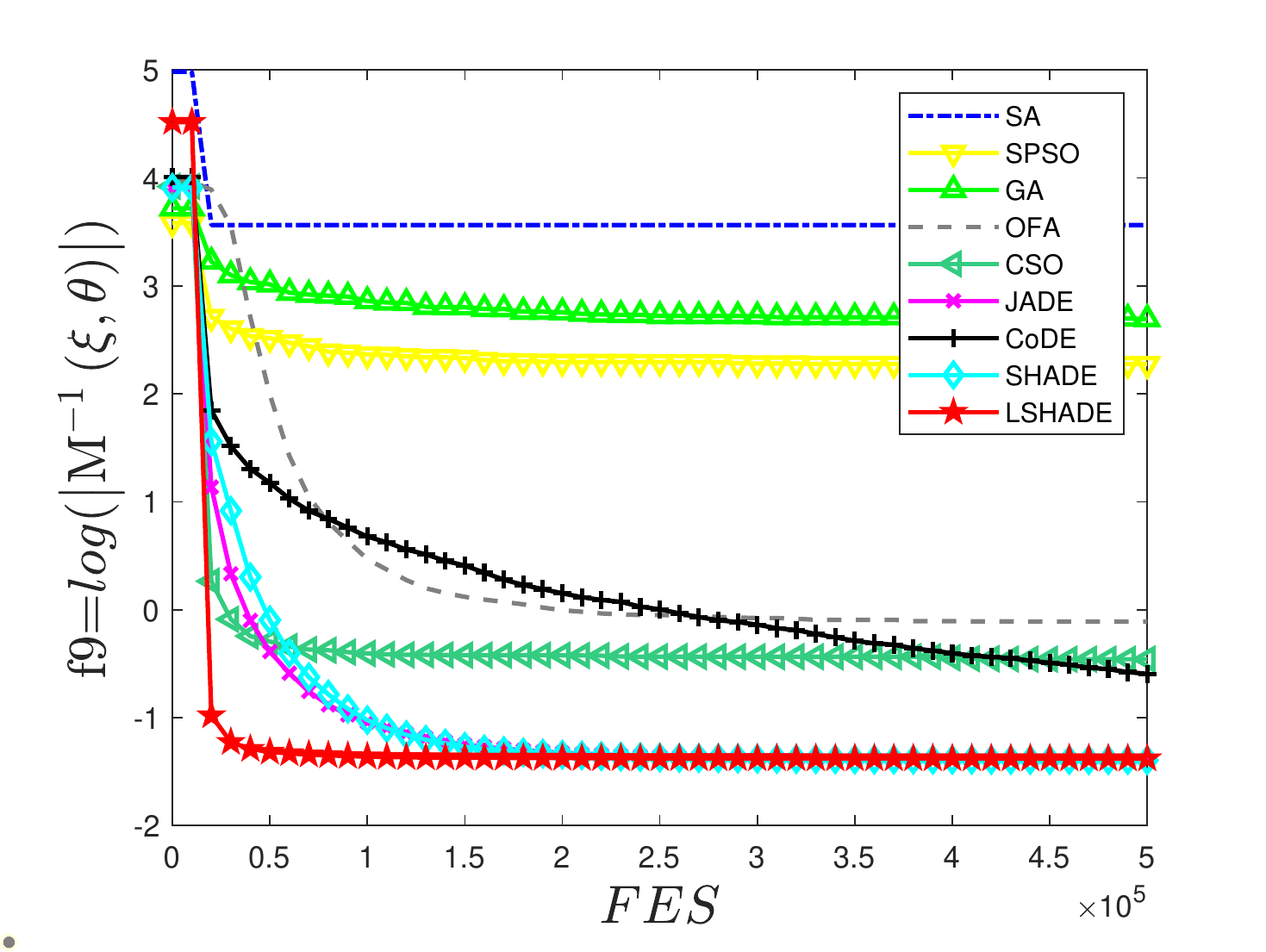}}
    \label{figure9D}\\
    {\includegraphics[width=0.32\linewidth]{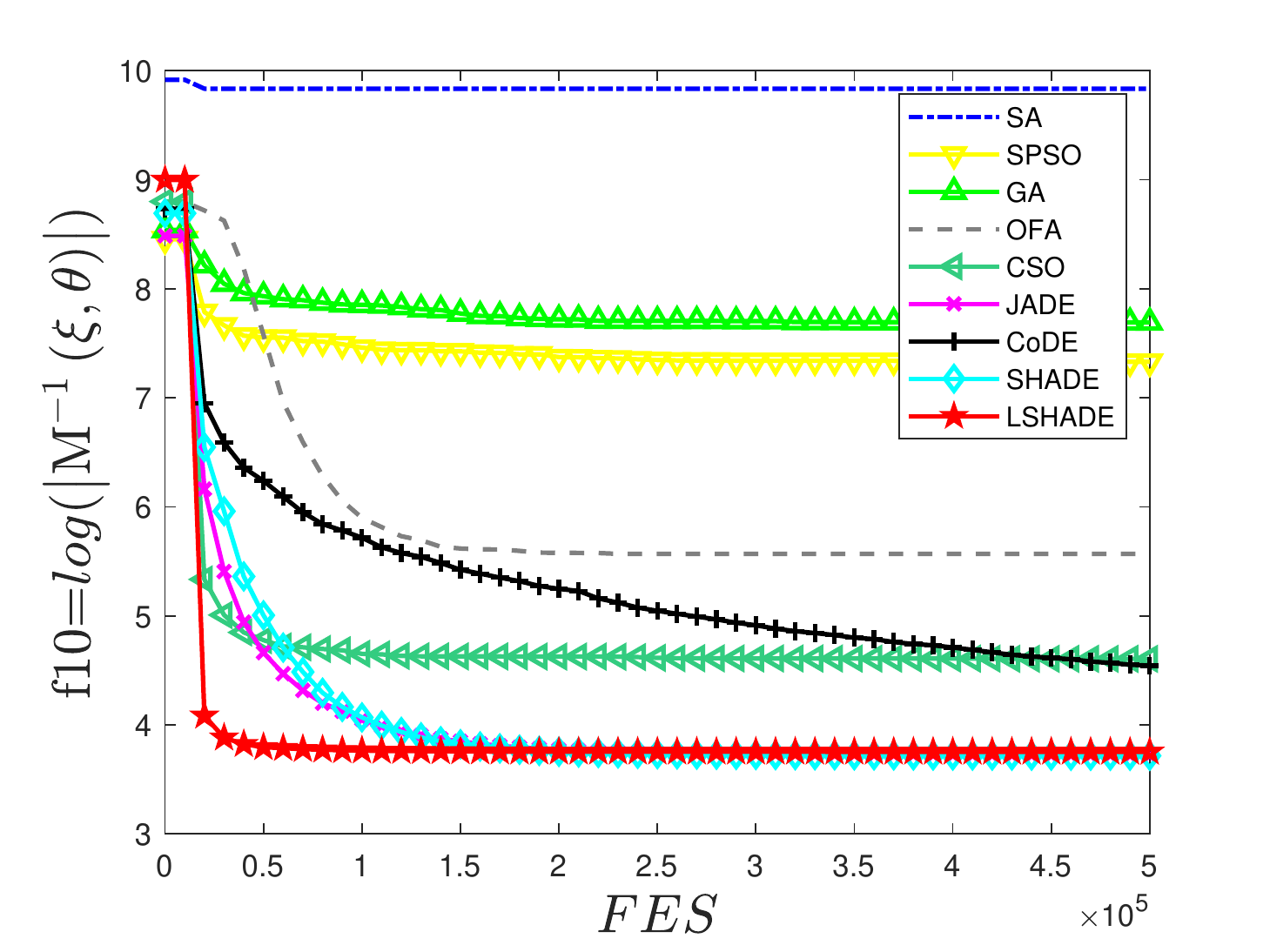}}
    \label{figure10D}\hfill
    {\includegraphics[width=0.32\linewidth]{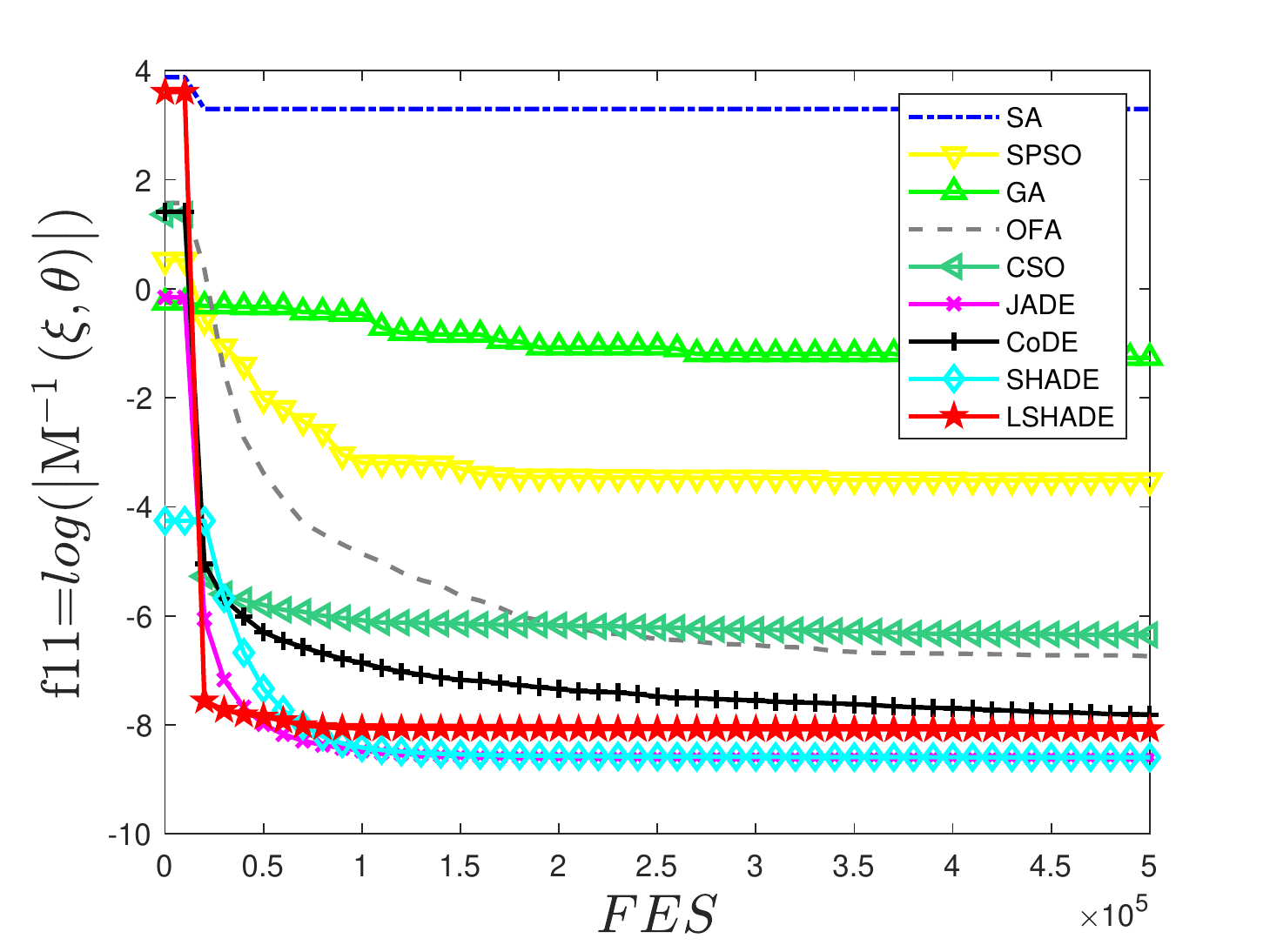}}
    \label{figure11D}\hfill
    {\includegraphics[width=0.32\linewidth]{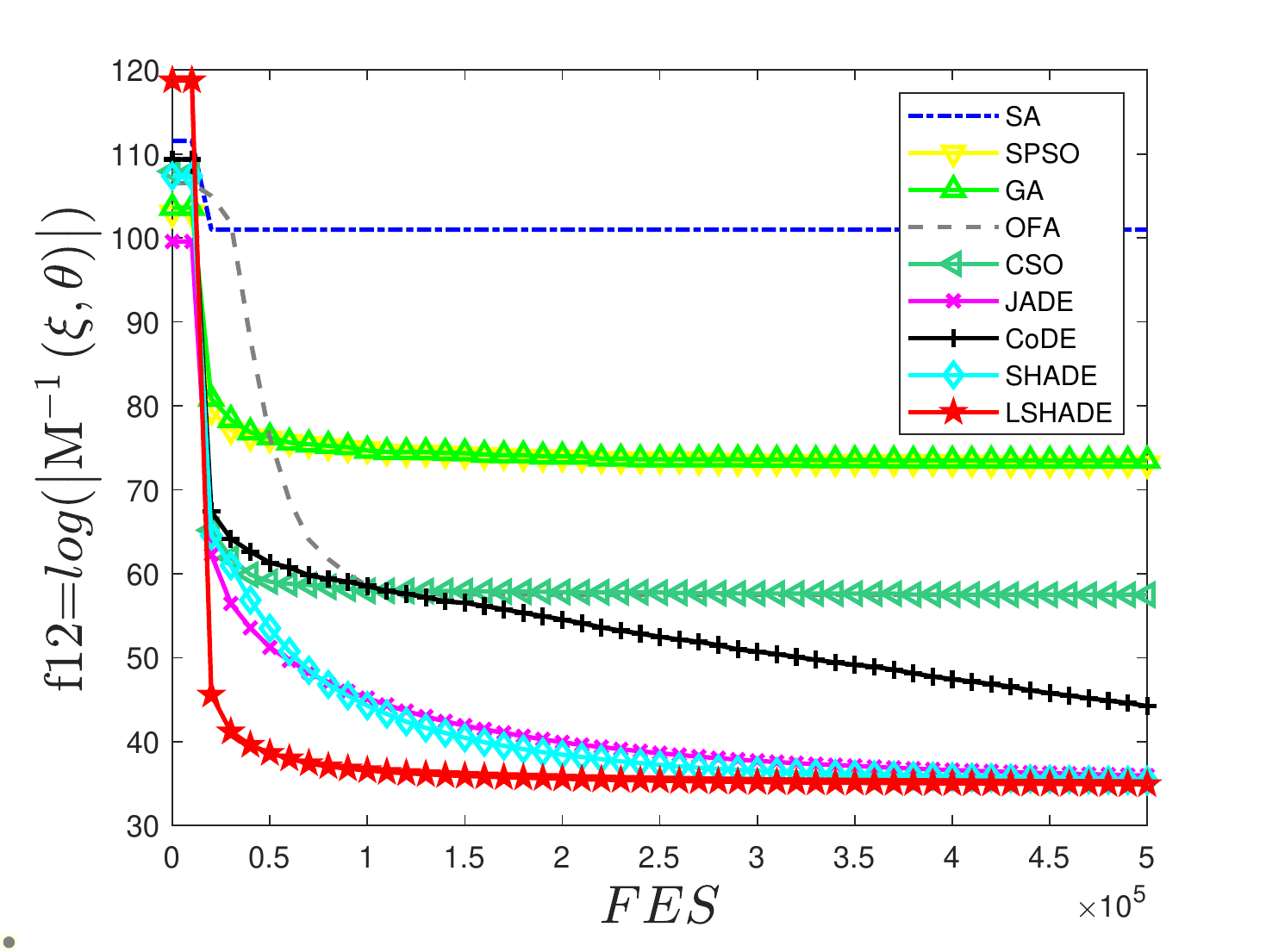}}
    \label{figure12D}\\
	\caption{Mean Performance of the SA, SPSO, GA, OFA, CSO, JADE, CoDE, SHADE, and LSHADE algorithms for finding $D$-optimal designs.}
	\label{figure_Dconvergence}
\end{figure}

Figure \ref{figure_Dconvergence} displays the convergence rate of all algorithms for finding the $D$-optimal designs for the statistical model. The X axis represents the number of function evaluations and the Y axis represents the mean $D$-optimal design criterion values on the log scale. The LSHADE algorithm shows better convergence performance than other comparative algorithms on the Problem 1-7, 9-10 and 12. JADE and SHADE obtain the convergence on the problem 8 and 11. Overall DE variants exhibit better convergence ability.

\newpage
\subsection{Designs found by various algorithms under the $A$-optimality criterion}
Tables \ref{tableA_1}, \ref{tableA_2}, and \ref{tableA_3} show the summary statistics for $A$-optimality criterion values of designs generated by different algorithms for models 1-12, such as the best, median, worst, and median value, standard deviation, and Computation time. From tables, we can observe that LSHADE can obtain better performance on the Problems 2 , 3, 5, 7, 9, and 12 in terms of the median values. On the problems 8, 10 and 11, SHADE achieves better result in terms of the median values. Except for the Problems 8 and 10, LSHADE obtains the best performance in terms of the best values. Compared with OFA and the traditional EAs (SA, SPSO, and GA), DE variants obtain superior results.

\begin{table}[htbp]
\tiny
  \centering
  \caption{Summary statistics for $A$-optimality criterion values of designs generated by different algorithms for models 1-4.}
    \begin{tabular}{cccccccc}
    \toprule
    Problem & Algorithm & best  & median & worst & mean  & std   &  time \\
    \midrule
    \multirow{9}[18]{*}{1} & SA    & 5.4106E+04 & 5.5325E+04 & 6.0497E+04 & 5.5603E+04 & 1.3189E+03 & 5.3486E+00 \\
\cmidrule{2-8}          & SPSO  & 5.3870E+04 & 5.5045E+04 & 5.9888E+04 & 5.5807E+04 & 1.8094E+03 & 5.5440E+00 \\
\cmidrule{2-8}          & GA    & 5.5324E+04 & 5.7162E+04 & 6.0459E+04 & 5.7363E+04 & 1.1872E+03 & 4.6051E+00 \\
\cmidrule{2-8}          & OFA   & 5.4040E+04 & 5.4604E+04 & 5.5677E+04 & 5.4686E+04 & 4.5277E+02 & 3.5254E+00 \\
\cmidrule{2-8}          & CSO   & 5.3797E+04 & 5.3797E+04 & 5.3800E+04 & 5.3797E+04 & 6.2337E-01 & 3.7374E+00 \\
\cmidrule{2-8}          & JADE  & 5.3797E+04 & 5.3797E+04 & 5.3797E+04 & 5.3797E+04 & 4.9717E-04 & 3.8483E+00 \\
\cmidrule{2-8}          & CoDE  & 5.3801E+04 & 5.3817E+04 & 5.4021E+04 & 5.3833E+04 & 4.7297E+01 & 4.4170E+00 \\
\cmidrule{2-8}          & SHADE & 5.3797E+04 & 5.3797E+04 & 5.3797E+04 & 5.3797E+04 & 2.1238E-04 & 6.0526E+00 \\
\cmidrule{2-8}          & LSHADE & 5.3797E+04 & 5.3797E+04 & 5.3806E+04 & 5.3797E+04 & 1.8110E+00 & 8.2115E+00 \\
    \midrule
    \multirow{9}[18]{*}{2} & SA    & 2.2132E+01 & 2.4535E+01 & 2.5309E+01 & 2.4355E+01 & 7.6792E-01 & 6.2339E+00 \\
\cmidrule{2-8}          & SPSO  & 2.4813E+01 & 2.9798E+01 & 3.2309E+01 & 2.9196E+01 & 2.1380E+00 & 8.5148E+00 \\
\cmidrule{2-8}          & GA    & 2.9523E+01 & 3.5465E+01 & 3.9113E+01 & 3.5036E+01 & 2.5656E+00 & 6.7971E+00 \\
\cmidrule{2-8}          & OFA   & 2.3854E+01 & 2.6579E+01 & 3.0040E+01 & 2.6834E+01 & 1.7220E+00 & 5.4673E+00 \\
\cmidrule{2-8}          & CSO   & 2.1135E+01 & 2.2366E+01 & 2.5764E+01 & 2.2770E+01 & 1.4652E+00 & 6.1731E+00 \\
\cmidrule{2-8}          & JADE  & 2.1370E+01 & 2.1962E+01 & 2.2542E+01 & 2.1944E+01 & 2.9728E-01 & 6.6322E+00 \\
\cmidrule{2-8}          & CoDE  & 2.5084E+01 & 2.6758E+01 & 2.9199E+01 & 2.6879E+01 & 1.0002E+00 & 1.2787E+01 \\
\cmidrule{2-8}          & SHADE & 2.2375E+01 & 2.2970E+01 & 2.3738E+01 & 2.2944E+01 & 3.9143E-01 & 1.0042E+01 \\
\cmidrule{2-8}          & LSHADE & 2.0953E+01 & 2.0953E+01 & 2.1986E+01 & 2.1045E+01 & 2.5188E-01 & 8.4403E+00 \\
    \midrule
    \multirow{9}[18]{*}{3} & SA    & 2.7080E+02 & 2.9571E+02 & 3.3889E+02 & 2.9688E+02 & 1.5384E+01 & 1.6590E+01 \\
\cmidrule{2-8}          & SPSO  & 3.2764E+02 & 3.7085E+02 & 3.9487E+02 & 3.6773E+02 & 1.8336E+01 & 2.0000E+01 \\
\cmidrule{2-8}          & GA    & 3.5742E+02 & 4.1121E+02 & 4.7246E+02 & 4.1178E+02 & 2.6043E+01 & 1.7498E+01 \\
\cmidrule{2-8}          & OFA   & 3.7008E+02 & 4.3167E+02 & 5.5963E+02 & 4.3882E+02 & 4.2038E+01 & 1.3134E+01 \\
\cmidrule{2-8}          & CSO   & 2.5645E+02 & 2.8817E+02 & 3.2157E+02 & 2.9018E+02 & 1.6787E+01 & 1.6083E+01 \\
\cmidrule{2-8}          & JADE  & 2.7998E+02 & 2.8812E+02 & 2.9649E+02 & 2.8778E+02 & 4.8596E+00 & 1.7286E+01 \\
\cmidrule{2-8}          & CoDE  & 3.0548E+02 & 3.4264E+02 & 3.8399E+02 & 3.4323E+02 & 1.5921E+01 & 2.2902E+01 \\
\cmidrule{2-8}          & SHADE & 2.8308E+02 & 2.9356E+02 & 3.1122E+02 & 2.9471E+02 & 7.9760E+00 & 2.4603E+01 \\
\cmidrule{2-8}          & LSHADE & 2.4507E+02 & 2.5082E+02 & 2.6520E+02 & 2.5269E+02 & 5.4921E+00 & 2.1897E+01 \\
    \midrule
    \multirow{9}[18]{*}{4} & SA    & 9.5679E+06 & 1.0404E+07 & 1.1958E+07 & 1.0613E+07 & 6.6248E+05 & 5.9636E+00 \\
\cmidrule{2-8}          & SPSO  & 9.4782E+06 & 1.0196E+07 & 1.1210E+07 & 1.0176E+07 & 5.0031E+05 & 6.2734E+00 \\
\cmidrule{2-8}          & GA    & 9.6035E+06 & 1.0087E+07 & 1.0787E+07 & 1.0102E+07 & 2.8905E+05 & 4.6499E+00 \\
\cmidrule{2-8}          & OFA   & 9.4452E+06 & 9.4790E+06 & 9.5824E+06 & 9.4837E+06 & 3.3395E+04 & 3.7862E+00 \\
\cmidrule{2-8}          & CSO   & 9.4050E+06 & 9.4050E+06 & 9.4070E+06 & 9.4051E+06 & 4.4342E+02 & 4.0363E+00 \\
\cmidrule{2-8}          & JADE  & 9.4050E+06 & 9.4050E+06 & 9.4050E+06 & 9.4050E+06 & 5.1666E-03 & 4.1759E+00 \\
\cmidrule{2-8}          & CoDE  & 9.4050E+06 & 9.4051E+06 & 9.4058E+06 & 9.4052E+06 & 2.0252E+02 & 4.0380E+00 \\
\cmidrule{2-8}          & SHADE & 9.4050E+06 & 9.4050E+06 & 9.4050E+06 & 9.4050E+06 & 7.1280E-03 & 6.3865E+00 \\
\cmidrule{2-8}          & LSHADE & 9.4050E+06 & 9.4050E+06 & 9.4448E+06 & 9.4068E+06 & 7.9927E+03 & 6.6705E+00 \\
    \bottomrule
    \end{tabular}%
  \label{tableA_1}%
\end{table}%

\begin{table}[htbp]
\tiny
  \centering
  \caption{Summary statistics for $A$-optimality criterion values of designs generated by different algorithms for nodels 5-8.}
    \begin{tabular}{cccccccc}
    \toprule
    Problem & Algorithm & best  & median & worst & mean  & std   &  time \\
    \midrule
    \multirow{9}[18]{*}{5} & SA    & 2.9758E+04 & 3.3929E+04 & 4.0145E+04 & 3.4326E+04 & 3.0119E+03 & 9.7461E+00 \\
\cmidrule{2-8}          & SPSO  & 3.5928E+04 & 3.9208E+04 & 4.2549E+04 & 3.9294E+04 & 1.6376E+03 & 1.1562E+01 \\
\cmidrule{2-8}          & GA    & 3.2974E+04 & 3.6597E+04 & 4.1268E+04 & 3.6511E+04 & 1.7600E+03 & 9.2352E+00 \\
\cmidrule{2-8}          & OFA   & 3.4009E+04 & 3.8511E+04 & 4.5745E+04 & 3.8785E+04 & 2.8842E+03 & 8.3596E+00 \\
\cmidrule{2-8}          & CSO   & 2.9262E+04 & 2.9707E+04 & 3.0658E+04 & 2.9853E+04 & 4.4212E+02 & 8.3292E+00 \\
\cmidrule{2-8}          & JADE  & 2.9317E+04 & 2.9576E+04 & 3.0045E+04 & 2.9623E+04 & 1.9048E+02 & 8.6112E+00 \\
\cmidrule{2-8}          & CoDE  & 3.2457E+04 & 3.5652E+04 & 3.8275E+04 & 3.5662E+04 & 1.6036E+03 & 1.2272E+01 \\
\cmidrule{2-8}          & SHADE & 2.9347E+04 & 2.9927E+04 & 3.0749E+04 & 2.9928E+04 & 3.7348E+02 & 1.3310E+01 \\
\cmidrule{2-8}          & LSHADE & 2.9159E+04 & 2.9159E+04 & 2.9230E+04 & 2.9163E+04 & 1.4794E+01 & 1.2400E+01 \\
    \midrule
    \multirow{9}[18]{*}{6} & SA    & 8.0175E+01 & 8.0193E+01 & 8.0352E+01 & 8.0216E+01 & 5.4783E-02 & 4.6936E+00 \\
\cmidrule{2-8}          & SPSO  & 8.0200E+01 & 8.0635E+01 & 8.3339E+01 & 8.1013E+01 & 8.6480E-01 & 5.0745E+00 \\
\cmidrule{2-8}          & GA    & 8.0210E+01 & 8.0799E+01 & 8.2541E+01 & 8.0838E+01 & 4.9661E-01 & 3.8337E+00 \\
\cmidrule{2-8}          & OFA   & 8.0176E+01 & 8.0216E+01 & 8.0523E+01 & 8.0256E+01 & 9.4868E-02 & 3.1545E+00 \\
\cmidrule{2-8}          & CSO   & 8.0174E+01 & 8.0174E+01 & 8.0174E+01 & 8.0174E+01 & 3.4688E-14 & 3.0322E+00 \\
\cmidrule{2-8}          & JADE  & 8.0174E+01 & 8.0174E+01 & 8.0174E+01 & 8.0174E+01 & 9.5485E-10 & 3.3564E+00 \\
\cmidrule{2-8}          & CoDE  & 8.0174E+01 & 8.0181E+01 & 8.0228E+01 & 8.0186E+01 & 1.3382E-02 & 4.2377E+00 \\
\cmidrule{2-8}          & SHADE & 8.0174E+01 & 8.0174E+01 & 8.0174E+01 & 8.0174E+01 & 7.3451E-11 & 3.6520E+00 \\
\cmidrule{2-8}          & LSHADE & 8.0174E+01 & 8.0174E+01 & 8.0174E+01 & 8.0174E+01 & 6.0380E-07 & 5.5921E+00 \\
    \midrule
    \multirow{9}[18]{*}{7} & SA    & 9.9471E+03 & 1.1067E+04 & 2.9991E+04 & 1.2937E+04 & 4.5044E+03 & 6.6180E+00 \\
\cmidrule{2-8}          & SPSO  & 2.5409E+04 & 5.5562E+04 & 2.1145E+05 & 7.1136E+04 & 4.9829E+04 & 5.7289E+00 \\
\cmidrule{2-8}          & GA    & 1.3296E+04 & 1.6761E+04 & 2.0522E+04 & 1.6604E+04 & 1.6807E+03 & 4.6489E+00 \\
\cmidrule{2-8}          & OFA   & 1.3151E+04 & 1.7404E+04 & 2.7657E+04 & 1.8156E+04 & 3.6027E+03 & 3.8027E+00 \\
\cmidrule{2-8}          & CSO   & 9.8906E+03 & 1.0043E+04 & 2.6240E+04 & 1.1833E+04 & 3.5279E+03 & 4.4800E+00 \\
\cmidrule{2-8}          & JADE  & 9.8928E+03 & 9.9235E+03 & 9.9805E+03 & 9.9337E+03 & 2.5456E+01 & 4.5579E+00 \\
\cmidrule{2-8}          & CoDE  & 1.1313E+04 & 1.3133E+04 & 1.4179E+04 & 1.2965E+04 & 6.8265E+02 & 5.9052E+00 \\
\cmidrule{2-8}          & SHADE & 9.9220E+03 & 9.9729E+03 & 1.0274E+04 & 1.0004E+04 & 8.4994E+01 & 5.1934E+00 \\
\cmidrule{2-8}          & LSHADE & 9.8712E+03 & 9.8714E+03 & 1.0726E+04 & 9.9328E+03 & 1.8271E+02 & 6.8315E+00 \\
    \midrule
    \multirow{9}[18]{*}{8} & SA    & 2.6164E+02 & 3.5394E+02 & 1.0245E+04 & 1.1754E+03 & 2.7314E+03 & 7.6358E+02 \\
\cmidrule{2-8}          & SPSO  & 2.0337E+02 & 2.4411E+02 & 2.6659E+02 & 2.4239E+02 & 1.7058E+01 & 7.7314E+02 \\
\cmidrule{2-8}          & GA    & 4.4853E+02 & 6.7084E+02 & 8.4174E+02 & 6.7716E+02 & 8.5607E+01 & 7.1019E+02 \\
\cmidrule{2-8}          & OFA   & 1.7198E+02 & 1.9189E+02 & 2.1300E+02 & 1.9224E+02 & 1.1243E+01 & 5.9679E+02 \\
\cmidrule{2-8}          & CSO   & 1.8350E+02 & 2.6286E+02 & 3.8066E+02 & 2.6064E+02 & 4.7158E+01 & 6.4924E+02 \\
\cmidrule{2-8}          & JADE  & 1.0702E+02 & 1.0723E+02 & 1.0746E+02 & 1.0723E+02 & 1.1690E-01 & 7.0780E+02 \\
\cmidrule{2-8}          & CoDE  & 1.3400E+02 & 1.3844E+02 & 1.4526E+02 & 1.3884E+02 & 2.6946E+00 & 7.7524E+02 \\
\cmidrule{2-8}          & SHADE & 1.0684E+02 & 1.0700E+02 & 1.0711E+02 & 1.0700E+02 & 5.6425E-02 & 9.4994E+02 \\
\cmidrule{2-8}          & LSHADE & 1.0707E+02 & 1.1236E+02 & 9.7052E+02 & 2.1764E+02 & 2.2440E+02 & 1.0174E+03 \\
    \bottomrule
    \end{tabular}%
  \label{tableA_2}%
\end{table}%

\begin{table}[htbp]
\tiny
  \centering
  \caption{Summary statistics for $A$-optimality criterion values of designs generated by different algorithms for 9-12 models}
    \begin{tabular}{cccccccc}
    \toprule
    Problem & Algorithm & best  & median & worst & mean  & std   &  time \\
    \midrule
    \multirow{9}[18]{*}{9} & SA    & 1.2369E+01 & 1.4007E+01 & 1.6018E+01 & 1.4199E+01 & 9.5456E-01 & 1.7186E+03 \\
\cmidrule{2-8}          & SPSO  & 1.1587E+01 & 1.1994E+01 & 1.2593E+01 & 1.2042E+01 & 3.0643E-01 & 2.0798E+03 \\
\cmidrule{2-8}          & GA    & 1.1821E+01 & 1.2866E+01 & 1.3311E+01 & 1.2818E+01 & 3.1679E-01 & 1.9393E+03 \\
\cmidrule{2-8}          & OFA   & 9.3931E+00 & 1.0589E+01 & 1.0999E+01 & 1.0456E+01 & 4.0683E-01 & 1.5424E+03 \\
\cmidrule{2-8}          & CSO   & 7.9295E+00 & 8.7028E+00 & 9.2882E+00 & 8.6300E+00 & 3.7578E-01 & 1.5904E+03 \\
\cmidrule{2-8}          & JADE  & 7.3583E+00 & 7.4057E+00 & 7.4414E+00 & 7.4015E+00 & 1.9565E-02 & 1.7625E+03 \\
\cmidrule{2-8}          & CoDE  & 8.0619E+00 & 8.3577E+00 & 8.5831E+00 & 8.3257E+00 & 1.5422E-01 & 1.8042E+03 \\
\cmidrule{2-8}          & SHADE & 7.3431E+00 & 7.3920E+00 & 7.4168E+00 & 7.3889E+00 & 2.1006E-02 & 2.6446E+03 \\
\cmidrule{2-8}          & LSHADE & 7.3293E+00 & 7.3878E+00 & 7.4841E+00 & 7.3907E+00 & 3.9136E-02 & 2.5987E+03 \\
    \midrule
    \multirow{9}[18]{*}{10} & SA    & 2.3820E+01 & 2.9747E+01 & 4.3875E+01 & 3.0050E+01 & 3.5895E+00 & 8.0972E+02 \\
\cmidrule{2-8}          & SPSO  & 2.3032E+01 & 2.5519E+01 & 2.6387E+01 & 2.5381E+01 & 8.9345E-01 & 1.2398E+03 \\
\cmidrule{2-8}          & GA    & 2.5859E+01 & 2.7250E+01 & 2.8201E+01 & 2.7223E+01 & 6.6991E-01 & 1.1124E+03 \\
\cmidrule{2-8}          & OFA   & 2.0515E+01 & 2.2553E+01 & 2.3555E+01 & 2.2246E+01 & 8.3694E-01 & 7.9705E+02 \\
\cmidrule{2-8}          & CSO   & 1.6770E+01 & 1.7887E+01 & 1.9104E+01 & 1.7945E+01 & 6.2586E-01 & 6.9893E+02 \\
\cmidrule{2-8}          & JADE  & 1.5769E+01 & 1.5812E+01 & 1.5863E+01 & 1.5814E+01 & 2.6860E-02 & 9.3603E+02 \\
\cmidrule{2-8}          & CoDE  & 1.7166E+01 & 1.7747E+01 & 1.8451E+01 & 1.7747E+01 & 2.8636E-01 & 9.6280E+02 \\
\cmidrule{2-8}          & SHADE & 1.5751E+01 & 1.5798E+01 & 1.5837E+01 & 1.5798E+01 & 2.2046E-02 & 1.0642E+03 \\
\cmidrule{2-8}          & LSHADE & 1.5754E+01 & 1.5841E+01 & 1.8002E+01 & 1.5926E+01 & 4.3540E-01 & 1.0955E+03 \\
    \midrule
    \multirow{9}[18]{*}{11} & SA    & 3.2939E+00 & 4.0178E+00 & 2.8862E+01 & 6.5907E+00 & 6.9270E+00 & 1.0999E+03 \\
\cmidrule{2-8}          & SPSO  & 2.4552E+00 & 3.8679E+00 & 6.2302E+00 & 3.7815E+00 & 7.6485E-01 & 1.1716E+03 \\
\cmidrule{2-8}          & GA    & 5.4945E+00 & 1.3534E+01 & 3.1080E+01 & 1.5247E+01 & 6.9933E+00 & 1.0327E+03 \\
\cmidrule{2-8}          & OFA   & 1.4953E+00 & 1.7660E+00 & 2.0155E+00 & 1.7579E+00 & 1.5643E-01 & 6.6999E+02 \\
\cmidrule{2-8}          & CSO   & 1.3351E+00 & 1.6648E+00 & 2.7873E+00 & 1.7952E+00 & 3.6727E-01 & 1.0173E+03 \\
\cmidrule{2-8}          & JADE  & 1.0674E+00 & 1.0674E+00 & 1.0674E+00 & 1.0674E+00 & 9.4758E-06 & 1.0578E+03 \\
\cmidrule{2-8}          & CoDE  & 1.2481E+00 & 1.2936E+00 & 1.3465E+00 & 1.2921E+00 & 2.3161E-02 & 1.1228E+03 \\
\cmidrule{2-8}          & SHADE & 1.0674E+00 & 1.0674E+00 & 1.0675E+00 & 1.0674E+00 & 2.5077E-05 & 1.0489E+03 \\
\cmidrule{2-8}          & LSHADE & 1.0674E+00 & 1.1116E+00 & 3.9579E+00 & 1.3923E+00 & 6.3747E-01 & 1.1310E+03 \\
    \midrule
    \multirow{9}[18]{*}{12} & SA    & 2.2814E+03 & 3.4212E+03 & 7.3525E+03 & 3.5450E+03 & 1.0615E+03 & 1.1807E+03 \\
\cmidrule{2-8}          & SPSO  & 2.3281E+03 & 2.9545E+03 & 3.3204E+03 & 2.9098E+03 & 2.3086E+02 & 1.1599E+03 \\
\cmidrule{2-8}          & GA    & 2.7812E+03 & 4.0523E+03 & 4.7256E+03 & 3.9017E+03 & 5.3264E+02 & 1.0356E+03 \\
\cmidrule{2-8}          & OFA   & 8.3454E+02 & 9.7932E+02 & 1.2060E+03 & 9.8595E+02 & 9.3728E+01 & 6.8357E+02 \\
\cmidrule{2-8}          & CSO   & 8.1973E+02 & 1.2846E+03 & 2.6201E+03 & 1.3300E+03 & 4.0423E+02 & 5.3893E+02 \\
\cmidrule{2-8}          & JADE  & 3.5190E+02 & 3.6169E+02 & 3.7907E+02 & 3.6279E+02 & 6.5371E+00 & 1.0194E+03 \\
\cmidrule{2-8}          & CoDE  & 3.6852E+02 & 4.0345E+02 & 5.5261E+02 & 4.0886E+02 & 4.0404E+01 & 1.5123E+03 \\
\cmidrule{2-8}          & SHADE & 3.3836E+02 & 3.4646E+02 & 3.5645E+02 & 3.4726E+02 & 4.7466E+00 & 1.3289E+03 \\
\cmidrule{2-8}          & LSHADE & 3.0982E+02 & 3.1866E+02 & 6.4707E+02 & 3.3166E+02 & 6.5911E+01 & 1.2138E+03 \\
    \bottomrule
    \end{tabular}%
  \label{tableA_3}%
\end{table}%

Table \ref{tab:addlabel7} reports results from the Wilcoxon rank tests whether the $A$-optimal design criterion are different at the 0.05 significance level. The notations `-', `+', and `=' indicate that the corresponding comparative algorithm in the column of table is significantly worse than, better than, and similar to the target algorithm shown in the row of the table, respectively. Table \ref{tab:addlabel7} shows that LSHADE outperforms than other comparative algorithms and the traditional evolutionary algorithms (SA, SPSO, and GA) and OFA are under-perform  DE variants, such as JADE, CoDE, SHADE, and LSHADE. We note that CSO is also inferior to JADE, SHADE and LSHADE.
\begin{table}[htbp]
 \tiny
  \centering
  \caption{Results of the Wilcoxon rank tests for the equality of the medians of  values of the $A$-optimality criterion values of designs  generated from various  algorithms at the 0.05 significance level.}
    \begin{tabular}{cccccccccc}
    \toprule
    -/+/= & SA    & SPSO  & GA    & OFA   & CSO   & JADE  & CoDE  & SHADE & LSHADE \\
    \midrule
    SA    & [0/0/12] & [6/5/1] & [3/9/0] & [7/5/0] & [11/0/1] & [12/0/0] & [8/4/0] & [11/0/1] & [12/0/0] \\
    \midrule
    SPSO  & [5/6/1] & [0/0/12] & [2/8/2] & [9/1/2] & [11/0/1] & [12/0/0] & [12/0/0] & [12/0/0] & [12/0/0] \\
    \midrule
    GA    & [9/3/0] & [8/2/2] & [0/0/12] & [9/2/1] & [12/0/0] & [12/0/0] & [11/0/1] & [12/0/0] & [12/0/0] \\
    \midrule
    OFA   & [5/7/0] & [1/9/2] & [2/9/1] & [0/0/12] & [9/2/1] & [12/0/0] & [11/0/1] & [12/0/0] & [12/0/0] \\
    \midrule
    CSO   & [0/11/1] & [0/11/1] & [0/12/0] & [2/9/1] & [0/0/12] & [6/2/4] & [4/7/1] & [5/2/5] & [9/1/2] \\
    \midrule
    JADE  & [0/12/0] & [0/12/0] & [0/12/0] & [0/12/0] & [2/6/4] & [0/0/12] & [0/12/0] & [2/6/4] & [6/4/2] \\
    \midrule
    CoDE  & [4/8/0] & [0/12/0] & [0/11/1] & [0/11/1] & [7/4/1] & [12/0/0] & [0/0/12] & [12/0/0] & [11/0/1] \\
    \midrule
    SHADE & [0/11/1] & [0/12/0] & [0/12/0] & [0/12/0] & [2/5/5] & [6/2/4] & [0/12/0] & [0/0/12] & [6/4/2] \\
    \midrule
    LSHADE & [0/12/0] & [0/12/0] & [0/12/0] & [0/12/0] & [1/9/2] & [4/6/2] & [0/11/1] & [4/6/2] & [0/0/12] \\
    \bottomrule
    \end{tabular}%
  \label{tab:addlabel7}%
\end{table}%

\begin{figure}[htp]
\centering
	\includegraphics[width=0.32\linewidth]{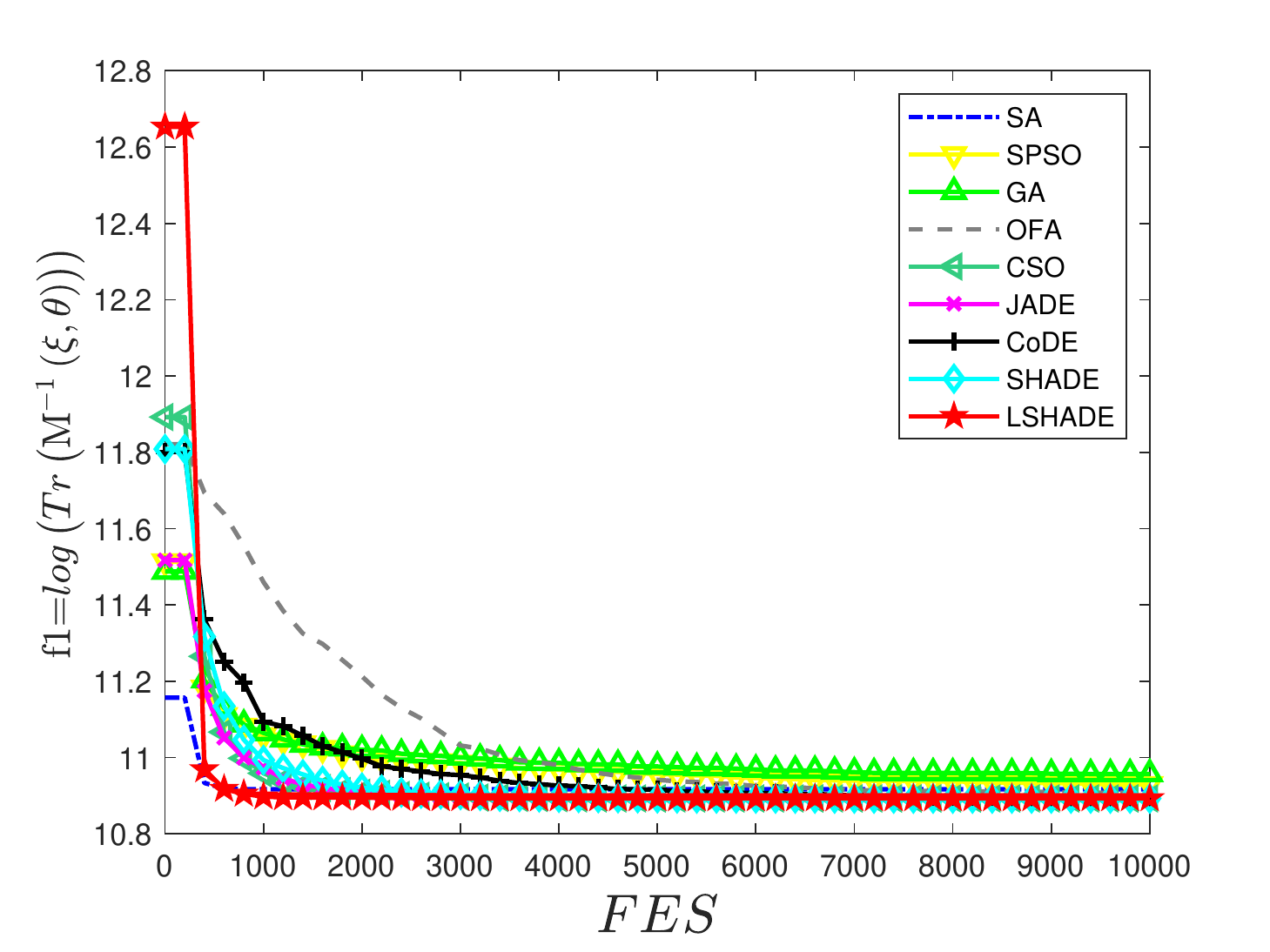}
    \label{figure1A}\hfill
	\includegraphics[width=0.32\linewidth]{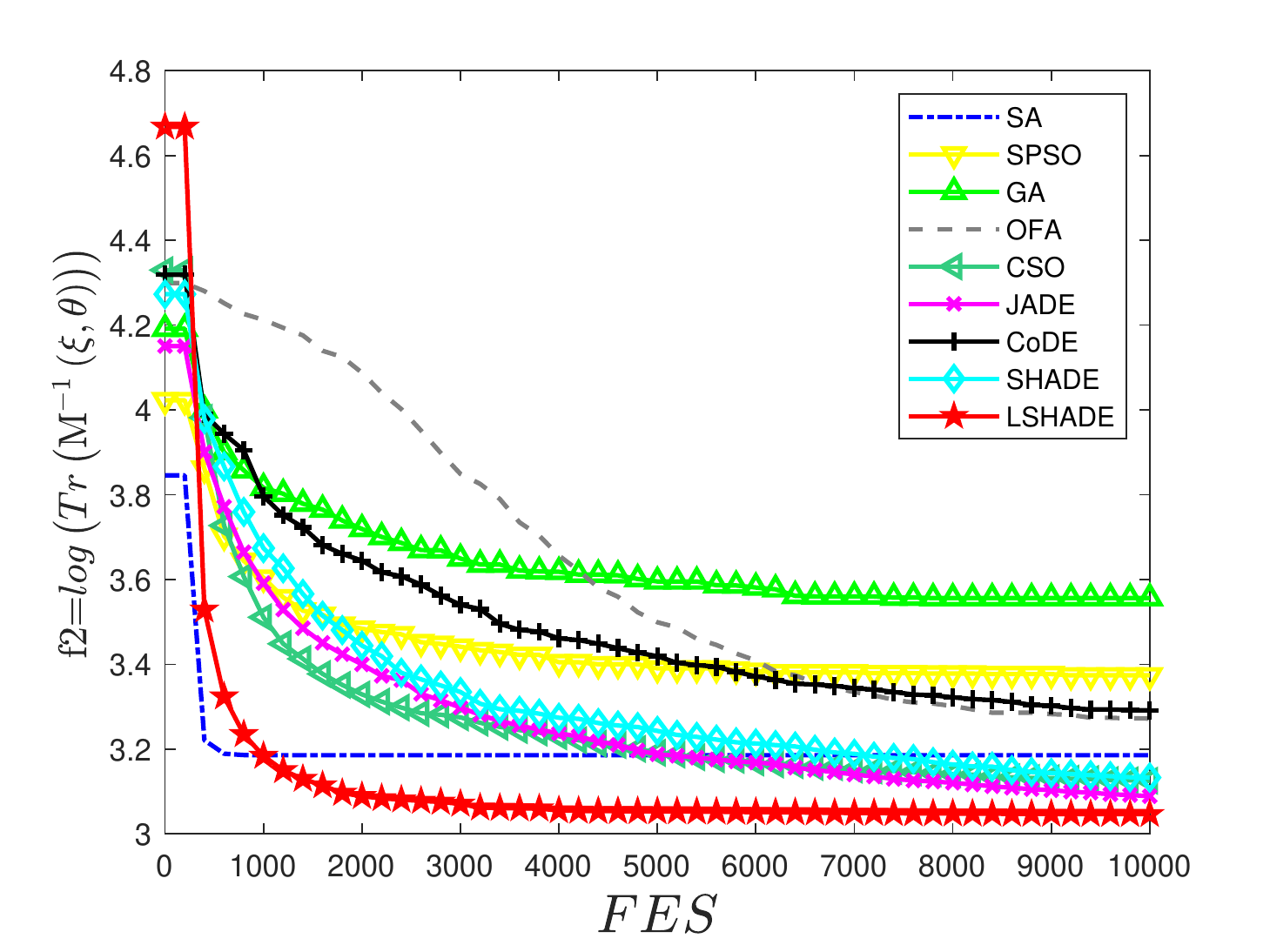}
    \label{figure2A}\hfill
    \includegraphics[width=0.32\linewidth]{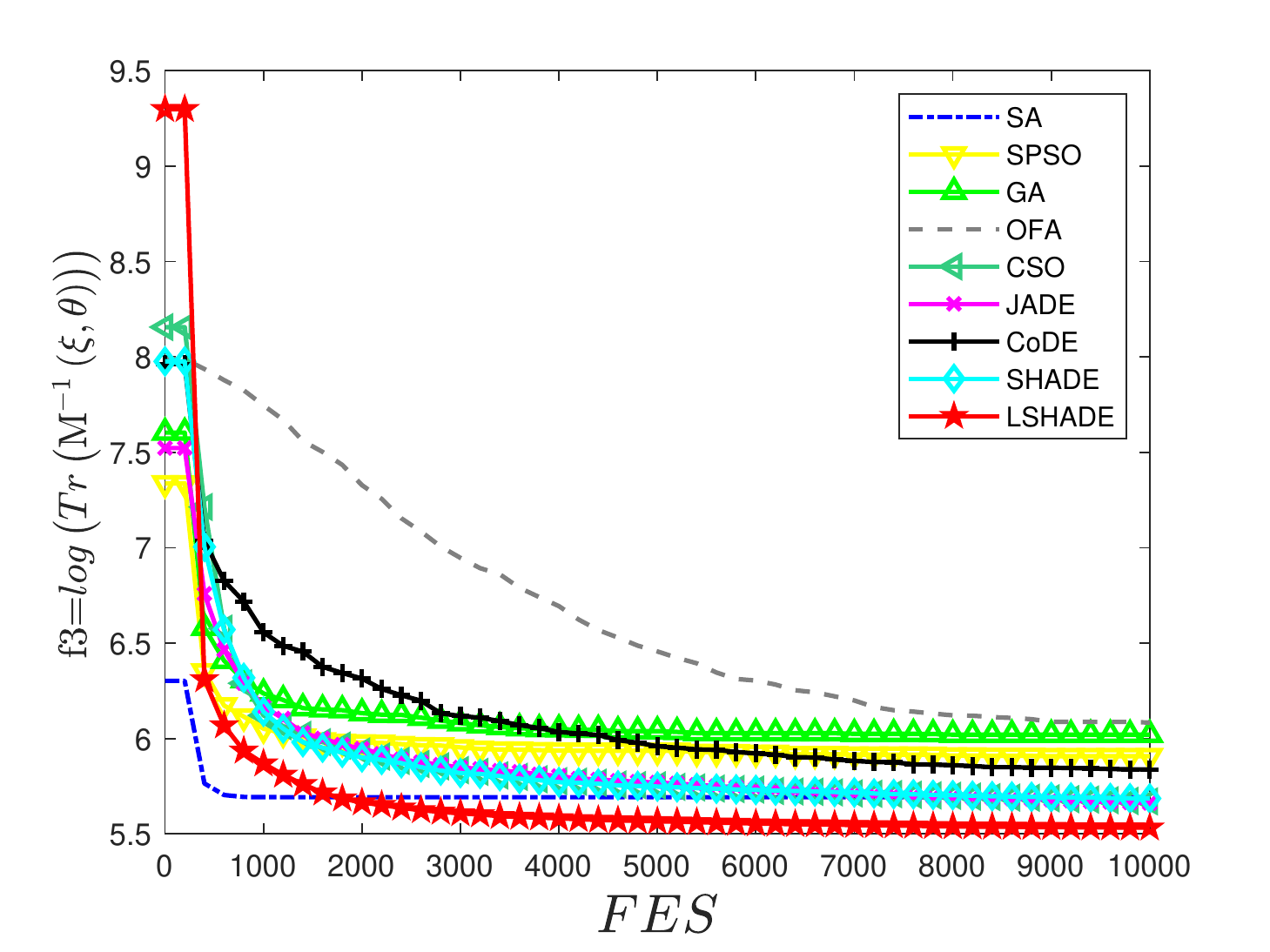}
    \label{figure3A}\\
    \includegraphics[width=0.32\linewidth]{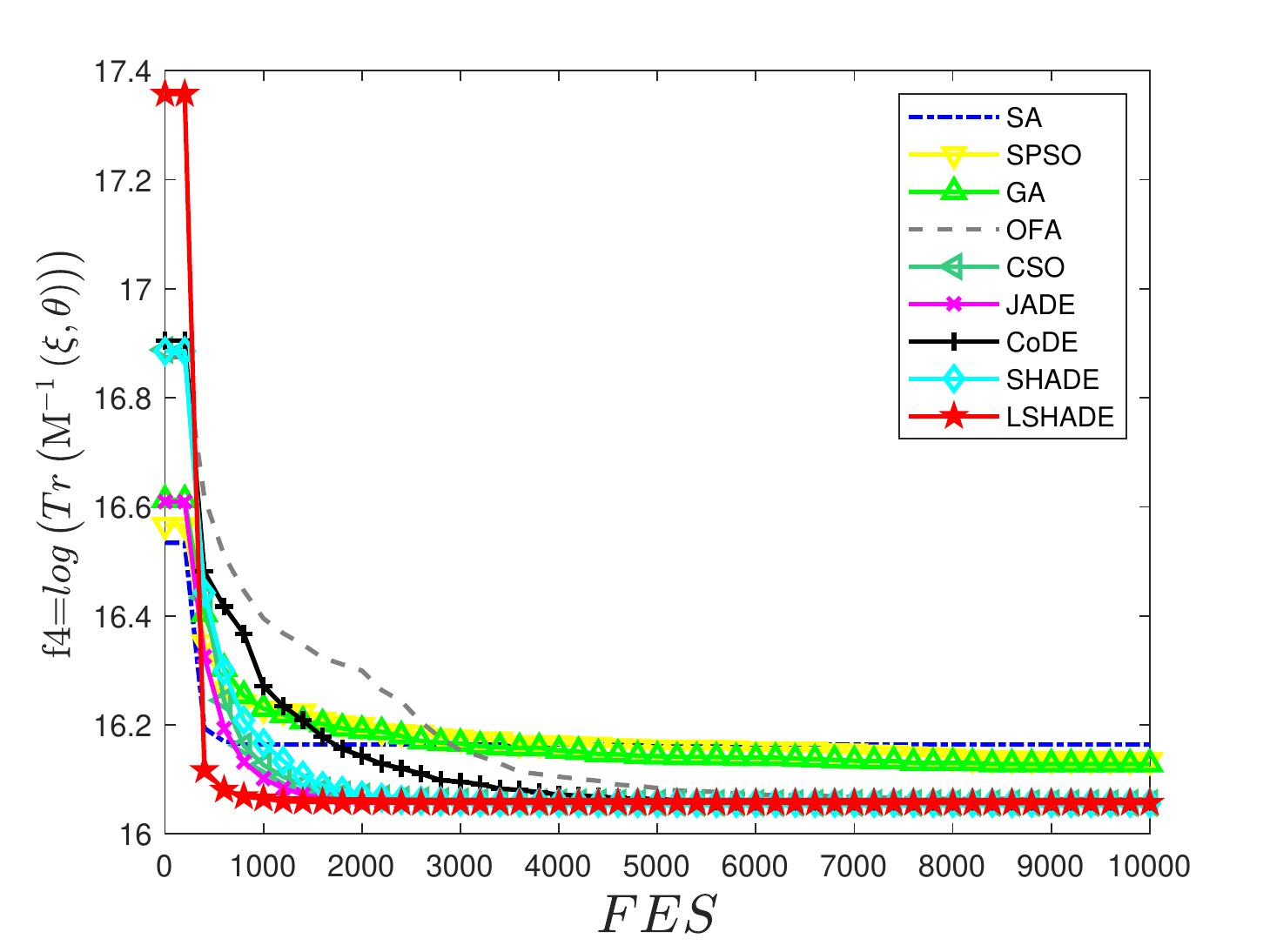}
    \label{figure4A}\hfill
    \includegraphics[width=0.32\linewidth]{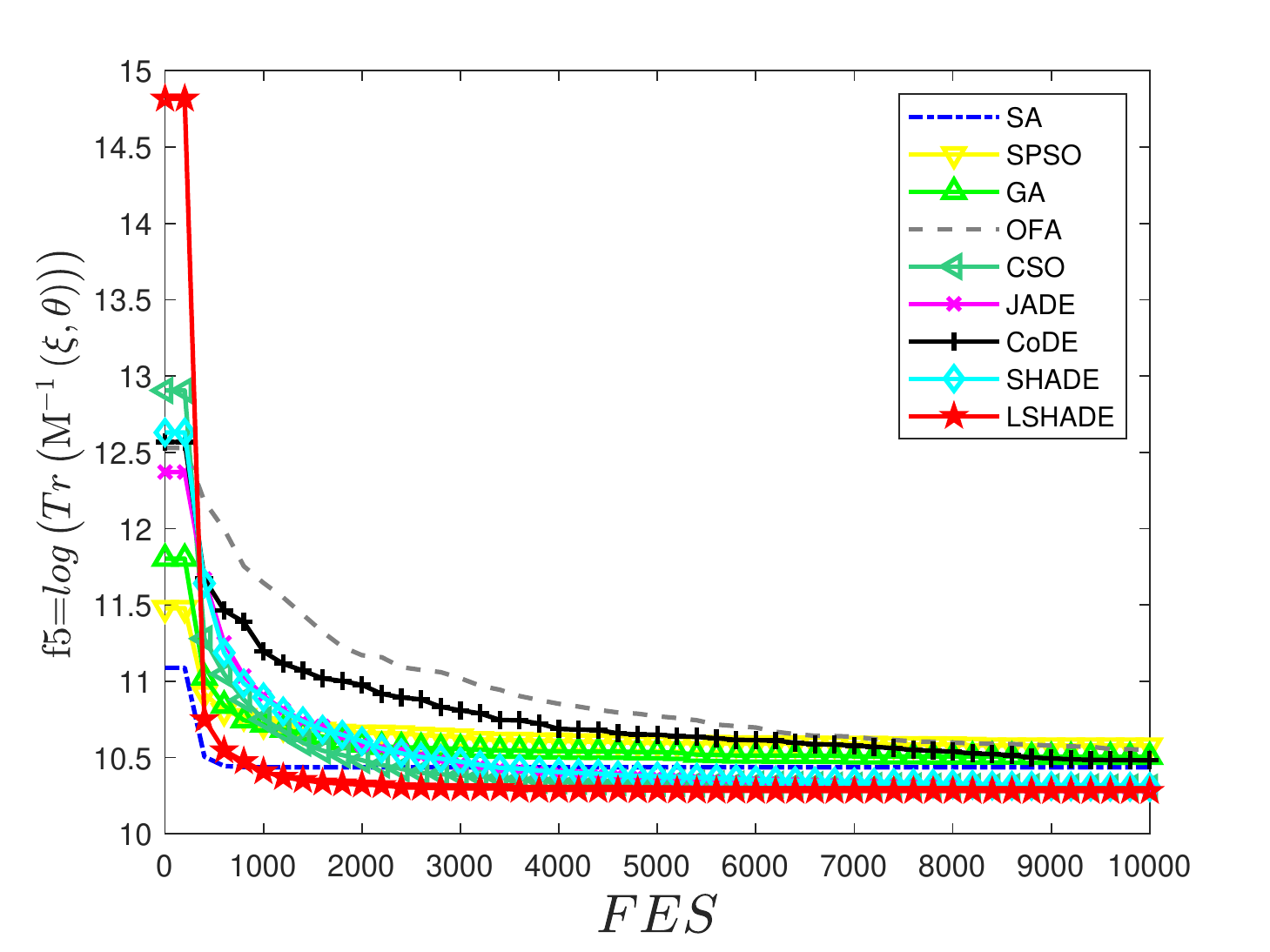}
    \label{figure5A}\hfill
    \includegraphics[width=0.32\linewidth]{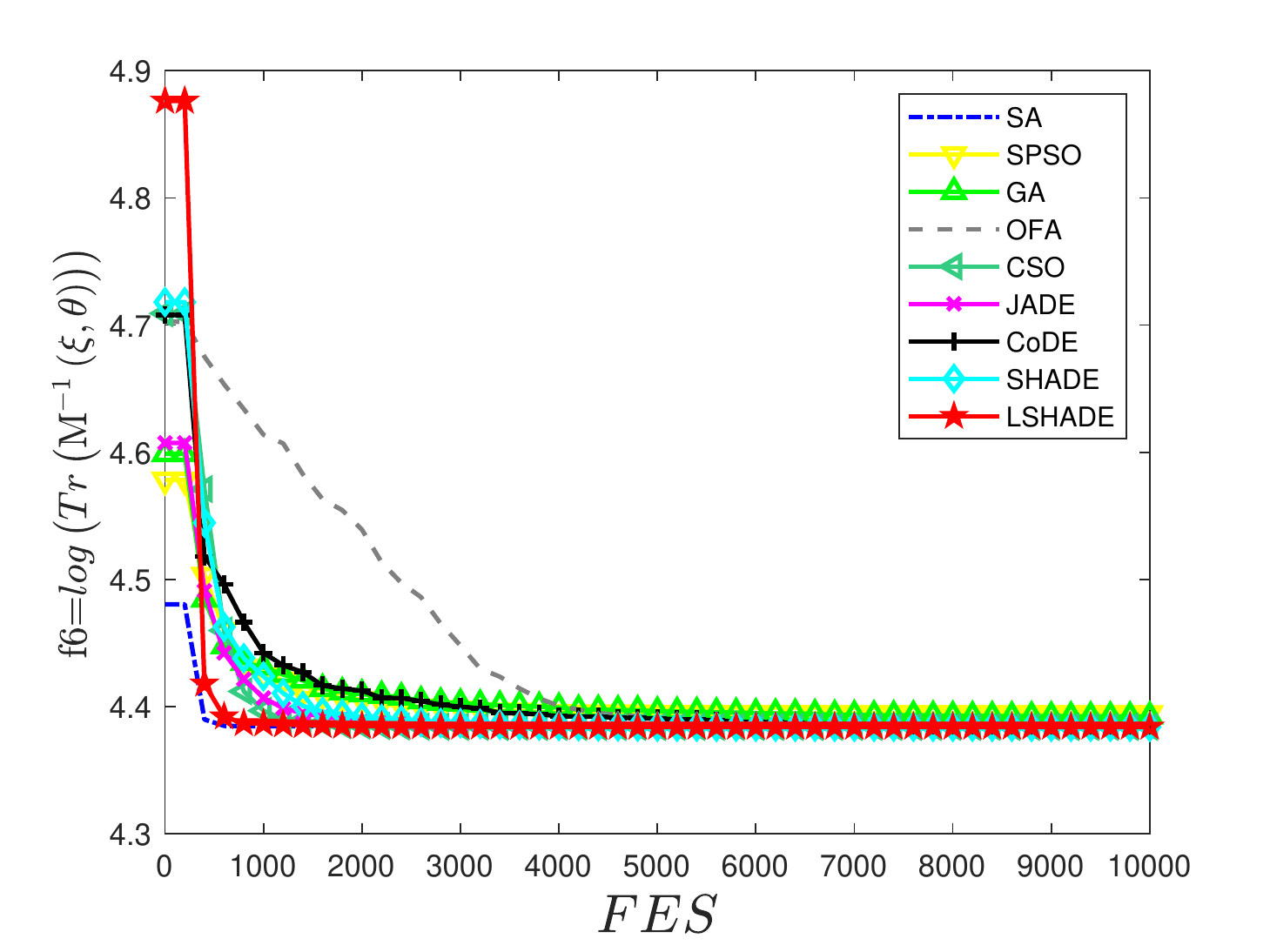}
    \label{figure6A}\\
    \includegraphics[width=0.32\linewidth]{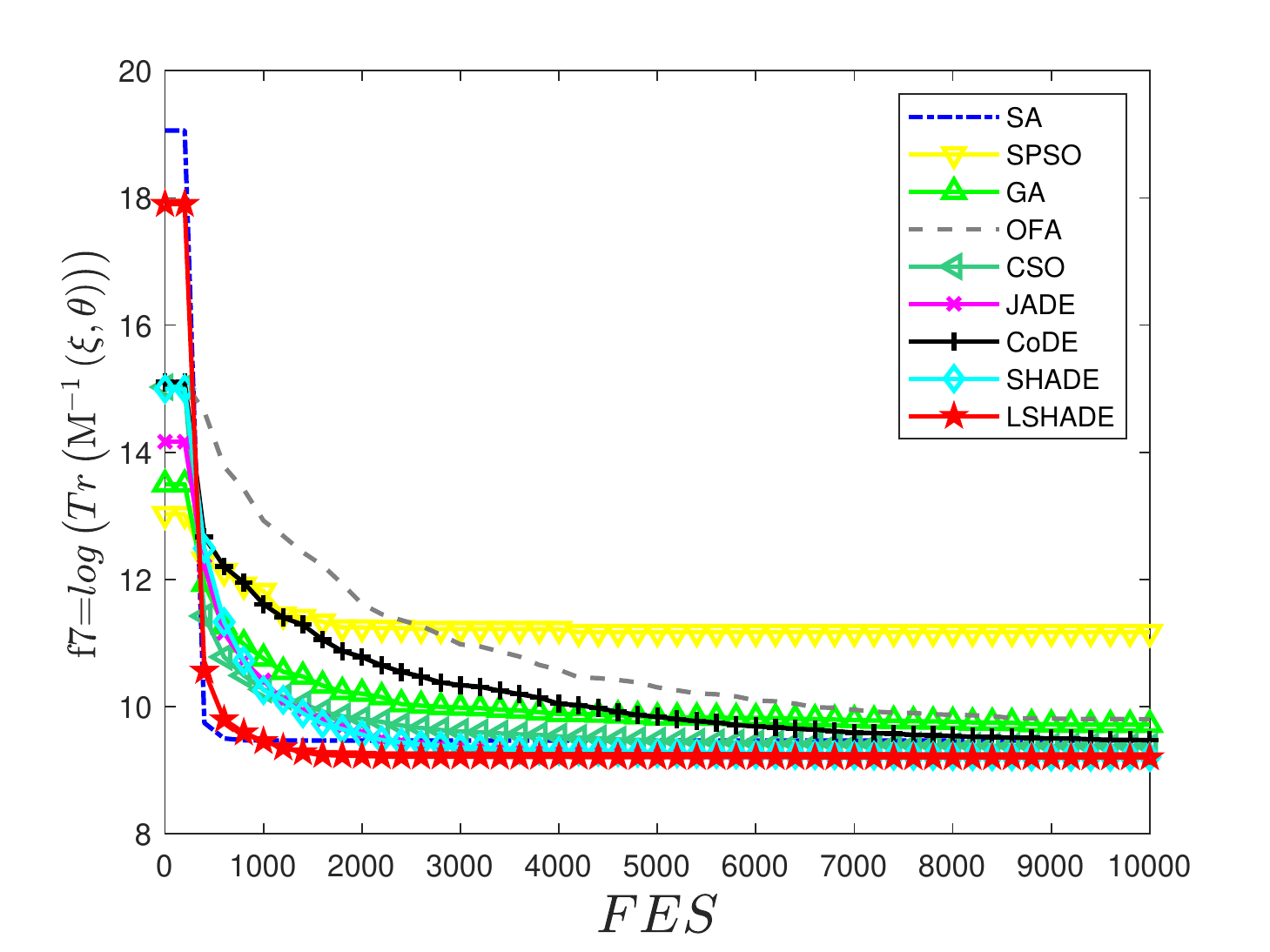}
    \label{figure7A}\hfill
	\includegraphics[width=0.32\linewidth]{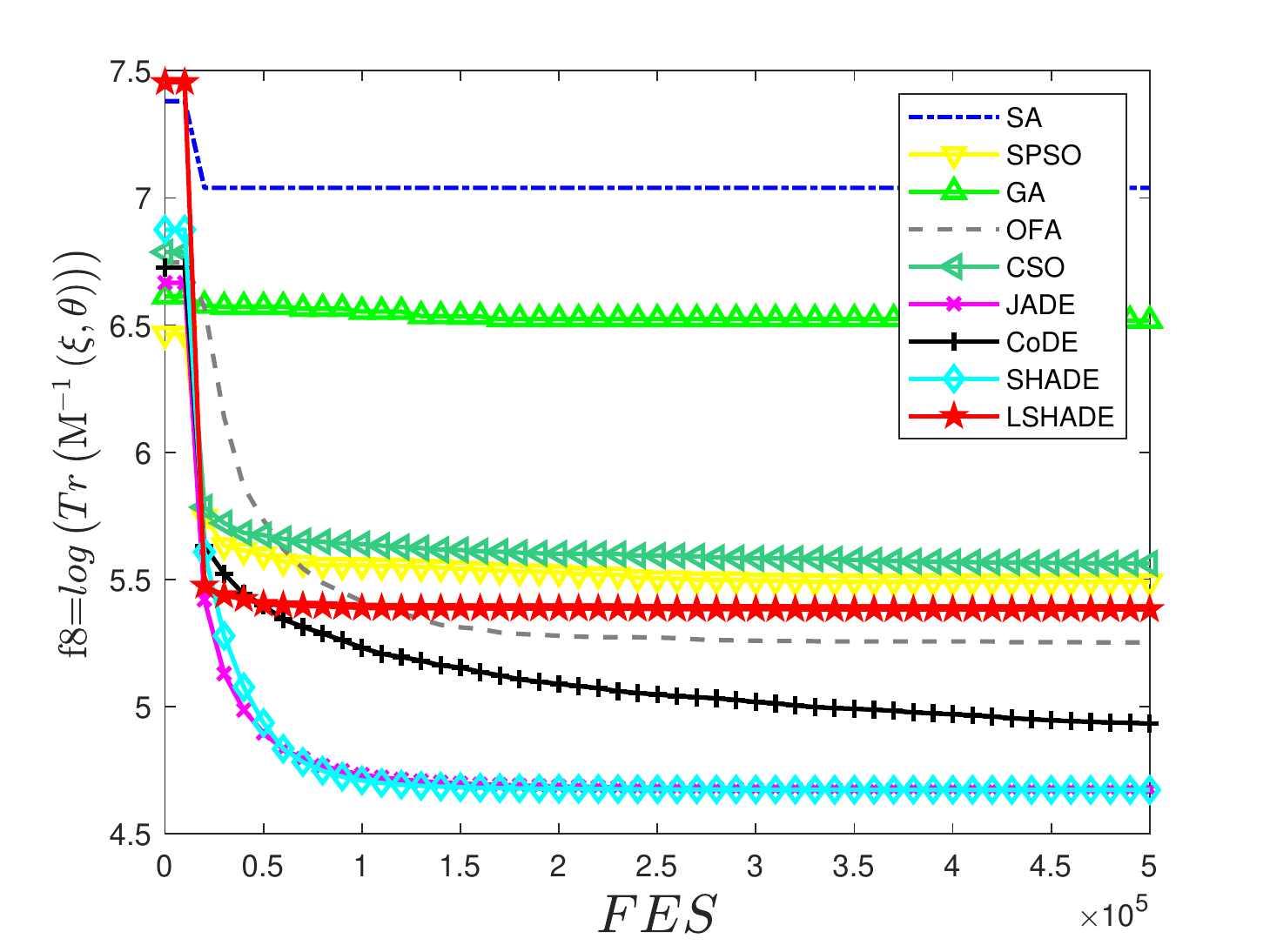}
    \label{figure8A}\hfill
    \includegraphics[width=0.32\linewidth]{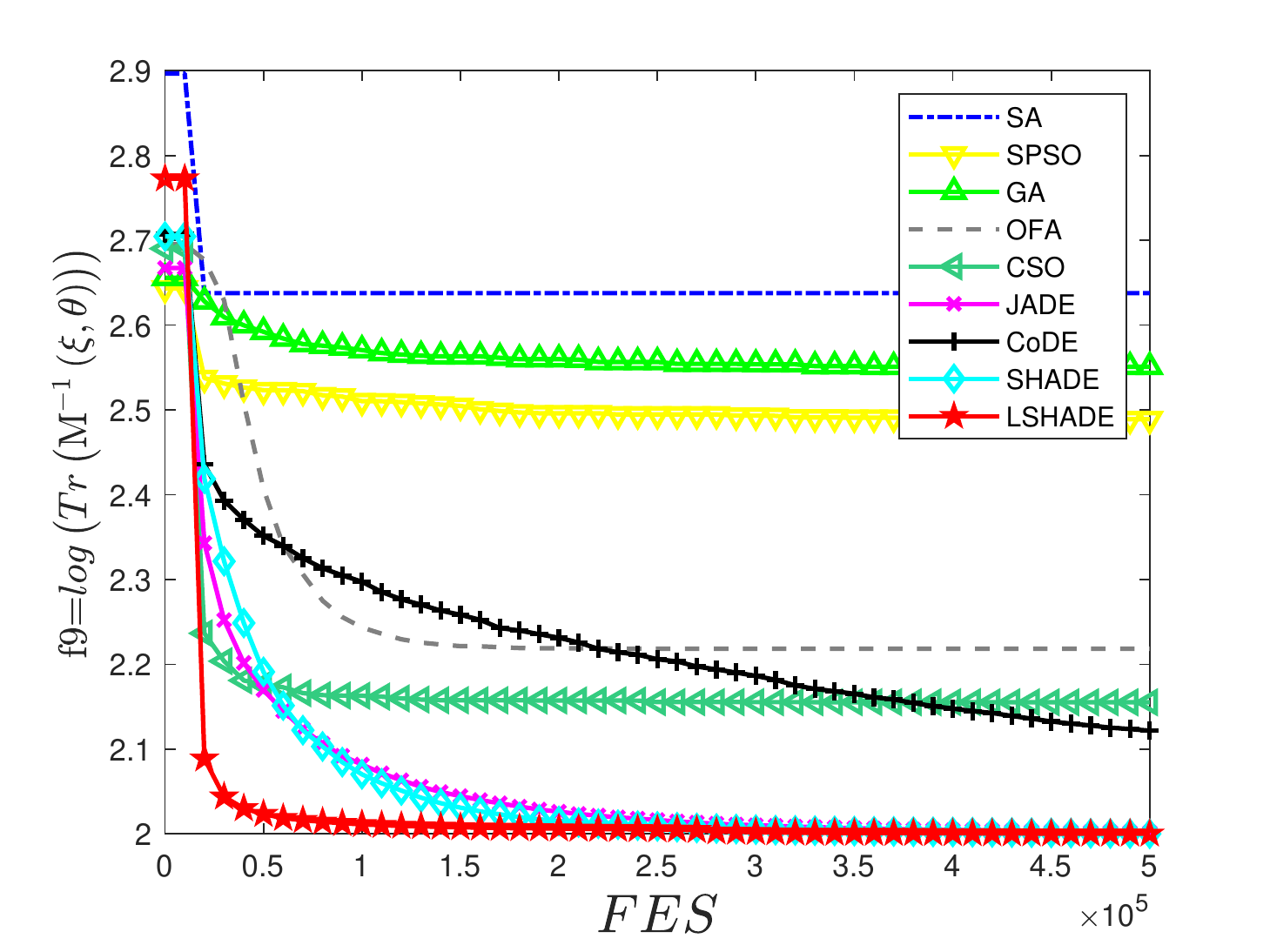}
    \label{figure9A}\\
    \includegraphics[width=0.32\linewidth]{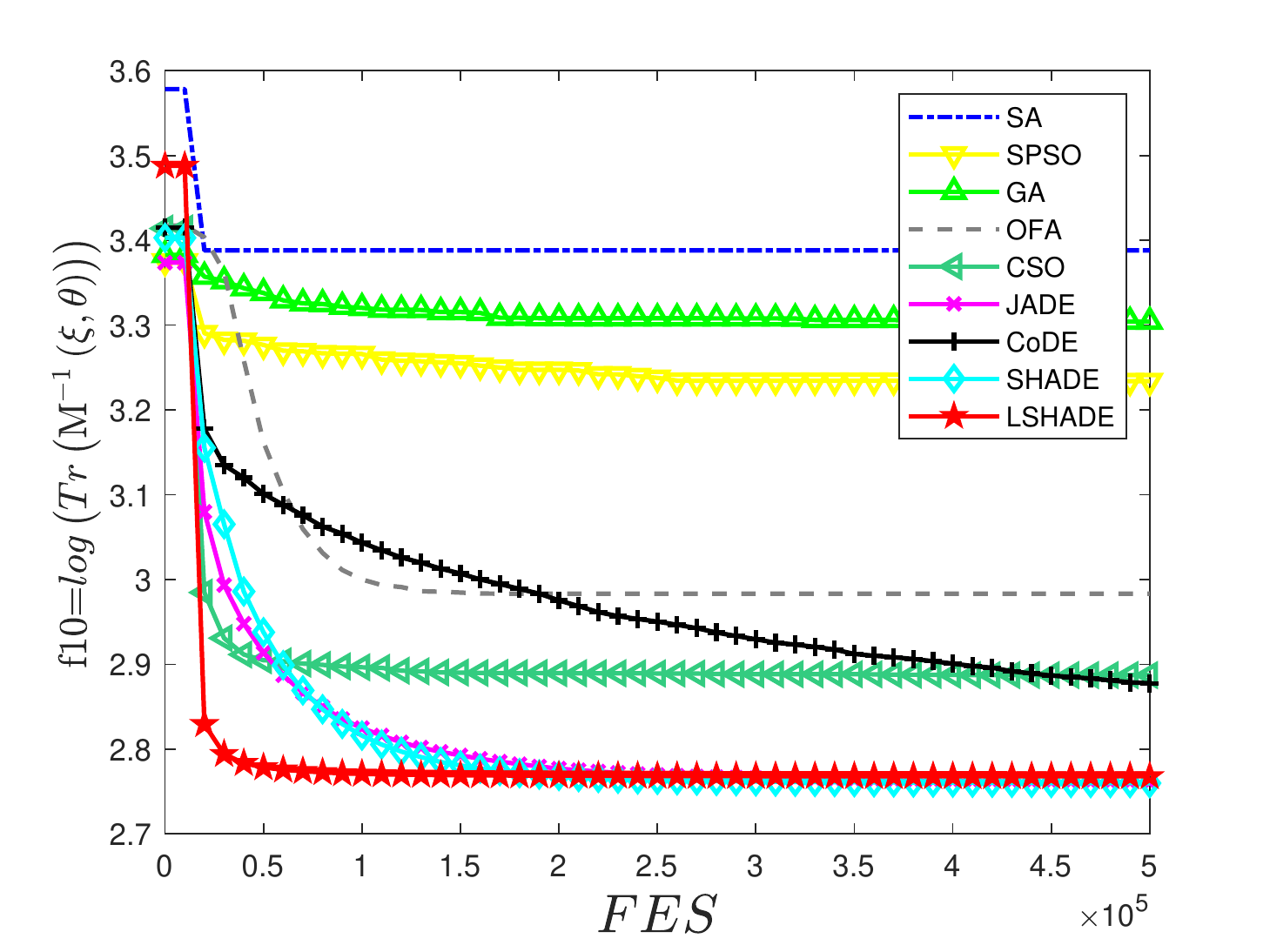}
    \label{figure10A}\hfill
    \includegraphics[width=0.32\linewidth]{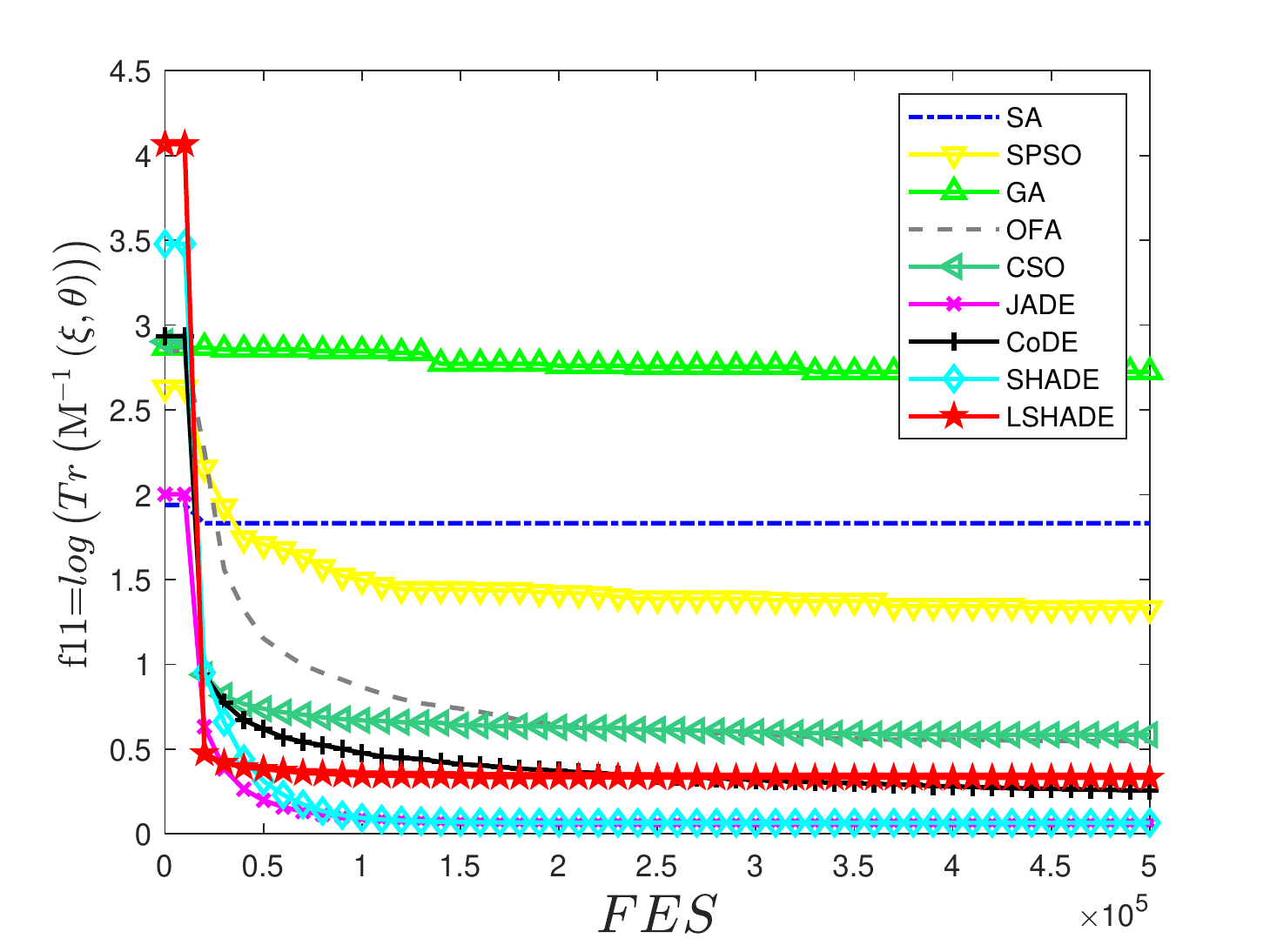}
    \label{figure11A}\hfill
    \includegraphics[width=0.32\linewidth]{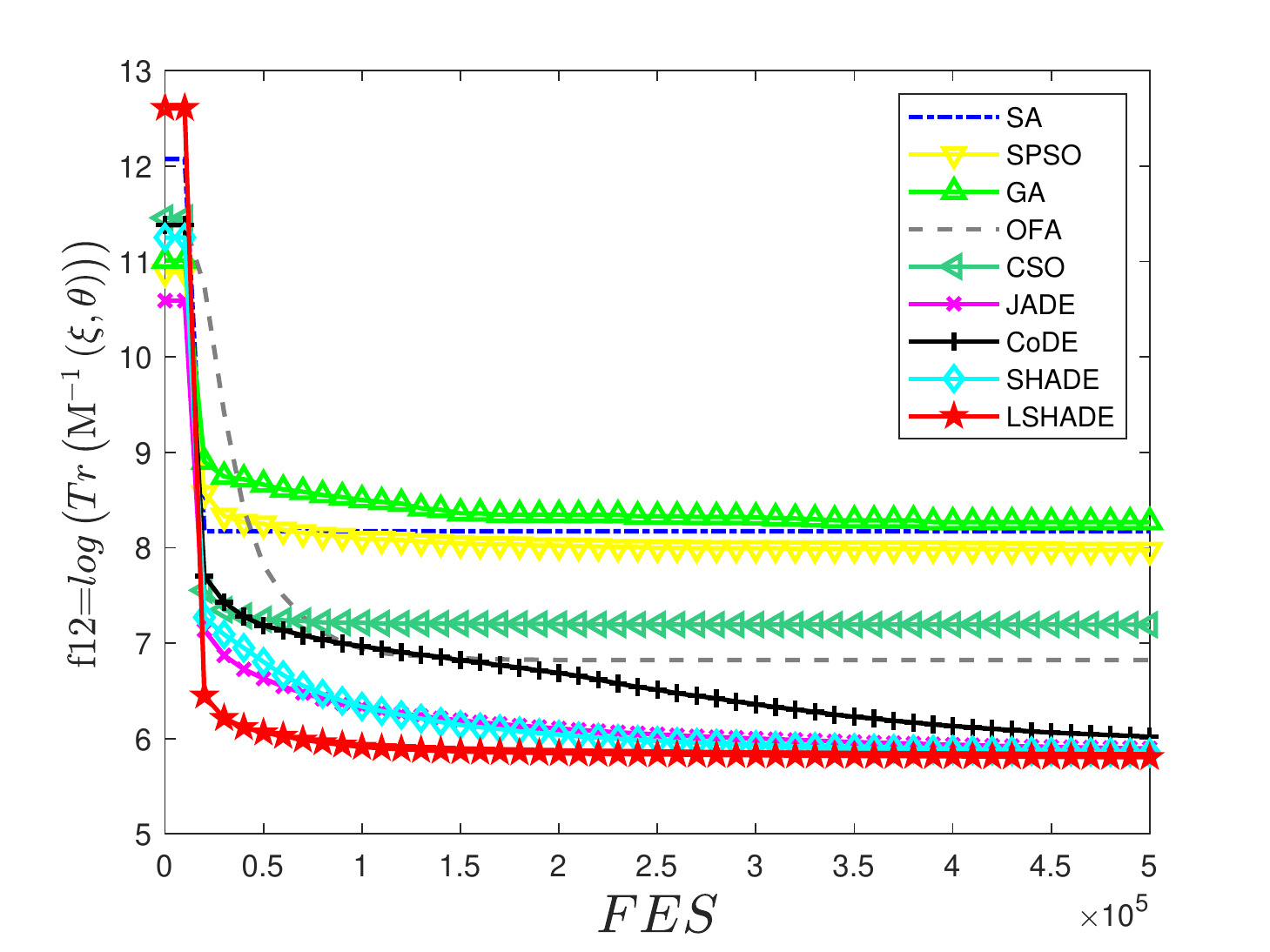}
    \label{figure12A}\\
	\caption{Mean Performance of the SA, SPSO, GA, OFA, CSO, JADE, CoDE, SHADE, and LSHADE algorithms for finding  $A$-optimal designs.}
	\label{figure3AA}
\end{figure}

Figure \ref{figure3AA} displays the convergence pattern of all algorithms. The X axial represents the number of function evaluations and the Y axial is the mean of the logarithmic $A$-optimal design criterion value. We observe that LSHADE has outstanding convergence performance for Problems 1-7, 9-10 and 12, while JADE and SHADE have better convergence characteristics for the Problems 8 and 11.

\subsection{Discussion}
The LSHADE framework is used to generate the mutation vector with the ``DE/current-to-pbest/1/bin" mutation strategy, and the parameter history update algorithm is used to adjust the adaptive parameters through the weighted Lehmer mean. Finally, the linear function is used to reduce the population scale gradually. In the repair process, since the number of  support points is unknown, similar support points in the individual are combined with their corresponding weights   along with very small weights from other  support points.  LSHADE performs very well for finding the $D$- and $A$-optimal design for many of the models in Table 1. However, in terms of convergence, the figures show LSHADE is inferior to JADE and SHADE for finding optimal designs for the 3-factor model in Problem 8, and  the gamma regression model with 5 factors with pairwise interaction terms in Problem 11.

Table \ref{tab:DA} reports the designs found  by LSHADE under the $D$- and $A$-optimality criterion for Problems 1, 2 and 4-7. The table shows their support points and corresponding weights, the sensitivity function values at each of the support points and the efficiency lower bound of the generated design. For Problems 1, 2, 4-7, the LSHADE algorithm identifies the correct number of support points, which are 4, 6, 4, 3, 2 and 4, respectively. The efficiency lower bounds in Table \ref{tab:DA} and the sensitivity plots in Figures 5 and 6 confirm the LSHADE-generated designs are either optimal or   close to the optimum with a $D$ or $A$-efficiency of at least 95\%.

\begin{table}[]
\centering
\tiny
\caption{Designs generated by the LSHADE algorithm under the $D$-optimality and $A$-optimality criteria  for Problem 1-2 and 4-7.}
\begin{tabular}{ccccccccccc}
\hline
\multirow{2}{*}{Problem} & \multicolumn{5}{c}{D-optimal design}                                                                                                                                                 & \multicolumn{5}{c}{A-optimal design}                                                                                                                                                           \\ \cline{2-11}
                         & \multicolumn{2}{c}{X}                                    & \multicolumn{1}{c}{W}      & \multicolumn{1}{c}{S} & \begin{tabular}[c]{@{}c@{}}D-efficiency\\ lower bound\end{tabular} & \multicolumn{2}{c}{X}                                    & \multicolumn{1}{c}{W}      & \multicolumn{1}{c}{S}          & \begin{tabular}[c]{@{}c@{}}A-efficiency \\ lower bound\end{tabular} \\ \hline
\multirow{4}{*}{1}       & \multicolumn{2}{c|}{0}                                    & \multicolumn{1}{c|}{0.25}   & \multicolumn{1}{c|}{0} & \multirow{4}{*}{1}                                                 & \multicolumn{2}{c|}{0}                                    & \multicolumn{1}{c|}{0.0857} & \multicolumn{1}{c|}{0.0074}     & \multirow{4}{*}{0.9999}                                             \\ \cline{2-5} \cline{7-10}
                         & \multicolumn{2}{c|}{0.3141}                               & \multicolumn{1}{c|}{0.25}   & \multicolumn{1}{c|}{0} &                                                                    & \multicolumn{2}{c|}{0.2723}                               & \multicolumn{1}{c|}{0.1957} & \multicolumn{1}{c|}{0.002}      &                                                                     \\ \cline{2-5} \cline{7-10}
                         & \multicolumn{2}{c|}{1.1307}                               & \multicolumn{1}{c|}{0.25}   & \multicolumn{1}{c|}{0} &                                                                    & \multicolumn{2}{c|}{1.1827}                               & \multicolumn{1}{c|}{0.2861} & \multicolumn{1}{c|}{-0.0118}    &                                                                     \\ \cline{2-5} \cline{7-10}
                         & \multicolumn{2}{c|}{2.7523}                               & \multicolumn{1}{c|}{0.25}   & \multicolumn{1}{c|}{0} &                                                                    & \multicolumn{2}{c|}{3}                                    & \multicolumn{1}{c|}{0.4325} & \multicolumn{1}{c|}{0.0054}     &                                                                     \\ \hline
\multirow{6}{*}{2}       & \multicolumn{1}{c|}{-1}     & \multicolumn{1}{c|}{0}      & \multicolumn{1}{c|}{0.1875} & \multicolumn{1}{c|}{0} & \multirow{6}{*}{1}                                                 & \multicolumn{1}{c|}{-1}     & \multicolumn{1}{c|}{0}      & \multicolumn{1}{c|}{0.1859} & \multicolumn{1}{c|}{0.0001}     & \multirow{6}{*}{0.9999}                                             \\ \cline{2-5} \cline{7-10}
                         & \multicolumn{1}{c|}{-1}     & \multicolumn{1}{c|}{1}      & \multicolumn{1}{c|}{0.1875} & \multicolumn{1}{c|}{0} &                                                                    & \multicolumn{1}{c|}{-1}     & \multicolumn{1}{c|}{1}      & \multicolumn{1}{c|}{0.1399} & \multicolumn{1}{c|}{-0.0003}    &                                                                     \\ \cline{2-5} \cline{7-10}
                         & \multicolumn{1}{c|}{0}      & \multicolumn{1}{c|}{1}      & \multicolumn{1}{c|}{0.125}  & \multicolumn{1}{c|}{0} &                                                                    & \multicolumn{1}{c|}{0}      & \multicolumn{1}{c|}{0}      & \multicolumn{1}{c|}{0.2287} & \multicolumn{1}{c|}{0.0001}     &                                                                     \\ \cline{2-5} \cline{7-10}
                         & \multicolumn{1}{c|}{0}      & \multicolumn{1}{c|}{0}      & \multicolumn{1}{c|}{0.125}  & \multicolumn{1}{c|}{0} &                                                                    & \multicolumn{1}{c|}{0}      & \multicolumn{1}{c|}{1}      & \multicolumn{1}{c|}{0.1197} & \multicolumn{1}{c|}{-0.0001}    &                                                                     \\ \cline{2-5} \cline{7-10}
                         & \multicolumn{1}{c|}{1}      & \multicolumn{1}{c|}{1}      & \multicolumn{1}{c|}{0.1875} & \multicolumn{1}{c|}{0} &                                                                    & \multicolumn{1}{c|}{1}      & \multicolumn{1}{c|}{1}      & \multicolumn{1}{c|}{0.1399} & \multicolumn{1}{c|}{0.0001}     &                                                                     \\ \cline{2-5} \cline{7-10}
                         & \multicolumn{1}{c|}{1}      & \multicolumn{1}{c|}{0}      & \multicolumn{1}{c|}{0.1875} & \multicolumn{1}{c|}{0} &                                                                    & \multicolumn{1}{c|}{1}      & \multicolumn{1}{c|}{0}      & \multicolumn{1}{c|}{0.1859} & \multicolumn{1}{c|}{0}          &                                                                     \\ \hline
\multirow{4}{*}{4}       & \multicolumn{2}{c|}{0}                                    & \multicolumn{1}{c|}{0.25}   & \multicolumn{1}{c|}{0} & \multirow{4}{*}{1}                                                 & \multicolumn{2}{c|}{0}                                    & \multicolumn{1}{c|}{0.1888} & \multicolumn{1}{c|}{-1009.4201} & \multirow{4}{*}{0.9999}                                             \\ \cline{2-5} \cline{7-10}
                         & \multicolumn{2}{c|}{0.3305}                               & \multicolumn{1}{c|}{0.25}   & \multicolumn{1}{c|}{0} &                                                                    & \multicolumn{2}{c|}{0.3011}                               & \multicolumn{1}{c|}{0.3509} & \multicolumn{1}{c|}{296.8851}   &                                                                     \\ \cline{2-5} \cline{7-10}
                         & \multicolumn{2}{c|}{0.7692}                               & \multicolumn{1}{c|}{0.25}   & \multicolumn{1}{c|}{0} &                                                                    & \multicolumn{2}{c|}{0.7926}                               & \multicolumn{1}{c|}{0.3119} & \multicolumn{1}{c|}{-252.4479}  &                                                                     \\ \cline{2-5} \cline{7-10}
                         & \multicolumn{2}{c|}{1}                                    & \multicolumn{1}{c|}{0.25}   & \multicolumn{1}{c|}{0} &                                                                    & \multicolumn{2}{c|}{1}                                    & \multicolumn{1}{c|}{0.1484} & \multicolumn{1}{c|}{1112.3723}  &                                                                     \\ \hline
\multirow{3}{*}{5}       & \multicolumn{1}{c|}{0.2804} & \multicolumn{1}{c|}{0}      & \multicolumn{1}{c|}{0.3333} & \multicolumn{1}{c|}{0} & \multirow{3}{*}{1}                                                 & \multicolumn{1}{c|}{0.2603} & \multicolumn{1}{c|}{0}      & \multicolumn{1}{c|}{0.4785} & \multicolumn{1}{c|}{-0.0007}    & \multirow{3}{*}{0.9999}                                             \\ \cline{2-5} \cline{7-10}
                         & \multicolumn{1}{c|}{3}      & \multicolumn{1}{c|}{0}      & \multicolumn{1}{c|}{0.3333} & \multicolumn{1}{c|}{0} &                                                                    & \multicolumn{1}{c|}{3}      & \multicolumn{1}{c|}{0}      & \multicolumn{1}{c|}{0.0595} & \multicolumn{1}{c|}{0.0008}     &                                                                     \\ \cline{2-5} \cline{7-10}
                         & \multicolumn{1}{c|}{3}      & \multicolumn{1}{c|}{0.7951} & \multicolumn{1}{c|}{0.3333} & \multicolumn{1}{c|}{0} &                                                                    & \multicolumn{1}{c|}{3}      & \multicolumn{1}{c|}{0.826}  & \multicolumn{1}{c|}{0.462}  & \multicolumn{1}{c|}{0.0007}     &                                                                     \\ \hline
\multirow{2}{*}{6}       & \multicolumn{2}{c|}{0.7143}                               & \multicolumn{1}{c|}{0.5}    & \multicolumn{1}{c|}{0} & \multirow{2}{*}{1}                                                 & \multicolumn{2}{c|}{0.5373}                               & \multicolumn{1}{c|}{0.6696} & \multicolumn{1}{c|}{0}          & \multirow{2}{*}{1}                                                  \\ \cline{2-5} \cline{7-10}
                         & \multicolumn{2}{c|}{5}                                    & \multicolumn{1}{c|}{0.5}    & \multicolumn{1}{c|}{0} &                                                                    & \multicolumn{2}{c|}{5}                                    & \multicolumn{1}{c|}{0.3304} & \multicolumn{1}{c|}{0}          &                                                                     \\ \hline
\multirow{4}{*}{7}       & \multicolumn{1}{c|}{3.1579} & \multicolumn{1}{c|}{0}      & \multicolumn{1}{c|}{0.25}   & \multicolumn{1}{c|}{0} & \multirow{4}{*}{1}                                                 & \multicolumn{1}{c|}{2.4402} & \multicolumn{1}{c|}{0}      & \multicolumn{1}{c|}{0.2651} & \multicolumn{1}{c|}{0.0001}     & \multirow{4}{*}{0.9999}                                             \\ \cline{2-5} \cline{7-10}
                         & \multicolumn{1}{c|}{4.0793} & \multicolumn{1}{c|}{2.6754} & \multicolumn{1}{c|}{0.25}   & \multicolumn{1}{c|}{0} &                                                                    & \multicolumn{1}{c|}{3.3919} & \multicolumn{1}{c|}{3.2516} & \multicolumn{1}{c|}{0.3234} & \multicolumn{1}{c|}{0.0016}     &                                                                     \\ \cline{2-5} \cline{7-10}
                         & \multicolumn{1}{c|}{30}     & \multicolumn{1}{c|}{0}      & \multicolumn{1}{c|}{0.25}   & \multicolumn{1}{c|}{0} &                                                                    & \multicolumn{1}{c|}{30}     & \multicolumn{1}{c|}{0}      & \multicolumn{1}{c|}{0.1398} & \multicolumn{1}{c|}{-0.0003}    &                                                                     \\ \cline{2-5} \cline{7-10}
                         & \multicolumn{1}{c|}{30}     & \multicolumn{1}{c|}{3.5789} & \multicolumn{1}{c|}{0.25}   & \multicolumn{1}{c|}{0} &                                                                    & \multicolumn{1}{c|}{30}     & \multicolumn{1}{c|}{4.7409} & \multicolumn{1}{c|}{0.2717} & \multicolumn{1}{c|}{-0.0018}    &                                                                     \\ \hline
\end{tabular}
\label{tab:DA}
\end{table}

\begin{figure}[htbp]
\centering
	\includegraphics[width=0.32\linewidth]{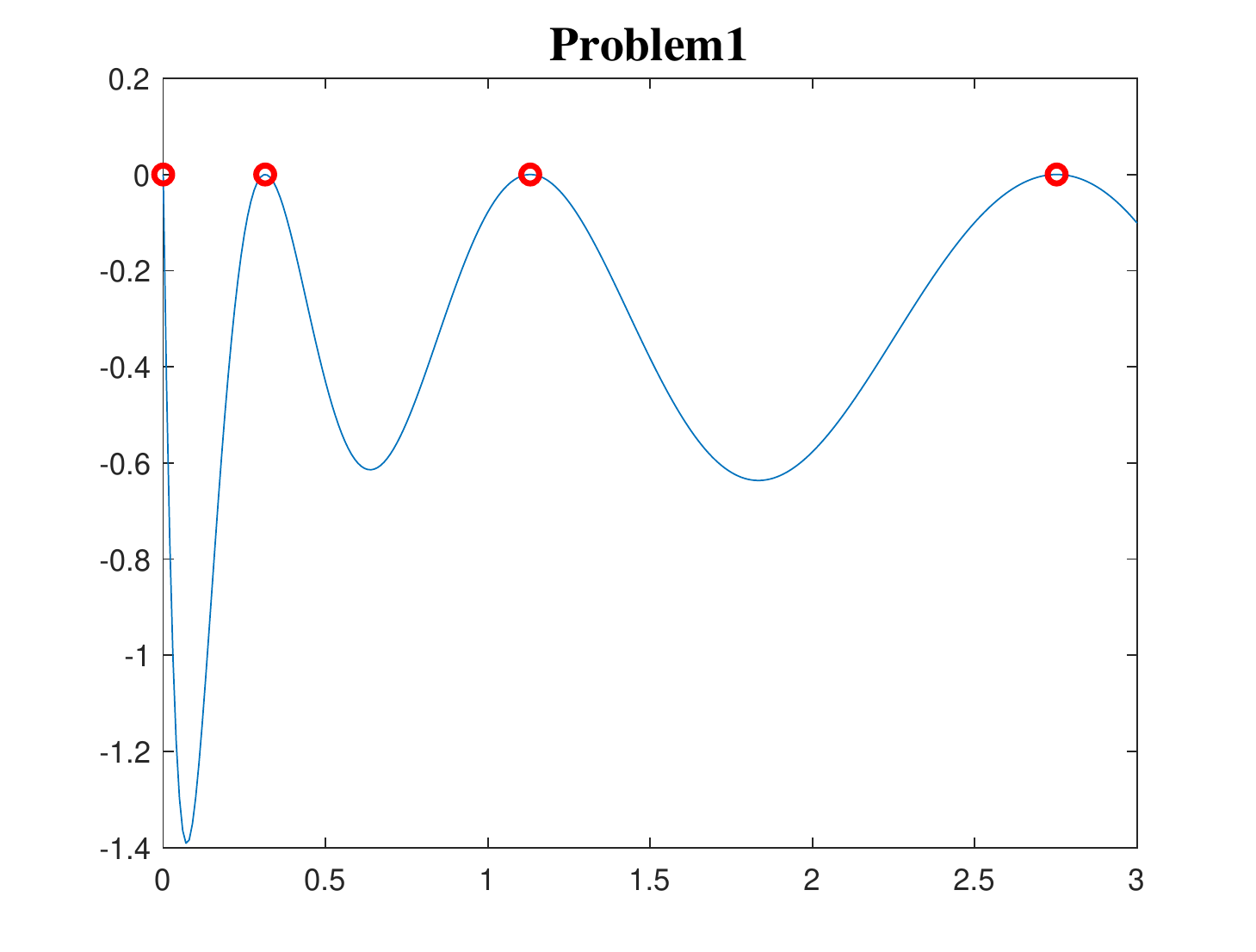}
    \label{DPro1_LSHADE}\hfill
	\includegraphics[width=0.32\linewidth]{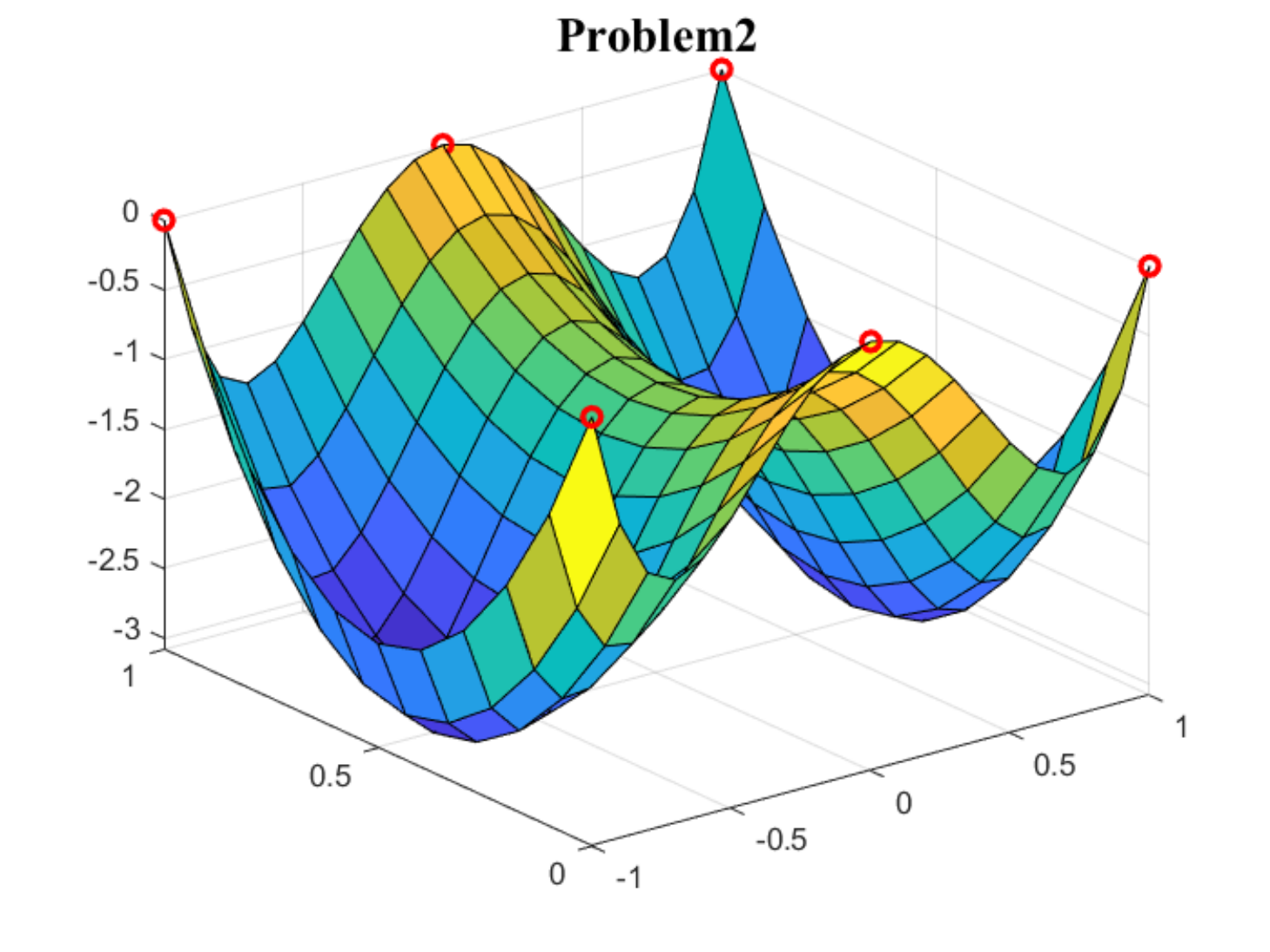}
    \label{DPro2_LSHADE}\hfill
    \includegraphics[width=0.32\linewidth]{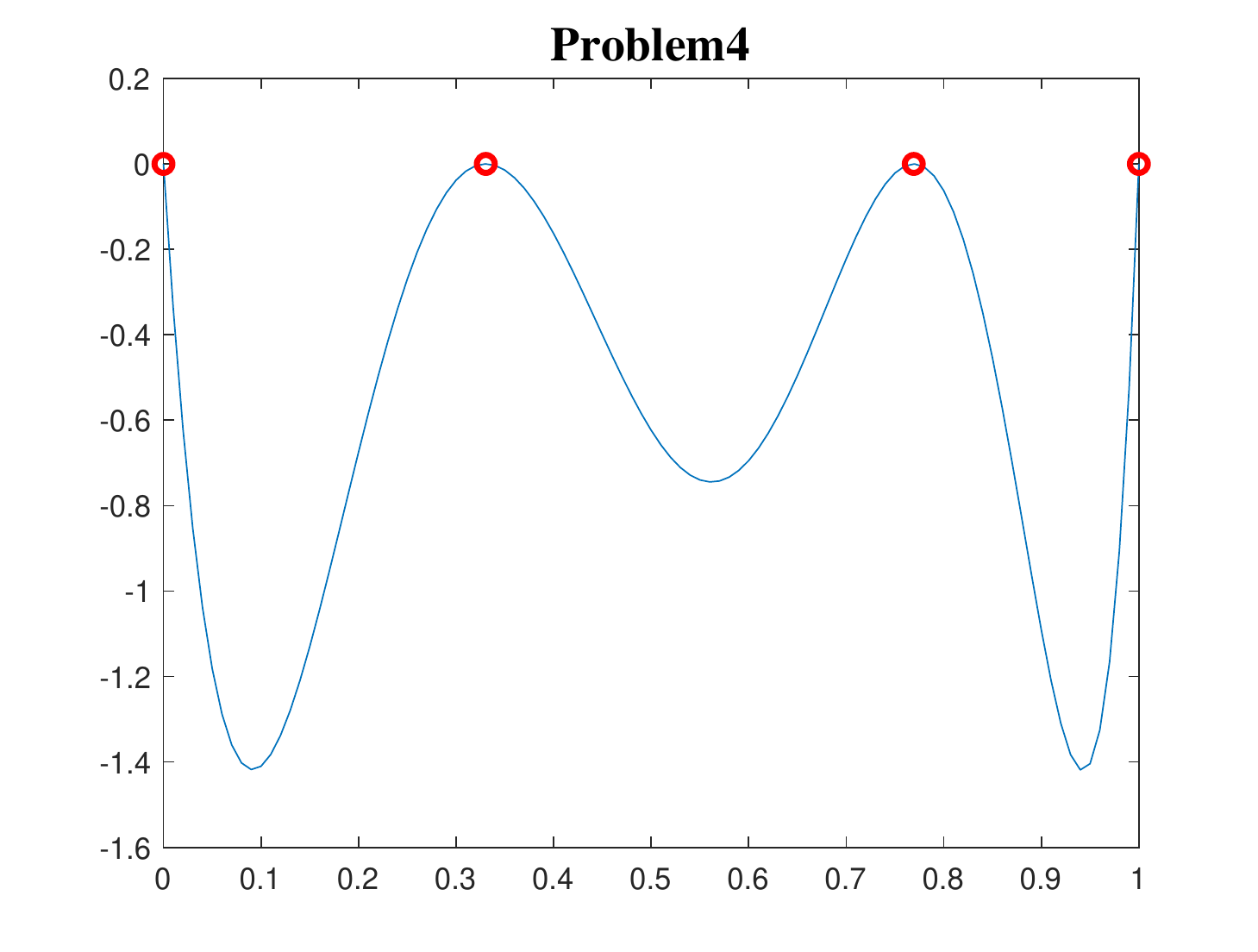}
    \label{DPro4_LSHADE}\\
    \includegraphics[width=0.32\linewidth]{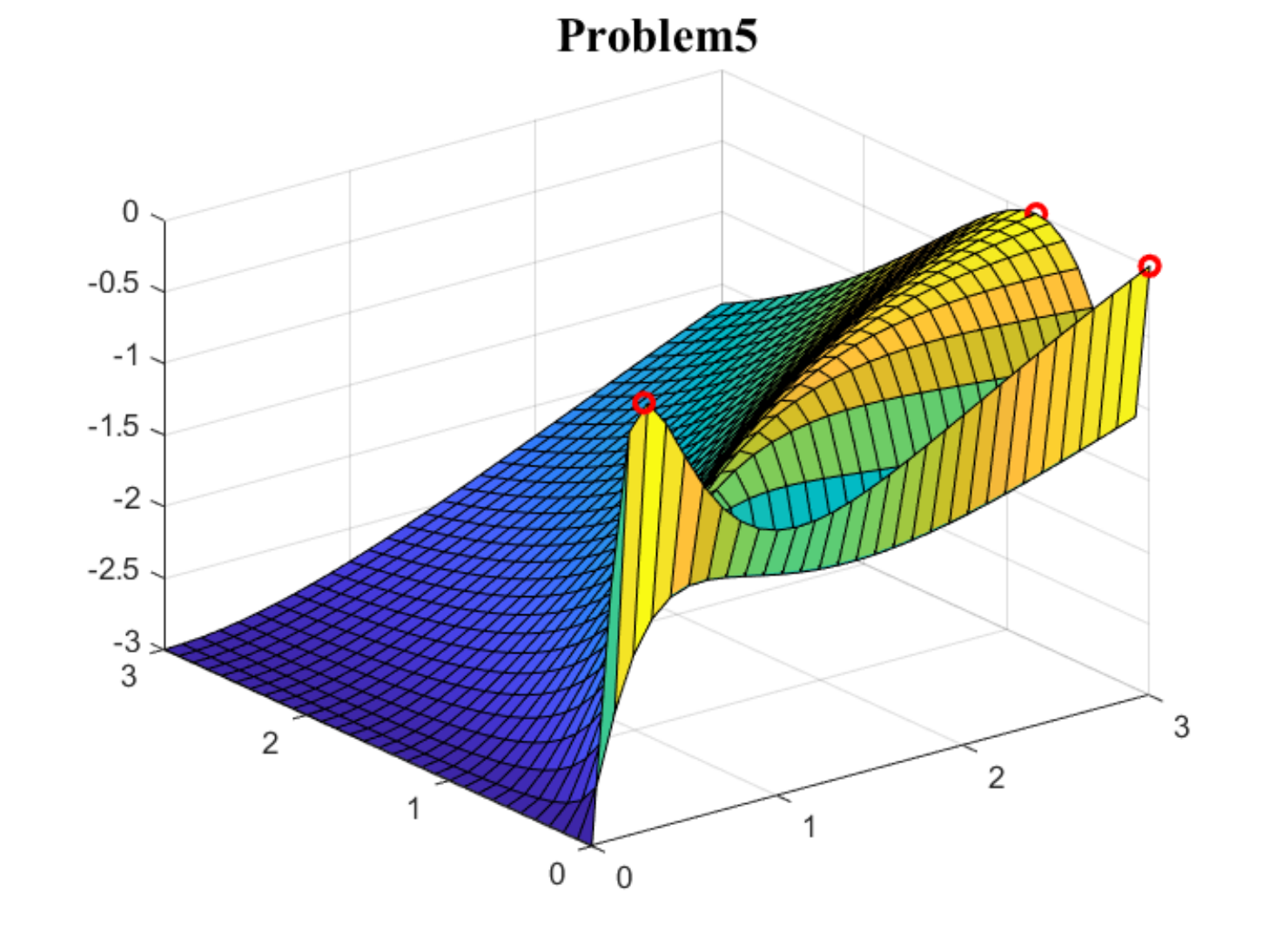}
    \label{DPro5_LSHADE}\hfill
    \includegraphics[width=0.32\linewidth]{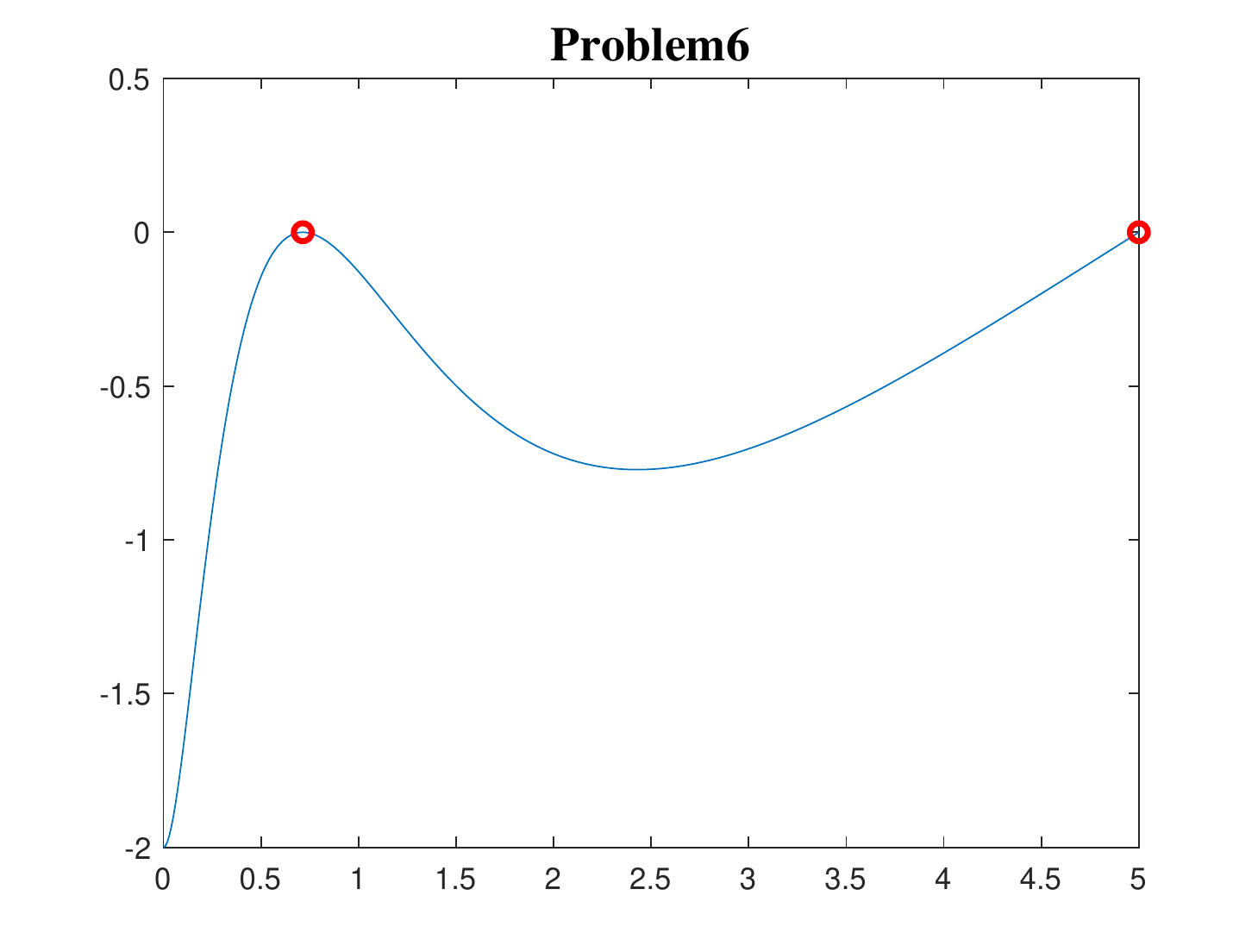}
    \label{DPro6_LSHADE}\hfill
    \includegraphics[width=0.32\linewidth]{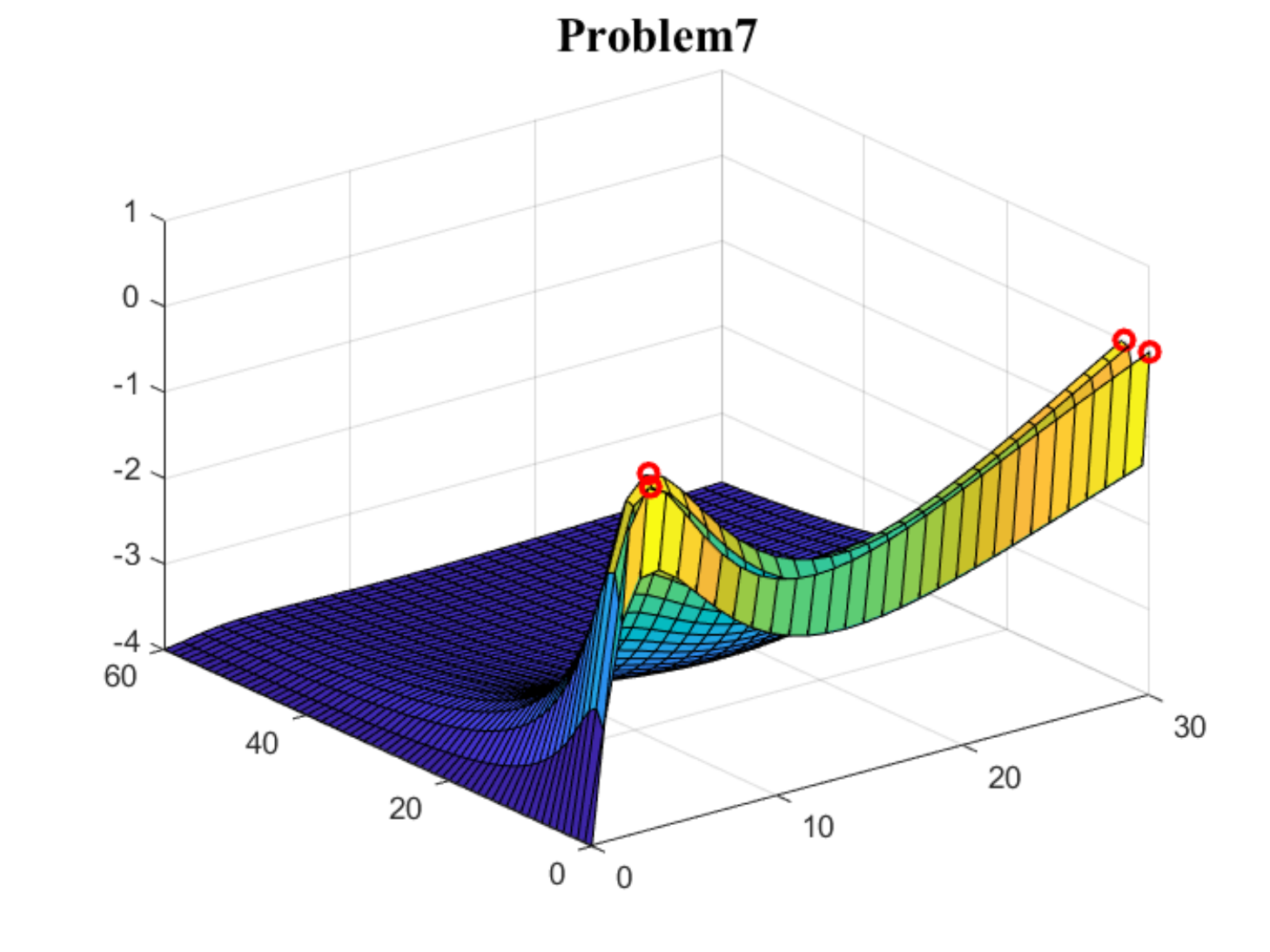}
    \label{DPro7_LSHADE}\\
	\caption{The sensitivity functions of the LSHADE-generated designs for Problems 1, 2, and 4-7 under the $D$-optimality criterion.}
	\label{figureD_sensitivity}
\end{figure}

\begin{figure}[htbp]
\centering
	\includegraphics[width=0.32\linewidth]{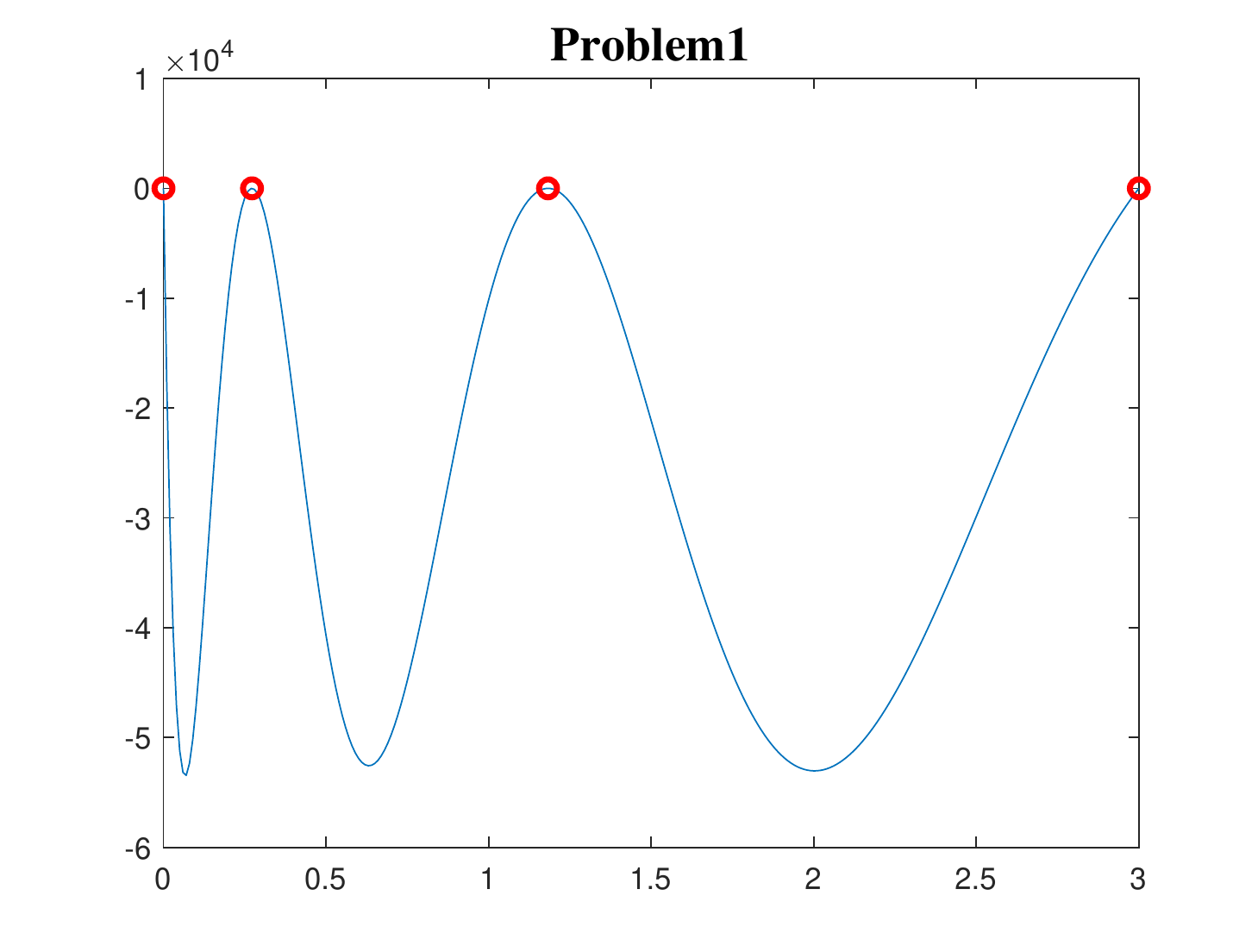}
    \label{APro1_LSHADE}\hfill
	\includegraphics[width=0.32\linewidth]{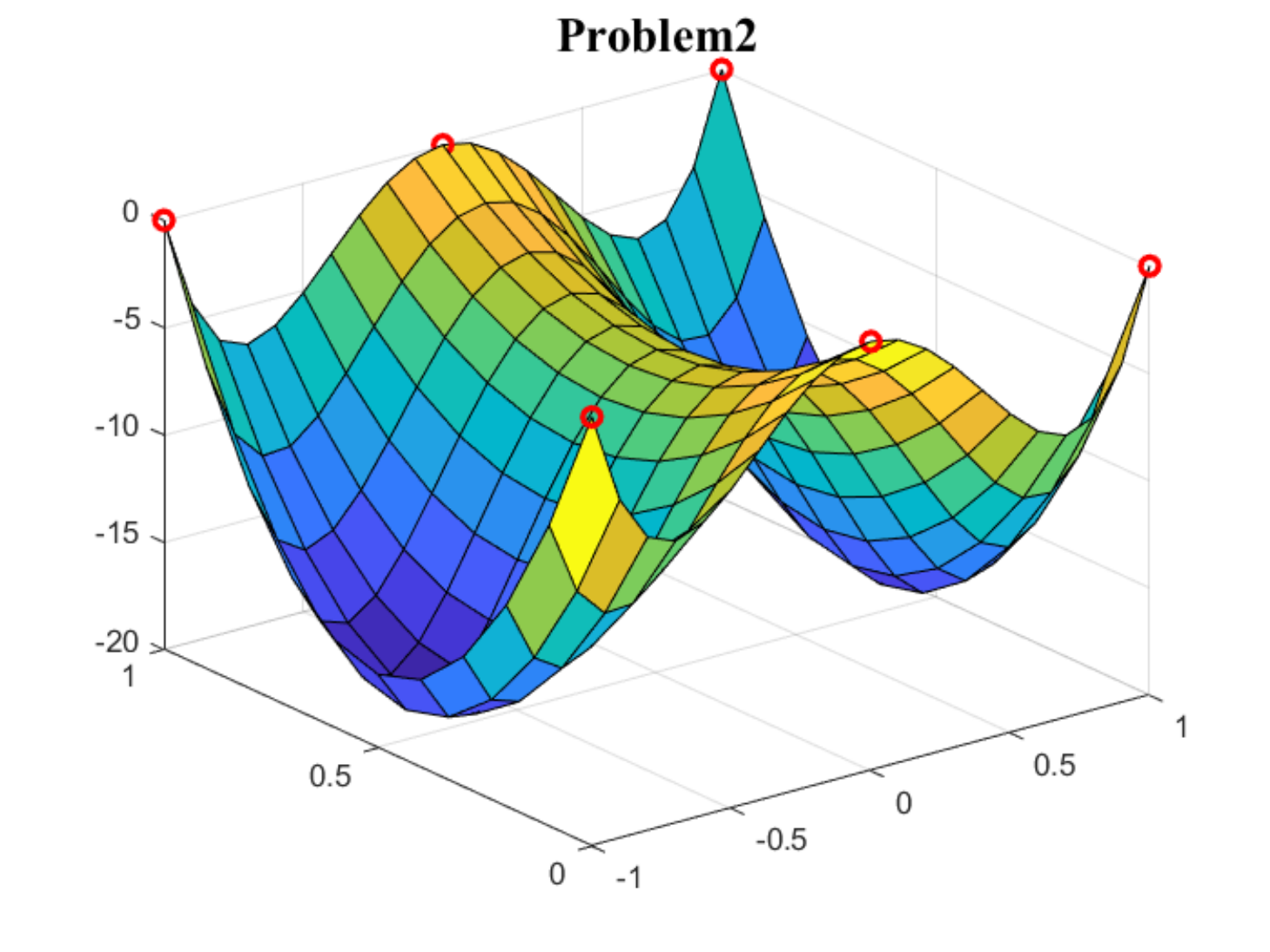}
    \label{APro2_LSHADE}\hfill
    \includegraphics[width=0.32\linewidth]{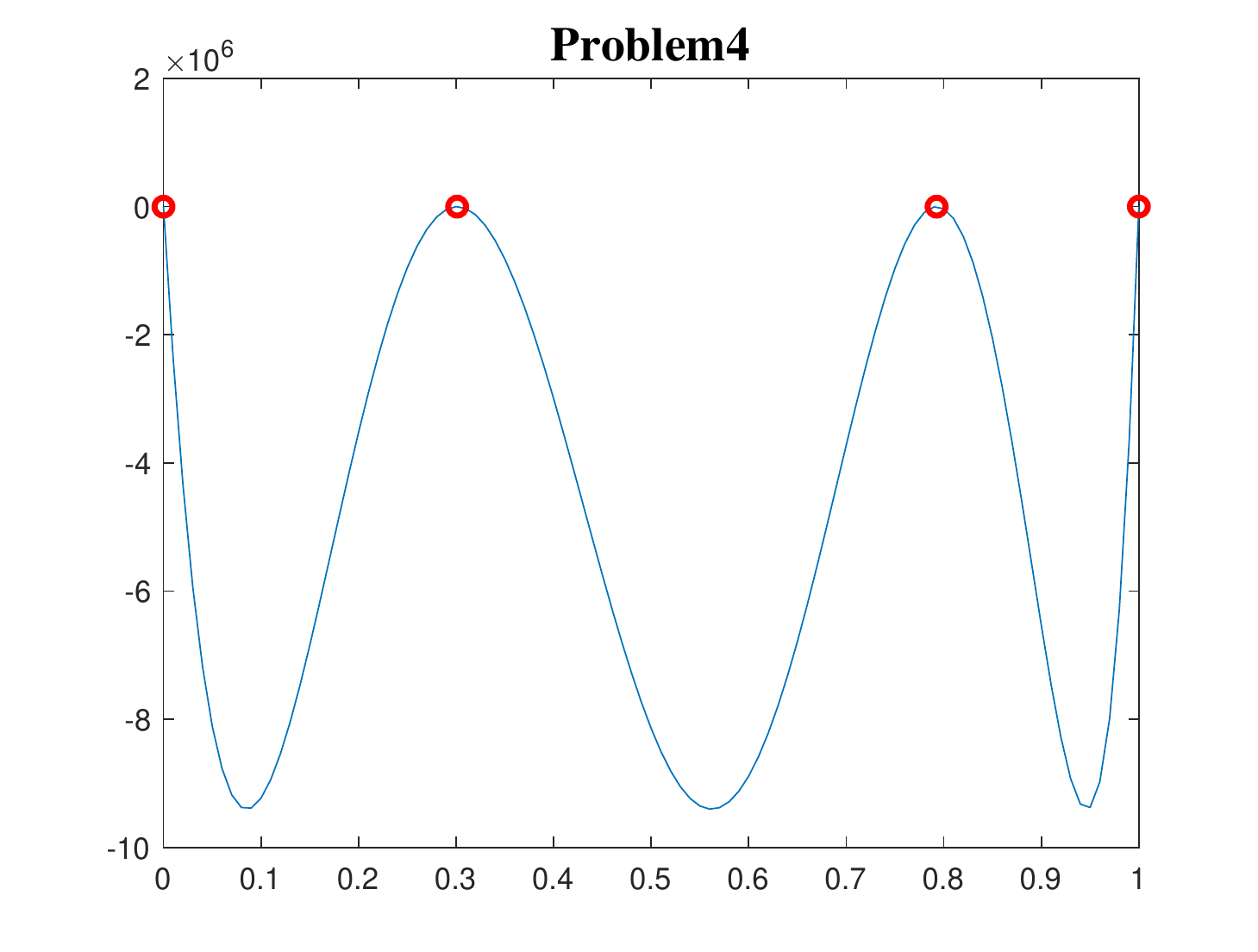}
    \label{APro4_LSHADE}\\
    \includegraphics[width=0.32\linewidth]{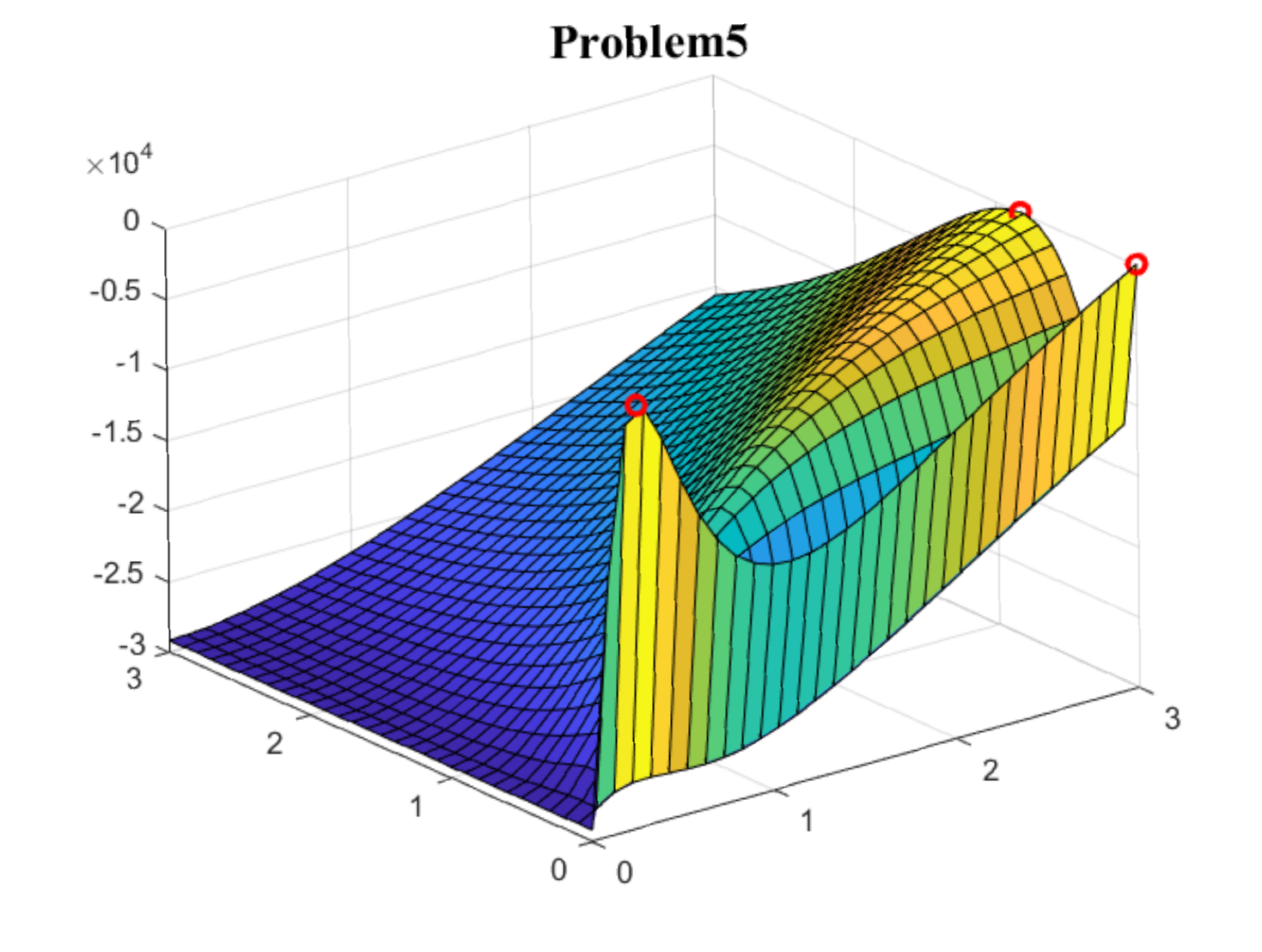}
    \label{APro5_LSHADE}\hfill
    \includegraphics[width=0.32\linewidth]{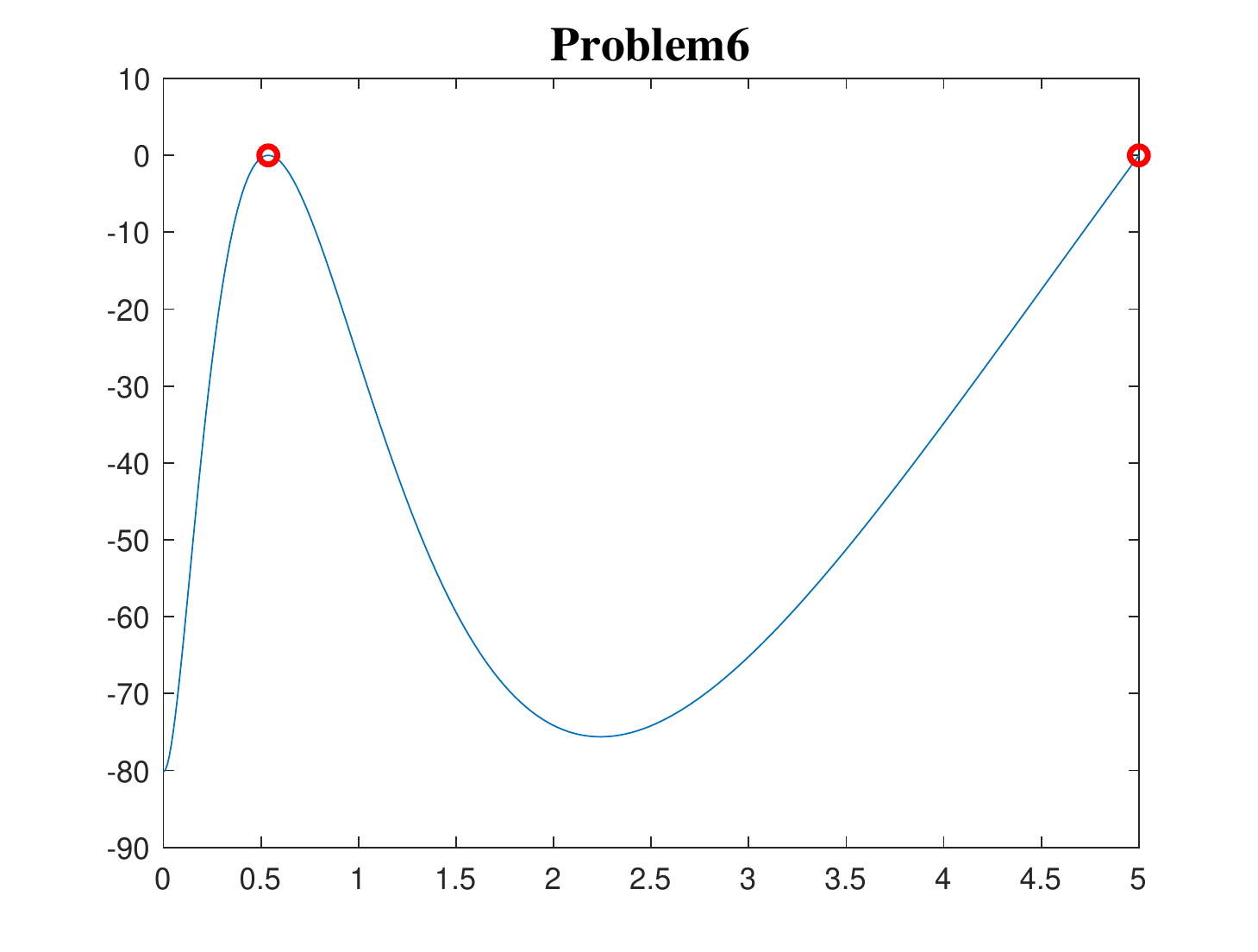}
    \label{APro6_LSHADE}\hfill
    \includegraphics[width=0.32\linewidth]{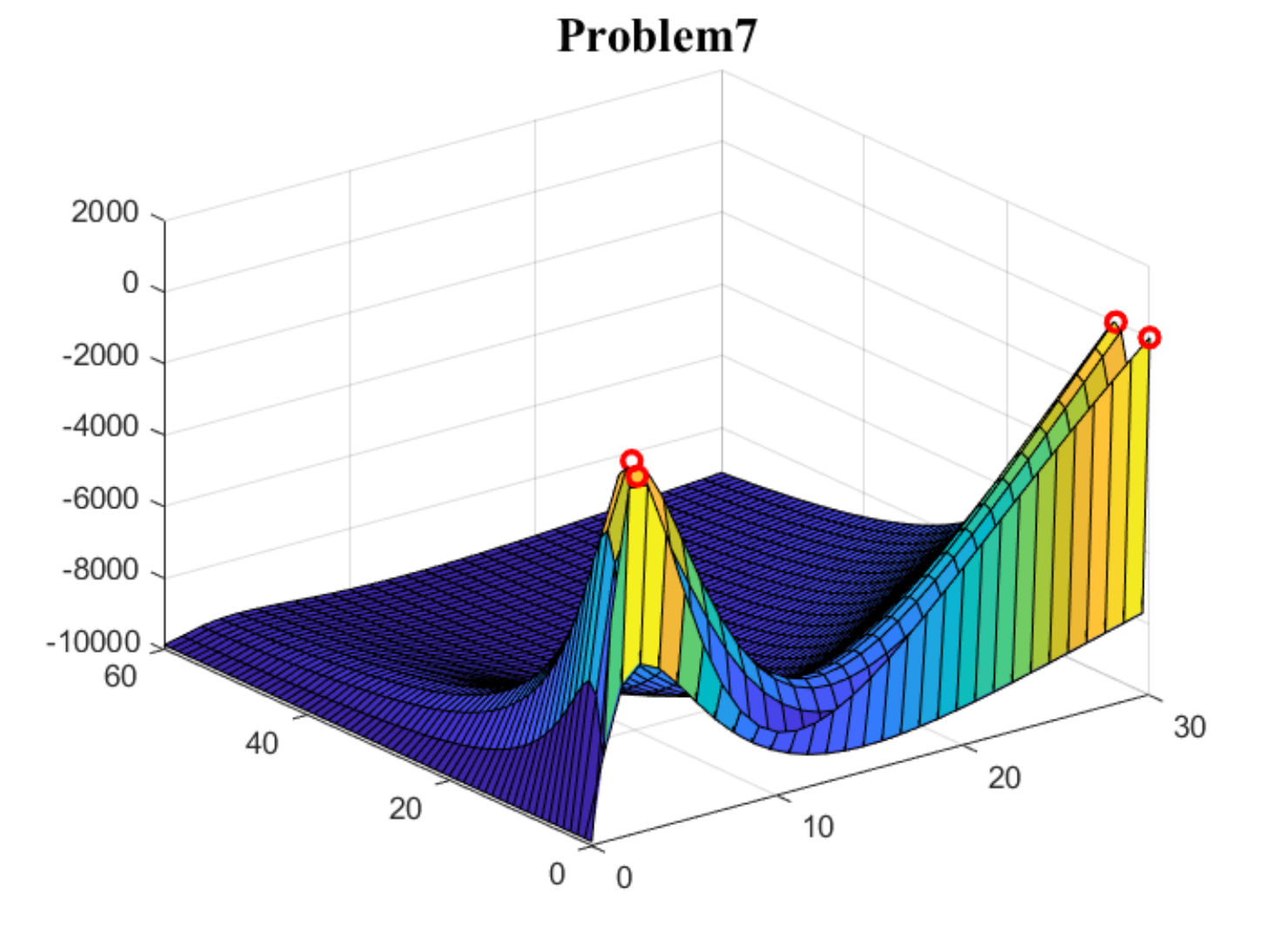}
    \label{APro7_LSHADE}\\
	\caption{The sensitivity functions of the LSHADE-generated designs for Problems 1, 2, and 4-7 under the $A$-optimality criterion.}
	\label{figureA_sensitivity}
\end{figure}

 Figure \ref{figureD_sensitivity} and  Figure \ref{figureA_sensitivity} display, respectively, the sensitivity functions of the LSHADE-generated designs for  Problem 1, 2, and 4-7 under the $D$- and $A$-optimality criteria. We omit sensitive functions  which have larger dimensions because they become harder to appreciate graphically. In the figure, the red dots are support points generated by LSHADE and the all plots in  Figure \ref{figureD_sensitivity} confirm optimality of the designs found by  LSHADE.

\section{Conclusions}
There are many evolutionary and nature-inspired metaheuristic algorithms for general optimization purposes.  DE is one of the most well known evolutionary algorithms  with many variants. A main goal in this study is to ascertain whether various EAs can find $D$- and $A$-optimal designs effectively for various types of statistical models in Table 1.  Performance of several variants of DE for finding $D$- and $A$-optimal designs were compared using several measures  and the LSHADE  variant is the clear winner. In this paper, we propose the repair method to merge similar support points with their corresponding weights based on Euclidean distance and eliminate the corresponding support point with less weight to identify the number of the support points. To handle the infeasible weight solutions of individuals, the repair operation guarantees that the infeasible weight solutions on the $D$- and $A$-optimal design problems are fixed into the feasible weight solutions. To further enrich our optimal design experiment, our work utilizes DE variants to find $D$- and $A$-optimal experimental design. Among the compared algorithms, simulation experiments on 12 statistical models reveal that LSHADE achieves the best performance on the $D$- and $A$-optimal problems.
 
In the future work, we will aims at designing a better evolutionary algorithm for E-optimal design. Next we will construct the different types of optimal design criteria as the multi-objective optimal design problem and design the multi-objective evolutionary algorithm to trade off multi-objective optimal design problem.


\bibliography{mybibfile}

\end{document}